\documentclass{article}

\PassOptionsToPackage{numbers}{natbib}
\usepackage{stfloats}  
\usepackage[preprint]{neurips_2025}
\usepackage{amsthm}
\usepackage{url}

\usepackage[utf8]{inputenc}  
\usepackage[T1]{fontenc}    
\usepackage{amssymb}        
\usepackage{hyperref}       
\usepackage{url}            
\usepackage{booktabs}       
\usepackage{amsfonts}        
\usepackage{nicefrac}       
\usepackage{microtype}     
\usepackage{xcolor}        
\usepackage{siunitx}        
\usepackage{amsmath}
\usepackage{amsthm}
\usepackage{algorithm}
\usepackage{algpseudocode}
\usepackage{multirow}  
\usepackage{tabularray}
\usepackage{bm}        
\usepackage{multirow}   
\usepackage{booktabs}  
\usepackage{graphicx}

\usepackage{booktabs} 
\usepackage{multirow} 
\usepackage{amssymb}   
\usepackage{xcolor}   
\usepackage{amsmath}
\usepackage{amssymb}
\usepackage{mathtools}
\usepackage{amsthm}
\usepackage{nccmath}
\usepackage{tabularx}
\usepackage{subcaption}

\usepackage[textsize=tiny]{todonotes}

 \usepackage[most]{tcolorbox}
\usepackage{listings}
\usepackage{microtype}

 \newtcolorbox{promptbox}[2][]{%
  enhanced,
  breakable,
  colback=gray!3,
  colframe=gray!60,
  boxrule=0.4pt,
  arc=2pt,
  left=6pt,right=6pt,top=6pt,bottom=6pt,
  fonttitle=\bfseries,
  title={#2},
  #1
}

 \lstdefinestyle{promptstyle}{
  basicstyle=\ttfamily\footnotesize,
  columns=fullflexible,
  breaklines=true,
  breakatwhitespace=true,
  keepspaces=true,
  showstringspaces=false,
  frame=none
}

 \usepackage{forest}
\usepackage{tikz}
\usetikzlibrary{arrows.meta, bending}

 \newcommand{\yes}{\checkmark}

 \usepackage{colortbl} 
\definecolor{graybg}{rgb}{0.9, 0.9, 0.9}

\newtheorem{theorem}{Theorem}

\newtheorem{corollary}[theorem]{Corollary}    

\title{Formatting Instructions For NeurIPS 2024}

\author{  
  Sizhe ~Tang
    \\
  The George Washington University\\
  \texttt{s.tang1@gwu.edu} \\
  \And
  Rongqian ~Chen\\
  The George Washington University\\
  \texttt{rongqianc@gwu.edu}
  \And
  Tian ~Lan\\
  The George Washington University\\
  \texttt{tlan@gwu.edu}
}

\begin{document}

\title{Agent Alpha: Tree Search Unifying Generation, Exploration and Evaluation for Computer-Use Agents}
\maketitle

\begin{abstract}
While scaling test-time compute through trajectory-level sampling has significantly improved Graphical User Interface (GUI) agents, the lack of regressive ability prevents the reuse of partial successes and the recovery from early missteps. In this paper, we introduce Agent Alpha, a unified framework that synergizes generation, exploration, and evaluation through step-level Monte Carlo Tree Search (MCTS). It enables active modeling or exploiting structures of the planning space. By integrating alpha-UCT guided search into the interaction loop, Agent Alpha enables deliberate planning, facilitating early pruning of suboptimal branches and efficient prefix reuse. 
We also employ comparison-driven evaluation to mitigate absolute scoring biases and diversity-constrained expansion to maintain a compact, informative search space. Regret bound of alpha-UCT is analyzed. On the OSWorld benchmark, Agent Alpha achieves a state-of-the-art success rate of $\sim$77\%, significantly outperforming trajectory-level baselines under equivalent compute. 
\end{abstract}

\section{Introduction}

\begin{figure}[t]
\centering
\includegraphics[width=\columnwidth]{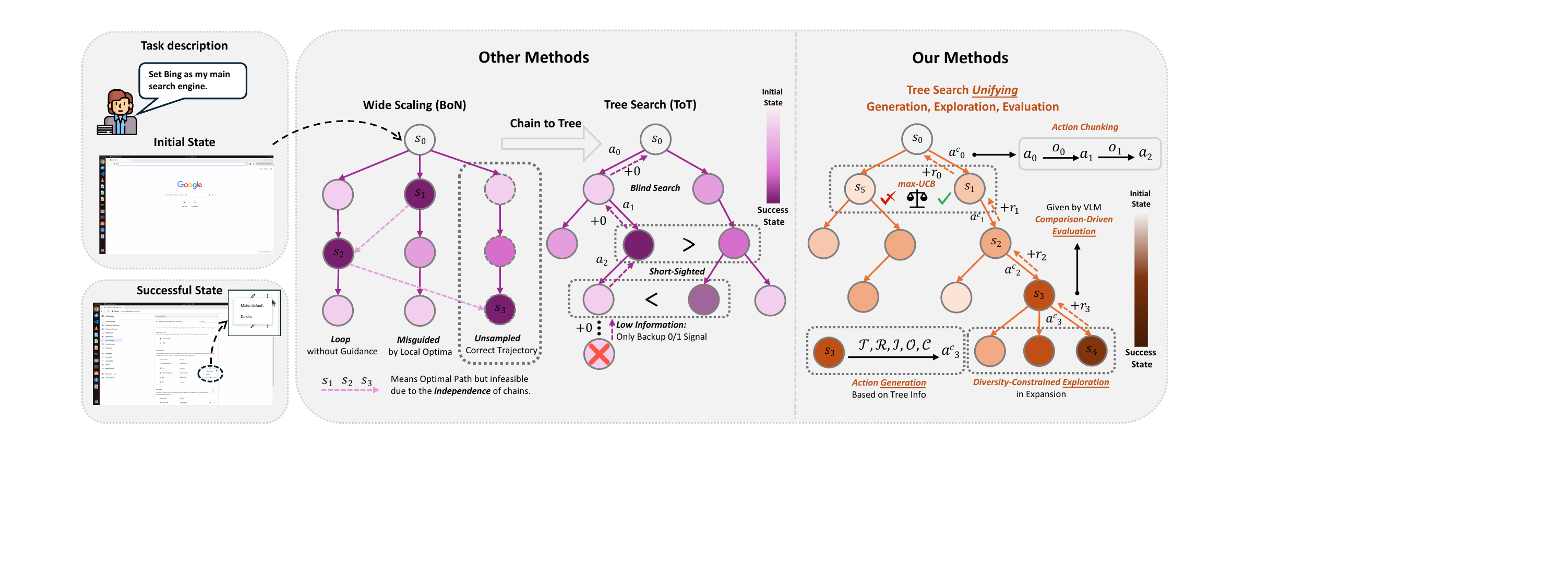}
%\vspace{-0.1in}
\caption{Overview of Agent Alpha. Agent Alpha could provide regressive planning that improve the utilization of generated trajectories with refined exploration and evaluation. We use nodes with deeper color to represent states with higher potential. \(\mathcal{T}, \mathcal{R}, \mathcal{I}, \mathcal{O}, \mathcal{C}\) denotes the information of trajectory, reflection, task instruction, observation, and operation context respectively. Existing approaches  operate primarily as unidirectional processes without the ability to model or exploit structures of the planning space in complex dynamic environments. Agent Alpha addresses this by synergizing generation, exploration, and evaluation capabilities of MLLMs through step-level MCTS under Alpha-UCT bound.}
\label{fig:average_nodes}
\end{figure}

Test-time scaling strategies such as Chain of Thoughts (CoT) \cite{wei2022chain} Tree of Thoughts (ToT) \cite{yao2023tree} and Behavior Best of N (bBoN)\cite{gonzalez2025unreasonable}, have demonstrated remarkable potential in enhancing computer-use agents (CUAs) based on Multimodal Large Language Models (MLLMs). Yet, despite their reasoning capabilities, these scaling approaches face a fundamental limitation in complex dynamic environments: they operate primarily as unidirectional processes \cite{snell2024scaling} without the ability to model or exploit structures of the planning space. In the absence of a mechanism to adapt to dynamic value feedback, existing methods cannot retrospectively evaluate the effectiveness of past actions, either to rectify course of actions when a suboptimal branch has been chosen, or to focus on sampling promising prefixes. Consequently, a single early misstep often cascades into an irreversible failure, preventing the agents from efficiently exploring and converging to the optimal solution. 

To address this, we introduce Agent Alpha, a unified framework that synergizes the generation, exploration, and evaluation capabilities of MLLMs through step-level Monte Carlo Tree Search (MCTS). By transforming linear execution into a regressive planning process, Agent Alpha maximizes the utility of every interaction and leverages evaluations to autonomously recover from early and compounding errors. By integrating search into the interaction loop, Agent Alpha enables fine-grained exploration that facilitates the reuse of promising action prefixes and the early termination of stalled progress. In particular, the search is guided by alpha-UCT bound that uses the maximum value on search-paths to augment exploration and takes into account dependent samples, caused by dependent evaluations and action generations (e.g., due to evolving context and dialog history~\cite{ma2025rethinking, wang2025assessing}).
We model the predictive residual as a martingale difference sequence and derive a regret bound for Alpha Agent by establishing a predictive-corrective coupling between internal reasoning and external evaluation. 
Agent Alpha achieves a tighter confidence bound~\cite{freedman1975tail, kato2020concentration} than standard MCTS. This reduced variance enables the algorithm to prune suboptimal trajectories with significantly fewer samples, efficiently scaling up test-time performance.

Our framework also introduces several methodological innovations designed specifically for the unique challenges of the computer-use domain. First, we propose the search-aware action generation mechanism, which allows agents to aggregate information from failed branches to refine its internal proposal distribution via cross-trajectory-information guided reflection. Second, to prevent the structural redundancy caused by the mode-seeking behavior of aligned models, we implement diversity-constrained expansion, using lexical normalization to construct a compact yet diverse search space. Third, we address the challenge of sparse reward signals in GUI navigation through comparison-driven consistent evaluation. By assessing sibling actions jointly rather than in isolation, our MLLM-based judgment obtains a discriminative relative signal that significantly reduces estimation bias and anchoring effects, otherwise commonly found in absolute scoring.

Empirically, we evaluate Agent Alpha on the OSWorld benchmark \cite{xie2024osworld}, the premier evaluation suite for multimodal computer control across diverse task scenarios. Agent Alpha establishes a new state-of-the-art success rate of \(\sim\)77\%, significantly outperforming trajectory-level scaling methods under comparable computational budgets. 

Our contributions are four-fold:
\vspace{-0.17in}
\begin{itemize}
    \item Agent Alpha synergizes generative, exploratory, and evaluative capabilities of MLLMs through step-level Monte Carlo Tree Search (MCTS), moving beyond unidirectional-process scaling.
    \vspace{-0.07in}
    \item We propose three novel designs for enhancing search-based CUAs, including action generation with tree-informed reflection, comparison-driven consistent evaluation, and diversity-constrained expansion.
    \vspace{-0.07in}
    \item We analyze the regret bound of Alpha-UCT and show that it achieves a tighter confidence bound and superior scaling efficiency compared to standard MCTS.
        \vspace{-0.07in}
    \item Our framework establishes a new state-of-the-art success rate of \(\sim\)77\% on OSWorld.
\end{itemize}

\section{Background}

\paragraph{Computer-Use Agents}
General-purpose computer use is modeled as a Partially Observable Markov Decision Process (POMDP), defined by the tuple $\mathcal{M} = \langle\mathcal{S}, \mathcal{A}, \mathcal{T}, \mathcal{R}, \Omega, \mathcal{O}\rangle$ \cite{fang2024coordinate}. The environment state $s_t \in \mathcal{S}$ represents the underlying status of the operating system and applications, which is not fully visible to the agent \cite{fang2024learning, fang2023implementing}. Instead, at each time step $t$, the agent receives an observation $o_t \in \Omega$ through the observation function $\mathcal{O}(s_t, \mathcal{I})$, where $\mathcal{I}$ denotes the user instruction. This observation $o_t$ comprises the current screenshot and structured accessibility metadata, such as the accessibility tree.

Recent literature has seen a surge in both general-purpose agentic frameworks \cite{song2024trial, wang2024executable, li2025inspo} and specialized GUI agents \cite{niu2024screenagent, rawles2024androidworld, xie2024osworld}. The standard paradigm involves training or prompting a single policy model to generate a sequence of actions. 

\vspace{-0.1in}
\paragraph{Test-Time Scaling.}
Scaling test-time compute for agents has proven effective for enhancing the capabilities of large models \cite{muennighoff2025s1, yu2025optimizing}, typically through parallel sampling \cite{zhu2025scaling, gonzalez2025unreasonable, yu2025look} or iterative refinement \cite{lightman2023verify}. Step-wise parallel decision strategies often fall into vicious cycles that are difficult to break free from. Recent trajectory-level scaling methods like bBoN \cite{gonzalez2025unreasonable}, which prioritize global exploration but suffer from high computational costs and lack of information sharing among different trajectories. Although these methods improves the ability of agents in complex tasks, they have several shortcomings due to the framework structure: 1) They generate actions in the one-directional prediction, which lack of the regressive ability to do in-time error recovery; 2) The diagram of isolated scaling makes it inefficient to make use interaction information across trajectories; 3) Exploration in current methods is not optimized facing the huge action space of computer-use tasks, in which blind sampling can hardly achieve optimal performance.

\vspace{-0.1in}
\paragraph{Monte Carlo Tree Search}
\label{sec:preliminaries_mcts}
Monte Carlo Tree Search (MCTS) \cite{uct, zhang2025lipschitz, zhang2025tail} is a heuristic search algorithm that builds a lookahead tree to estimate the optimal policy \cite{alphazero, tang2025malinzero}. The search proceeds iteratively, where each iteration consists of four phases:  Starting from the root node $s_0$, the algorithm traverses the tree by selecting the child node that maximizes the Upper Confidence Bound applied to Trees (UCT) \cite{uct}. For a state $s$ and action $a$, the selection rule is:
\(    a_t = \arg\max_{a \in \mathcal{A}(s)} \quad Q(s, a) + c_{\text{uct}} \sqrt{\frac{\ln N(s)}{N(s, a)}} ,
\)
where $Q(s, a)$ is the estimated value of taking action $a$ in state $s$, $N(s)$ is the visit count of the parent node, $N(s, a)$ is the visit count of the child node, and $c_{\text{uct}}$ is a constant. Once a leaf node $s_L$ is reached, if it is not a terminal state and has unvisited children, one or more child nodes are added to the tree, corresponding to available actions $a \in \mathcal{A}(s_L)$. From the newly expanded node, the value \(v\) is estimated and propagated backward from the leaf to the root. For every node $(s, a)$ along the traversed path, the visit counts and value estimates are updated:
\(
     N(s, a) \leftarrow N(s, a) + 1, Q(s, a) \leftarrow Q(s, a) + \frac{v - Q(s, a)}{N(s, a)}.
\)

\section{Methodology}

\begin{figure*}[t]
\centering
\includegraphics[width=1\textwidth]{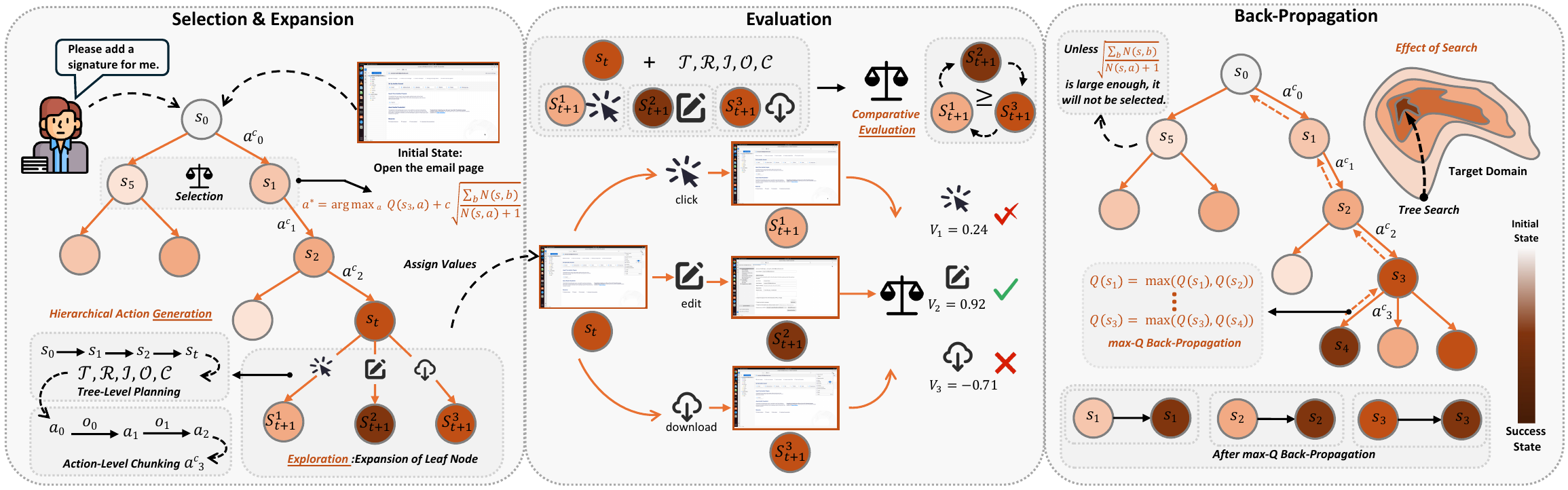}
\caption{The detailed search process applied in Agent Alpha. The process consists of \textit{Selection}, \textit{Expansion}, \textit{Evaluation}, and \textit{Back-Propagation}. The search process will stop if the budget runs out or the task is successful.}
\label{fig:framework}
\end{figure*}

\subsection{Overview of Agent Alpha}

\vspace{-0.1in}
\paragraph{Problem Statement}
In this work, we consider the transition function $g(s_{t+1}|s_t, a_t)$ as deterministic but unknown, governed by the system's logic. For computer-use tasks, rewards $r \in \mathcal{R}$ tend to be sparse, typically satisfying $r = 0$ for $t < T$ and only being obtained at the end of execution. Following recent advancements in GUI agents \cite{peng2024grounding, qin2025ui}, we instantiate our agent using a Vision-Language Model (VLM), denoted as $\pi_\theta$. The action space $\mathcal{A}$ consists of precise computer commands, such as \texttt{click(x, y)}, \texttt{type(text)}, or \texttt{scroll(direction, length)}. In standard reactive paradigms like Chain-of-Thought (CoT) \cite{yao2022react}, the agent generates actions autoregressively conditioned on the interaction history $h_t = (o_0, a_0, \dots, o_t)$ as 
\(
    p(a_t | h_t, \mathcal{I}) = \sum_{\tau_t} \pi_\theta(a_t | \tau_t, h_t, \mathcal{I}) \pi_\theta(\tau_t | h_t, \mathcal{I})
\)
where $\tau_t$ represents the internal reasoning trace generated based on the trajectory and $a_t$ is the predicted action. This process repeats until the agent emits a special \texttt{terminate} token or reaches the maximum step limit $T$.

\vspace{-0.1in}
\paragraph{Unifying Generation, Exploration, and Evaluation.}
To address the challenges in long-horizon computer-use tasks, we propose Agent Alpha, an integrated MCTS-based framework that synergizes the generation, exploration and evaluation capabilities of LMMs through deliberate planning. Viewing the initial task state as the root node, Agent Alpha constructs a search tree iteratively until the search budget runs out or tasks are completed. Each iteration consists of three phases: \textit{Selection}, \textit{Expansion}, and \textit{Back-Propagation}. During the \textit{Selection} stage, the agent traverses the tree to choose the most promising transition action \(a^\prime\) by maximizing our proposed customized Alpha-UCT metric: 
\begin{equation}
a^\prime = \arg\max_{a \in \mathcal{A}(v)} \mbox{Alpha-UCT}(v,Q,N)
\end{equation}

In the \textit{Expansion} phase, we apply the \(\pi_\theta\) empowered by LLMs to sample the set of potential actions \(\mathcal{A}^{(i)} = \{a^{(i)}_1, a^{(i)}_2,\dots,a^{(i)}_k \}\). Then, to avoid semantically repeated nodes in the tree that compromises the search efficiency, only nodes with actions in the normalized set \(\mathcal{\bar A}^{(i)} = \phi(\mathcal{A}^{(i)})\) will be added to the tree, where \(\phi(\cdot)\) is a normalization operator to move the repeated actions. Then, each added node \(v\) will be assigned values \(Q(v)\) via the comparative judgment. For the \textit{Back-Propagation}, statistics (e.g., value and visiting count) of nodes in the search path will be updated according to the max-value back-propagation.

\vspace{-0.1in}
\paragraph{Novel Search Designs} Our framework introduces several innovations designed specifically for the unique challenges of the computer-use domain. {The action generation of Agent Alpha is based on the tree-level dynamic reflection, which gathers more comprehensive information starting from the search implicitly compared with existing single-pass reflection. The proposed diversity-constraint exploration makes the search focus on the most potential and meaningful computer-use path, avoiding the redundancy and inefficiency incurred by standard MCTS. To distinguish actions with subtle difference and obtain consistent guidance for the search, we employ the comparison-driven evaluation instead of independent scoring for nodes. For computer-use tasks, a semantically meaningful action tends to consists of several atomic actions while current agentic frameworks tend to operate these atomic actions separately. This prevents the agent to obtain the long-horizontal planning ability for the hardness to provide proper feedback for the corresponding action. Thus, in Agent Alpha, we replace the single action paradigm for each node with action chunking, which aggregates a sequence of consecutive planning-action pairs to improve the search efficiency.}

\vspace{-0.1in}
\paragraph{Alpha-UCT Bound}   

Standard MCTS algorithms typically view all actions/rewards as independent samples in a multi-armed bandit formulation~\cite{uct}. However, samples in Agent Alpha could be correlated. The tree-level reflection and comparative-driven judgment applied in Agent Alpha lead to dependent evaluations and action generations (e.g., due to evolving context and dialog history~\cite{ma2025rethinking, wang2025assessing}), thus embedding additional unconfirmed knowledge~\cite{kato2020concentration} across search iterations. We take into account such dependent samples in MCTS and develop a new Alpha-UCT bound that ensures more efficient search compared with the standard UCT~\cite{uct}. We further leverage the maximum value on search-paths to augment the exploration term, to more quickly identify missteps and promising prefixes. The regret of Alpha-UCT is analyzed.

\subsection{Search-Aware Action Generation via Tree-Informed Reflection} 

While standard paradigms such as CoT \cite{wei2022chain} improve reliability by generating internal reasoning traces, both action and reasoning generation can only be conditioned on the history in a single pass. Agent Alpha transcends this limitation by employing an informed action generation mechanism to make use of interaction information gathered during the whole tree.

The reflection $\mathcal{R}$ is not a static descriptor, which gathers interaction information over the topology of the search tree in an implicit manner. With Agent Alpha updates increasing statistics of nodes through multiple iterations of search, the trial-and-error information of all paths in the tree is distilled into the selection. 
Let $\mathcal{T}_t^{(i)} = \{ (a^{(i)}_{1:k}, o^{(i)}_{1:k}, Q^{(i)}_{1:k}) \}_{k=1}^t$ be the trajectory of actions, observations, and node values starting from root to step \(t\) in search iteration $i$.  We can note that \(\mathcal{T}^{(i)}_t\) is obtained based on all the trajectories from the first iteration, that is 
\begin{equation}
    \mathcal{T}^{(i)}_t = f_{\operatorname{search}} (\mathcal{T}^{(i-1)}_{\operatorname{leaf\_depth}}),
\end{equation}
which implies that the trajectory information can be inherited across search iterations.

Let \(\mathcal{R}^{(i)}\) denote the reflection used in the \(i\)-th iteration of search. Since the reflection is only generated in the leaf node, then we get
\begin{equation}
    \mathcal{R}^{(i)} = f_{\operatorname{reflect}} \left(\mathcal{T}^{(i-1)}_{\operatorname{leaf\_depth}}, \mathcal{I}\right). 
\end{equation}
Hence, even though the agent will follow the search path it took before, the reflection for each node does not keep the same. The action generator synthesizes the collective intelligence $\mathcal{R}_t^{(i)}$ by aggregating insights from diverse explored branches. This process ensures that the reasoning prior is refined by distilled guidelines in the form of reflection from prior tree search experiences, optimizing the proposal distribution against known failure modes and stalled progress identified during simulation. 

Let \(\operatorname{ctx^{(i)}}\) denote the context of the current workspace such as clickable buttons in the website. The planning \(\mathcal{P}_t^{(i)}\) of the leaf node in iteration \(i\) is predicted as 
\begin{equation}
    \mathcal{P}^{(i)}_t = f_{\operatorname{plan}}\left(\mathcal{I}, o^{(i)}_{t},\mathcal{R}^{(i)}_t, \operatorname{ctx}^{(i)}_t \right).
\end{equation}
The planning, observation and context are input into the grounding model to generate the specific action such as \texttt{click(300, 450)} by
\begin{equation}
    a_t = f_{\operatorname{predict}} \left(\mathcal{P}^{(i)}_t, o^{(i)}_t \right).
\end{equation}

To form an action chunking, the above atomic action \(a_t\) and the resulting observation \(o_t\) will be generated repeatedly in a sequence manner. Then, the atomic actions will be grouped to an action chunk and the last resulted state will be assigned to the transited node. In this manner, action chunking only make the agent do the planning with a longer vision and avoid the case that a certain atomic action is optimal judged meaningless by LLMs.

\subsection{Diversity-Constrained Exploration}
To expand a tree, Agent Alpha needs to do the above action generation for the same leaf node, which incurs the problem of efficient exploration. That is, given identical state $s^{(i)}_t$, independent sampling $a^{(i)}_{t,1}, \dots, a^{(i)}_{t,k} \sim \pi_\theta(\cdot|s^{(i)}_t)$ often collapses onto a narrow subset of high-probability tokens. This results in structural redundancy where sibling branches explore overlapping regions of the action space, creating an exploration illusion where the search tree grows in size without a corresponding increase in state-space coverage. 

We address this through a constrained exploration operator that enforces uniqueness among sibling nodes via online semantic de-duplication. Let \(\phi: \mathcal{A}^{(i)} \to \mathcal{\bar A}^{(i)}\) be a normalization function that maps a raw action set to its functional representation set by stripping superficial formatting variations while preserving essential elements such as coordinates and command types. For example, if the planner indicates that agent needs to click the \texttt{download} button, then the action \texttt{click(300, 450)} and \texttt{click(300, 452)} should share the same function and one of them should be rejected for sampling. During the expansion of node $v$, candidate actions are elicited sequentially. An action $a^{(i)}_{k}$ is admitted to the search tree if and only if its normalized representation is unique relative to previously accepted siblings:
\begin{equation}
    \mathcal{\bar A}^{(i)} = \{ a^{(i)}_{k} \sim \pi_\theta \mid \forall j < k, \phi(a^{(i)}_k) \neq \phi(a^{(i)}_j) \}
\end{equation}
This mechanism yields an adaptive branching factor $b^*(v) = |\phi(\mathcal{\bar A}^{(i)})| \le K$. If a state presents a narrow decision bottleneck, the exploration gracefully yields fewer branches, thereby concentrating the search compute on genuinely distinct decision paths. This ensures that the search budget is allocated proportionally to actual uncertainty, transforming the expansion phase from a blind sampling process into a diversity-aware construction of a compact and informative search space.

\subsection{Comparison-Driven Consistent Evaluation}

Evaluating the quality of candidate actions is central to effective tree search, yet language models exhibit significant range instability and scale-interpretation bias when tasked with independent scalar scoring \cite{lorentz2016using}. In the context of GUI navigation, absolute judgments often suffer from overconfidence or excessive hedging, where minor prompt perturbations lead to unpredictable score shifts \cite{gu2024survey}. This lack of calibration is particularly detrimental when the search algorithm must resolve subtle decision boundaries between semantically similar sibling actions, as independent queries fail to provide a consistent reference frame.

Agent Alpha addresses this through comparison-driven consistent evaluation, a relative assessment paradigm that transforms the evaluation phase into a joint assessment task. Instead of scoring nodes in isolation, the MLLM-based evaluator jointly assesses the set of sibling nodes expanded from a common parent. Let \(v\) be the parent node at state $s_t$, and $\{v'_{1}, \dots, v'_{k}\}$ be its children generated by actions $\{a_{t,1}, \dots, a_{t,k}\}$. The evaluator $f_{\text{judge}}$ computes the node values as a normalized vector:
\begin{equation}
    [V(v'_1), \dots, V(v'_k)] = f_{\text{judge}}(s_t, \{(a_{t,j}, o'_{t,j})\}_{j=1}^k, \mathcal{I})
\end{equation}
where each $(a_{t,j}, o'_{t,j})$ pair represents the proposed action and its resulting interaction consequence, grounded in the user instruction $\mathcal{I}$. The \(Q\) values of expanded nodes are initialized with predicted values. 

In standard MCTS, the predicted node values will be updated as the cumulative mean value through the search path. However, some slight operation difference in the computer-use scenario will incur huge difference to the outcome. Thus, the metric of mean value will make it longer to expose the wrong operation with low value since the negative feedback is easy to be hidden by the high \(Q\)-value of the previous path. To this end, we update the exploitation term in UCT-type bound using the max-value of nodes in the search path. For each node \(v\) in the search path except for the leaf, the \(Q\) value is updated as
\begin{equation}
    Q(v) = \max (Q(v), V_{\operatorname{new}})
\end{equation}
where \(V_{\operatorname{new}}\) is the new value to be updated into the tree.
In this way, the agent can select the most potential action from the perspective of the whole tree, which prevents the case that the fatal action can not be discriminated. To better distinguish the meaningful state-action pair and the wrongly guided one, we set the judgment score for each node in the range of \([-1, 1]\).

\subsection{Alpha-UCT and Regret Analysis}
\label{sec:theoretical_analysis}

Standard MCTS algorithms typically view actions/rewards as independent samples~\cite{uct}. However, Agent Alpha leverages the dynamic reflection and comparative evaluation mechanism, resulting in correlated samples during search, due to evolving context and dialog history~\cite{ma2025rethinking, wang2025assessing}). More precisely, reward correlation/dependence arises in the evaluation since rewards are evaluated in a comparative manner; 
the reflection mechanism introduces a temporal dependency across search iterations, altering the independence assumption of exploration and action sampling. Such correlation/dependence may serve as unconfirmed knowledge~\cite{kato2020concentration} in MCTS.

We model the value estimation process as a martingale concentration problem. Let $X_t$ be the evaluation score and $\hat{\theta}_t$ be the reflection prior generated by the agent. The efficiency of the search is fundamentally governed by the residual variance of the prediction error rather than the raw variance of the environment:
\begin{equation}
        \sigma_{\operatorname{res}}^2 = \mathbb{E}\left[ (X_t - \hat{\theta}_t)^2 \mid \mathcal{F}_{t-1} \right].
\end{equation}
Here, the filtration \(\mathcal{F}_{t-1}\) represents the history of observed trajectories and reflections. Unlike independent samples where variance is constant, the conditional variance here decreases as the reflection \(\hat\theta_t\) becomes more accurate.

Leveraging the concentration inequality~\cite{kato2020concentration} for martingales, we obtain Alpha-UCT bound as action our selection policy as follows:
\begin{equation}
\label{eq:alpha_uct_max}
    a^* = \arg\max_{a \in \mathcal{A}(v)} \quad  Q_{\text{max}}(v, a) + c \sqrt{\frac{\sum_b N(v,b)}{N(v, a) + 1}}.
\end{equation}
Here $Q_{\text{max}}(v, a)$ denotes the maximum evaluation score observed in the subtree of action $a$:
\begin{equation}
    Q_{\text{max}}(v, a) = \max_{k \in \mathcal{T}(v, a)} V_k,
\end{equation}
where $\mathcal{T}(v, a)$ is the set of all completed trajectories passing through node $(v, a)$, and $V_k$ is the comparative judgment score. Next, we analyze the regret bound of Alpha-UCT.

\begin{theorem}
\label{thm:regret_bound_kato}
\textbf{(Regret Bound with Unconfirmed Knowledge).} 
The cumulative regret $R_T$ for an agent leveraging unconfirmed knowledge satisfies:
\begin{equation}
    R_T \le \sum_{i \in \mathcal{A} \setminus \{i^*\}} \left( \frac{8 \sigma_{\text{res}, i}^2 \ln T}{\Delta_i} + \frac{16 \ln T}{3} + 2\Delta_i \right).
\end{equation}
\end{theorem}

This theorem leads to a critical insight regarding thes complexity of our framework compared to standard methods.

\begin{corollary}
\label{cor:complexity_order}
\textbf{(Asymptotic Regret Complexity).} 
Under the reflection-guided framework of Agent Alpha, the cumulative regret $R_T$ over horizon $T$ with branching factor $K$ satisfies the complexity order:
\begin{equation}
    R_T = \mathcal{O}\left( K \cdot \sigma_{\text{res}}^2 \cdot \ln T \right).
\end{equation}
\end{corollary}

This implies that the quality of the agent's reflection directly dictates search speed: as the reflection $\hat{\theta}$ aligns closer to the outcome $X$, $\sigma_{\text{res}}^2 \to 0$, effectively minimizing the regret. To quantify this advantage, we compare the sample complexity of Alpha-UCT against the standard UCT baseline that considers all actions are independent.

\begin{corollary}
\label{cor:efficiency_gain}
\textbf{(Efficiency Gain via Variance Reduction).} 
Let $\sigma_X^2 = \text{Var}(X)$ be the unconditional variance of the evaluator (blind search variance). The effective regret reduction of Agent Alpha relative to standard UCT is proportional to the ratio of the residual variance to the unconditional variance:
\begin{equation}
    \frac{R_T^{\text{Alpha}}}{R_T^{\text{UCT}}} \approx \frac{\sigma_{\operatorname{res}}^2}{\sigma_X^2} < 1.
\end{equation}
\end{corollary}

This corollary theoretically validates that Agent Alpha converts predictive accuracy into search efficiency. The term $\sigma_{\operatorname{res}}^2 / \sigma_X^2$ represents the fraction of uncertainty remaining after accounting for the explored search path.

\section{Experiments and Analysis}

\begin{table*}[t]
\centering
\small
\setlength{\tabcolsep}{3pt}
\caption{Success rate comparison on OSWorld tasks.
Agent Alpha achieves superior performance across diverse domains by using GPT-5.2 base model with optimized parameters and action chunking.
\textbf{\textcolor{red}{Red}} indicates the best performance, and \textbf{\textcolor{blue}{Blue}} indicates the second best.}
\label{tab:osworld_100}
\begin{tabularx}{\textwidth}
{l >{\centering\arraybackslash}X *{10}{>{\centering\arraybackslash}X}}
\toprule
\textbf{Methods}
& \textbf{Chrome}
& \textbf{Calc}
& \textbf{Impress}
& \textbf{Writer}
& \textbf{GIMP}
& \textbf{VSCode}
& \textbf{Multi}
& \textbf{Thunder}
& \textbf{OS}
& \textbf{VLC}
& \textbf{Avg.} \\
\midrule
Agent s3 w/ bBoN (N=10)
& 62.96 & \textcolor{red}{87.23} & \textcolor{blue}{72.26} & \textcolor{blue}{86.83} & 73.08 & 82.61 & \textcolor{red}{63.90} & \textcolor{blue}{73.33} & \textcolor{blue}{79.17} & 62.94 & \textcolor{blue}{72.58} \\
UiPath Screen Agent
& \textcolor{red}{69.48} & \textcolor{blue}{80.43} & 70.19 & 69.57 & 69.23 & 78.26 & 52.97 & \textcolor{blue}{73.33} & 70.83 & 60.82 & 67.14 \\
OS-Symphony
& 62.96 & 76.60 & 48.81 & 78.13 & \textcolor{blue}{80.77} & 78.26 & 55.30 & \textcolor{red}{80.00} & 78.26 & 61.53 & 65.77 \\
GBOX Agent
& 62.96 & 74.47 & 63.74 & 68.18 & 76.92 & \textcolor{blue}{91.30} & 49.73 & 66.67 & 70.83 & 46.29 & 64.22 \\
GTA1
& 58.61 & 63.83 & 65.53 & 60.74 & 76.92 & 82.61 & 50.91 & \textcolor{red}{80.00} & \textcolor{blue}{79.17} & 57.88 & 63.41 \\
claude-sonnet-4-5
& 62.96 & 72.34 & 68.00 & 82.48 & 53.85 & 73.91 & 49.54 & 60.00 & 70.83 & 58.18 & 62.88 \\
CoACT-1
& 54.26 & 70.21 & 50.28 & 73.91 & 65.38 & 78.26 & 47.87 & \textcolor{blue}{73.33} & 75.00 & \textcolor{red}{71.94} & 60.76 \\

\midrule
\textbf{Agent Alpha (Ours)}
& \textbf{\textcolor{blue}{67.39}} & \textbf{\textcolor{red}{87.23}} & \textbf{\textcolor{red}{78.72}} & \textbf{\textcolor{red}{91.30}}
& \textbf{\textcolor{red}{96.15}} & \textbf{\textcolor{red}{100}} & \textbf{\textcolor{blue}{59.14}}
& \textbf{\textcolor{red}{80.00}} & \textbf{\textcolor{red}{95.83}} & \textbf{\textcolor{blue}{70.58}} & \textbf{\textcolor{red}{77.29}} \\
\bottomrule
\end{tabularx}
\end{table*}

\begin{table*}[t]
\centering
\small
\setlength{\tabcolsep}{2pt}
\caption{Improvement Analysis: Breakdown of tasks failed by Agent S3 but successfully solved by Agent Alpha across 10 domains.}
\label{tab:s3_fail_improvement_breakdown}
\begin{tabularx}{\textwidth}{l *{11}{>{\centering\arraybackslash}X}}
\toprule
\textbf{Metric}
& \textbf{Chrome} & \textbf{Calc} & \textbf{Impress} & \textbf{Writer} & \textbf{GIMP}
& \textbf{VSCode} & \textbf{Multi} & \textbf{Thunder} & \textbf{OS} & \textbf{VLC} & \textbf{Total} \\
\midrule
\# S3 Failed
& 21 & 16 & 23 & 7 & 12 & 6 & 58 & 5 & 6 & 11 & 165 \\
\# Solved by Alpha
& 5 & 8 & 4 & 3 & 11 & 6 & 7 & 1 & 5 & 6 & 56 \\
\midrule
\textbf{improvement Rate (\%)}
& 23.8 & 50.0 & 17.4 & 42.9 & 91.7 & 100.0 & 12.1 & 20.0 & 83.3 & 54.5 & 33.9 \\
\bottomrule
\end{tabularx}
\end{table*}

\begin{table}[t]
\centering
\small
\setlength{\tabcolsep}{3pt}
\caption{Performance comparison between Agent Alpha(Exp. Node=5, Action Chunking =1) and Agent S3(N=3), both use gpt-5-mini as base model.}
\label{tab:s3_fail_alpha_success}
\begin{tabularx}{\columnwidth}{l *{4}{>{\centering\arraybackslash}X}}
\toprule
\textbf{Method} & \textbf{Success Rate (\%)} & \textbf{Alignment Score(\%)} & \textbf{Average Step} & \textbf{Avg Time (s)} \\
\midrule
Agent S3    & 54.29 & 69.5 & 8.88 & 313.37 \\
\textbf{Agent Alpha} & \textbf{64.27} & \textbf{82.2} & \textbf{7.98} & \textbf{1116.5} \\
\bottomrule
\vspace{-0.5cm}
\end{tabularx}
\end{table}

\subsection{Experiment Setting.}

\paragraph{Benchmark.}
We conduct our experiments on the OSWorld benchmark \cite{xie2024osworld}, a comprehensive evaluation platform designed for multimodal agents in real computer environments. The benchmark encompasses diverse task scenarios, ranging from fundamental OS operations and daily office workflows to professional software manipulation and complex cross-application tasks. 
%Following established evaluation protocols \cite{xie2024osworld}, we exclude the subset of multi-application tasks requiring Google Drive credentials, as these are incompatible with the standard sandbox environment.

\paragraph{Baselines.}
We compare the performance and efficiency of Agent Alpha against state-of-the-art methods listed on the OSWorld leaderboard, including UiPath Screen Agent, Claude Sonnet 4.5, Agent S3 ~\cite{gonzalez2025unreasonable}, CoAct-1 ~\cite{song2025coact}, GTA1 ~\cite{yang2025gta1}, OS-symphony~\cite{yang2026symphony}) and GBOX Agent.

\subsection{Main Results}

In this section, we evaluate the performance of Agent Alpha against seven baseline models on the OSWorld benchmark. The Agent Alpha experimental setup utilizes GPT-5.2 with optimized hyperparameters: expansion factor of 5 and maximum MCTS iterations of 20. For complex, long-horizon tasks (such as Multi-apps), we utilize action chunking number of 5 and reduce iterations to 15 to balance efficiency.

As illustrated in Table 1, Agent Alpha establishes a new state-of-the-art with an average Success Rate (SR) of 77.29\%. This outperforms the strongest baseline, Agent s3 w/ Opus 4.5 + GPT-5 bBoN ($N=10$)~\cite{gonzalez2025unreasonable}, by a margin of 4.71\%. Notably, given that human performance on this benchmark is approximately 72\%~\cite{xie2024osworld}, Agent Alpha successfully surpasses human-level capability.

For the category-specific performance, our method exhibits robust generalization capabilities. It achieves the highest success rates(in red) in 7 out of 10 categories, including a 100\% success rate in VSCode and 96.15\% in GIMP. In categories where Agent Alpha does not rank first (Chrome, Multi, and VLC), it consistently secures the second-best performance (highlighted in blue), ensuring reliability across diverse operating system tasks.

\begin{table*}[t]
\centering
\caption{Ablation study on different module configurations. We analyze the impact of each sub-module choice on success rate and runtime.}
\label{tab:ablation}
\setlength{\tabcolsep}{5pt}

\begin{tabular}{l cc cc cc rr}
\toprule
\multirow{2}{*}{\textbf{Setting}} & 
\multicolumn{2}{c}{\textbf{Judge Mech.}} & 
\multicolumn{2}{c}{\textbf{Backup Mech.}} & 
\multicolumn{2}{c}{\textbf{Tree Parallel.}} & 
\multicolumn{2}{c}{\textbf{Results}} \\
\cmidrule(lr){2-3} \cmidrule(lr){4-5} \cmidrule(lr){6-7} \cmidrule(lr){8-9}

 & Comp. & Indep. & Max & Mean & Action & Env. & SR (\%) & Speed \\
\midrule
w/o Comparative Judge & & \yes & \yes&  & \yes & \yes & 57.96 & - \\
w/o Max Backup & \yes & & & \yes & \yes & \yes & 45.42 & - \\
w/ Action Parallelism  & \yes & & \yes &  & \yes &  & - & 4.4x \\
w/ Env. Parallelism  & \yes & & \yes & & & \yes & - & 2.1x \\
\midrule
\rowcolor{graybg}
\textbf{Alpha Agent (Full)} & \yes & & \yes & & \yes & \yes & \textbf{64.27} & - \\

\bottomrule
\end{tabular}
\end{table*}

\subsection{Comparison with SOTA}

Tables \ref{tab:s3_fail_improvement_breakdown} and \ref{tab:s3_fail_alpha_success} compare Agent Alpha against the SOTA baseline, Agent S3. To ensure a fair comparison, both base models use the same GPT-5-mini. The primary distinction lies in the search strategy: Agent S3 uses Best-of-N ($N=3$) and maximum trajectory depth of 30 steps, while Agent Alpha employs MCTS with an expansion factor of 5, max iteration of 20 times and no action chunking. We evaluate performance using Success Rate (SR) and Alignment Score, which measures adherence to human logic. We also report Average Step and Average Time to assess efficiency. These metrics are computed exclusively on successful tasks, and the reported running time explicitly excludes system initialization overhead to focus solely on inference.

As illustrated in Table \ref{tab:s3_fail_alpha_success}, Agent Alpha demonstrates significant improvements over the baseline, achieving a Success Rate of 64.27\% compared to 54.29\% for Agent S3. Our model also exhibits superior reasoning capabilities, evidenced by a marked increase in the Alignment Score (82.2\% vs. 69.5\%) and a reduction in the average number of steps required to solve tasks (7.98 vs. 8.88). However, this performance gain involves a computational trade-off: due to the comprehensive search process, Agent Alpha requires a higher average inference time (1116.5s) compared to the faster, but less accurate, Agent S3 (313.37s). To further investigate the robustness of our approach, Table \ref{tab:s3_fail_improvement_breakdown} provides a breakdown of tasks that Agent S3 failed to solve but were successfully completed by Agent Alpha. Out of 165 tasks failed by the baseline, Agent Alpha successfully succeed on 56, yielding an overall improvement rate of 33.9\%.

\subsection{Analysis of Hyperparameters}

\begin{figure}[t]\centering\includegraphics[width=1\columnwidth]{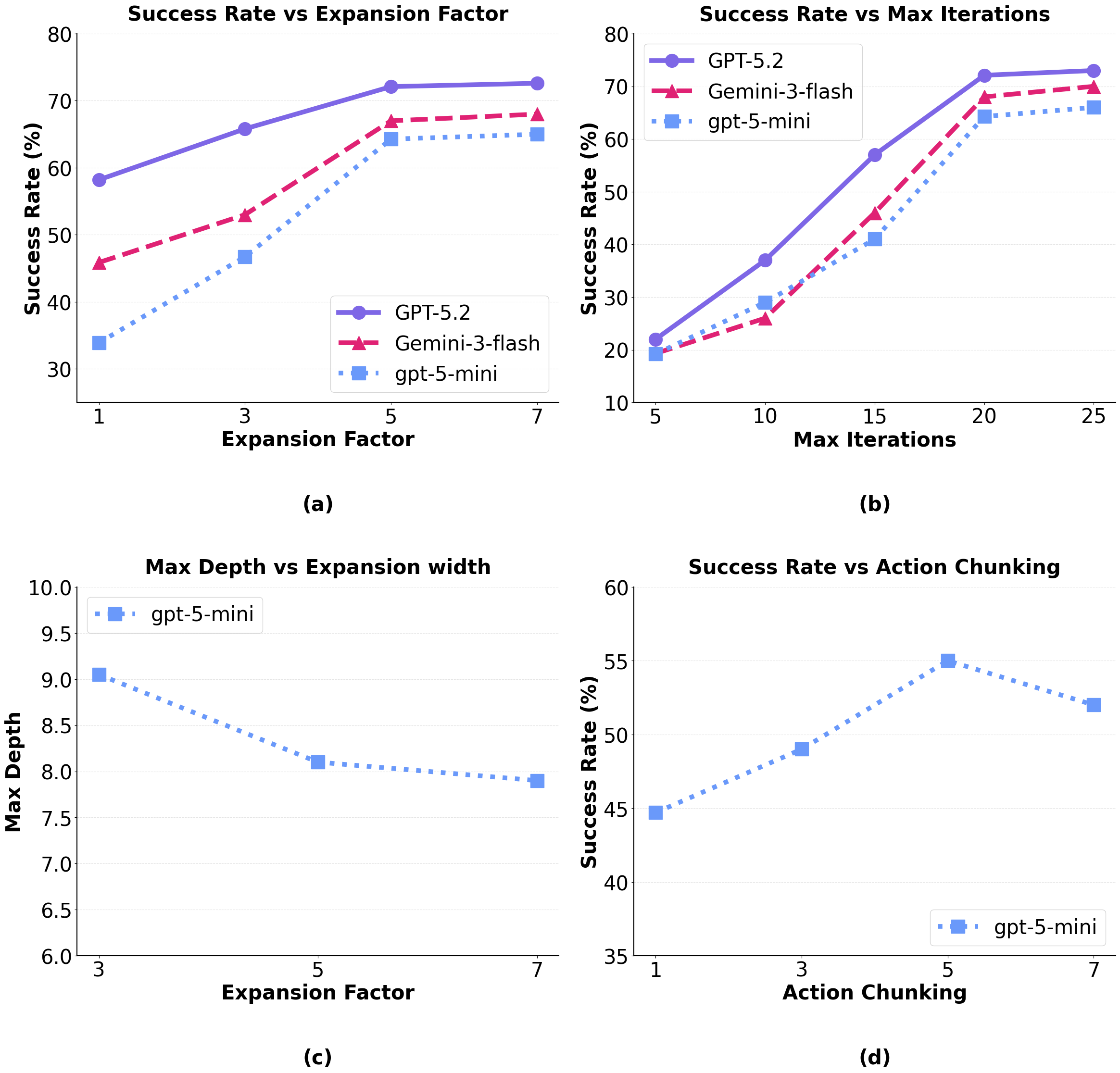}
\caption{Analysis of hyperparameters: (a) Success rate vs. Expansion Nodes, (b) Success rate vs. Maximum Iterations, (c) Max Depth vs. Expansion factor, and (d) Success rate vs. Action chunking on long-horizon tasks.}
\vspace{-0.5cm}
\label{fig:parameter_exp}
\end{figure}

In this section, we conduct an ablation study to evaluate the impact of key hyperparameters on Agent Alpha's performance, as illustrated in Figure \ref{fig:parameter_exp}.
\paragraph{Impact of Expansion factor ($N$):} In Figure \ref{fig:parameter_exp}(a), we analyze the effect of varying the expansion factor while fixing the maximum MCTS iterations to 20. Performance improves significantly as the node count increases from 1 to 5. However, the gain saturates beyond $N=5$. This suggests that for most OSWorld tasks, the optimal action lies within the top-5 probabilities generated by the policy model. Consequently, increasing $N$ beyond 5 offers diminishing returns with unnecessary computational overhead.

\paragraph{Impact of MCTS Iterations:} Figure \ref{fig:parameter_exp}(b) displays the success rate as a function of maximum iterations (with $N=5$ and chunking=1). The results show a steep performance increase up to 20 iterations, after which the curve plateaus. Lower iteration counts fail to explore the search tree sufficiently for complex tasks, while 20 iterations provide a robust balance between search depth and inference time.
\vspace{-0.1cm}
\paragraph{Search Breadth vs. Trajectory Length:} Figure \ref{fig:parameter_exp}(c) highlights an inverse relationship between expansion width and trajectory depth. As the expansion factor increases, the average depth of successful trajectories decreases. This indicates that a broader search allows the agent to discover more efficient shortcuts and optimal paths, rather than relying on sub-optimal, longer sequences.
\vspace{-0.1cm}
\paragraph{Action Chunking for Long-Horizon Tasks:} In Figure \ref{fig:parameter_exp}(d), we evaluate action chunking on long-horizon domains (Chrome and Multi-apps). While increasing the chunk size to 5 improves performance by allowing the agent to plan macro-actions, increasing it further to 7 causes a performance drop. This degradation is likely due to the loss of granular control and the increased difficulty in error recovery when too many actions are committed simultaneously without intermediate feedback.

Across all experiments, \textbf{GPT-5.2} consistently achieves the highest success rates, outperforming both Gemini-3-flash and GPT-5-mini. We categorize the remaining failure cases into three primary causes: (1) context fragmentation and 
memory loss during long horizons, (2) environment state reconstruction errors, and (3) domain-specific knowledge gaps. Future work will focus on enhancing long-term memory mechanisms to mitigate these in extended tasks.

\subsection{Ablation Study}

To investigate the contribution of each component in Agent Alpha, we conduct an ablation study as shown in Table \ref{tab:ablation}. We verify the effectiveness of our design choices by replacing or removing specific modules from the full model, while using the base model of GPT-5-mini, fixing the maximum MCTS iteration to 20, expansion factor N=5, and action chunking =1.

\textbf{Impact of Judge Mechanism.} We first examine the effectiveness of the Comparative Judge. As shown in the results, replacing the Comparative Judge with an Independent Judge leads to a performance drop from 64.27\% to 57.96\%. This indicates that comparative reasoning/evaluation is crucial to accurately distinguish between superior and inferior actions, whereas independent evaluation struggles to capture the relative quality of candidate steps in complex computer-use environments.

\textbf{Impact of Backup Mechanism.} We then analyze the value back-propagation strategy. Replacing the \textit{Max} backup with a \textit{Mean} backup results in a significant performance degradation of 18.85\% (dropping to 45.42\%). 
This sharp decline suggests that for open-ended OS tasks, an "optimistic" update strategy is essential. 
Since the goal is to find \textit{one} successful trajectory, the Max backup correctly prioritizes high-potential paths, without diluting the signal by sub-optimal siblings as in Mean backup.

\textbf{Efficiency of Parallelism.} Finally, we evaluate the efficiency gains from our parallel computing strategies. We observe that Action Parallelism contributes most significantly to efficiency, providing a 4.4x speedup, while Environment Parallelism yields a 2.1x speedup. The full Agent Alpha integrates both strategies to achieve maximizing inference efficiency without compromising the search performance while maintaining the scalability in running time.

\section{Conclusion}
In this paper, we presented Agent Alpha, a MCTS-based framework tailored for general-purpose computer-use tasks, designed to address the efficiency bottlenecks of coarse-grained trajectory sampling in existing test-time scaling methods. We developed Alpha-UCT that leverages maximum values on search-paths to augment exploration and takes into account dependent samples. Through step-level regressive planning, Agent Alpha efficient error recovery and prefix reuse. We anlayze the regret bound of Agent Alpha. Empirical results on the OSWorld benchmark demonstrate that Agent Alpha not only significantly outperforms state-of-the-art baselines in success rate but also exhibits strong scalability and efficiency of test-time inference.

\bibliography{ref}      

@article{fang2024learning,
  title={Learning from random demonstrations: Offline reinforcement learning with importance-sampled diffusion models},
  author={Fang, Zeyu and Lan, Tian},
  journal={arXiv preprint arXiv:2405.19878},
  year={2024}
}

@article{zhang2024modeling,
  title={Modeling other players with bayesian beliefs for games with incomplete information},
  author={Zhang, Zuyuan and Imani, Mahdi and Lan, Tian},
  journal={arXiv preprint arXiv:2405.14122},
  year={2024}
}

@article{ma2025rethinking,
  title={Rethinking Testing for LLM Applications: Characteristics, Challenges, and a Lightweight Interaction Protocol},
  author={Ma, Wei and Yang, Yixiao and Hu, Qiang and Ying, Shi and Jin, Zhi and Du, Bo and Xing, Zhenchang and Li, Tianlin and Shi, Junjie and Liu, Yang and others},
  journal={arXiv preprint arXiv:2508.20737},
  year={2025}
}

@article{wang2025assessing,
  title={Assessing consistency and reproducibility in the outputs of large language models: Evidence across diverse finance and accounting tasks},
  author={Wang, Julian Junyan and Wang, Victor Xiaoqi},
  journal={arXiv preprint arXiv:2503.16974},
  year={2025}
}

@inproceedings{fang2023implementing,
  title={Implementing first-person shooter game ai in wild-scav with rule-enhanced deep reinforcement learning},
  author={Fang, Zeyu and Zhao, Jian and Zhou, Wengang and Li, Houqiang},
  booktitle={2023 IEEE Conference on Games (CoG)},
  pages={1--8},
  year={2023},
  organization={IEEE}
}

@article{alphazero,
  title={Mastering chess and shogi by self-play with a general reinforcement learning algorithm},
  author={Silver, David and Hubert, Thomas and Schrittwieser, Julian and Antonoglou, Ioannis and Lai, Matthew and Guez, Arthur and Lanctot, Marc and Sifre, Laurent and Kumaran, Dharshan and Graepel, Thore and others},
  journal={arXiv preprint arXiv:1712.01815},
  year={2017}
}

@article{zhang2025tail,
  title={Tail-Risk-Safe Monte Carlo Tree Search under PAC-Level Guarantees},
  author={Zhang, Zuyuan and Ghosh, Arnob and Lan, Tian},
  journal={arXiv preprint arXiv:2508.05441},
  year={2025}
}

@article{tang2025human,
  title={Human-centric service offloading with cnn partitioning in cloud-edge computing-empowered metaverse networks},
  author={Tang, Sizhe and Xia, Xiaoyu and Bilal, Muhammad and Dou, Wanchun and Xu, Xiaolong},
  journal={IEEE Transactions on Consumer Electronics},
  year={2025},
  publisher={IEEE}
}

@article{fang2024coordinate,
  title={Coordinate-aligned multi-camera collaboration for active multi-object tracking},
  author={Fang, Zeyu and Zhao, Jian and Yang, Mingyu and Lu, Zhenbo and Zhou, Wengang and Li, Houqiang},
  journal={Multimedia Systems},
  volume={30},
  number={4},
  pages={221},
  year={2024},
  publisher={Springer}
}

@article{song2024trial,
  title={Trial and Error: Exploration-Based Trajectory Optimization for LLM Agents},
  author={Song, Chan Hee and Wu, Jiaman and Washington, Clayton and Sadler, Brian M and Chao, Wei-Lun and Su, Yu},
  journal={arXiv preprint arXiv:2403.02502},
  year={2024}
}

@article{wang2024executable,
  title={Executable Code Actions Elicit Better LLM Agents},
  author={Wang, Xingyao and Yang, Hao and Hassan, Hany and Awadallah, Ahmed},
  journal={arXiv preprint arXiv:2402.01030},
  year={2024}
}

@inproceedings{xie2024osworld,
  title={OSWorld: Benchmarking Multimodal Agents for Open-Ended Tasks in Real Computer Environments},
  author={Xie, Tianbao and Zhang, Danyang and Chen, Jixuan and Li, François and Zhao, Siheng and Ruan, Kunsheng and Wang, Bo and Nowak, Sarah and O'Connell, Lisa and Agrawal, Maneesh and others},
  booktitle={Advances in Neural Information Processing Systems (NeurIPS)},
  year={2024}
}

@article{zhang2025lipschitz,
  title={Lipschitz Lifelong Monte Carlo Tree Search for Mastering Non-Stationary Tasks},
  author={Zhang, Zuyuan and Lan, Tian},
  journal={arXiv preprint arXiv:2502.00633},
  year={2025}
}

@article{rawles2024androidworld,
  title={Androidworld: A dynamic benchmarking environment for autonomous agents},
  author={Rawles, Christopher and Clinckemaillie, Sarah and Chang, Yifan and Waltz, Jonathan and Lau, Gabrielle and Fair, Marybeth and Li, Alice and Bishop, William and Li, Wei and Campbell-Ajala, Folawiyo and others},
  journal={arXiv preprint arXiv:2405.14573},
  year={2024}
}

@article{niu2024screenagent,
  title={Screenagent: A vision language model-driven computer control agent},
  author={Niu, Runliang and Li, Jindong and Wang, Shiqi and Fu, Yali and Hu, Xiyu and Leng, Xueyuan and Kong, He and Chang, Yi and Wang, Qi},
  journal={arXiv preprint arXiv:2402.07945},
  year={2024}
}

@inproceedings{li2024crowdsensing,
  title={A Crowdsensing Service Pricing Method in Vehicular Edge Computing},
  author={Li, Zheng and Tang, Sizhe and Tian, Hao and Xiang, Haolong and Xu, Xiaolong and Dou, Wanchun},
  booktitle={2024 IEEE International Symposium on Parallel and Distributed Processing with Applications (ISPA)},
  pages={82--89},
  year={2024},
  organization={IEEE}
}

@article{snell2024scaling,
  title={Scaling LLM Test-Time Compute Optimally can be More Effective than Scaling Model Parameters},
  author={Snell, Charlie and et al.},
  journal={arXiv preprint arXiv:2408.03314},
  year={2024}
}

@article{jiang2022intelligent,
  title={Intelligent monitoring for infectious diseases with fuzzy systems and edge computing: A survey},
  author={Jiang, Qinting and Zhou, Xuanhong and Wang, Ruili and Ding, Weiping and Chu, Yi and Tang, Sizhe and Jia, Xiaoyun and Xu, Xiaolong},
  journal={Applied Soft Computing},
  volume={123},
  pages={108835},
  year={2022},
  publisher={Elsevier}
}

@inproceedings{lightman2023verify,
  title={Let's Verify Step by Step},
  author={Lightman, Hunter and Kosaraju, Vineet and Burda, Yura and Edwards, Harri and Baker, Bowen and Lee, Teddy and Leike, Jan and Schulman, John and Sutskever, Ilya and Cobbe, Karl},
  booktitle={International Conference on Learning Representations (ICLR)},
  year={2024}
}

@inproceedings{peng2024grounding,
  title={Grounding multimodal large language models to the world},
  author={Peng, Zhiliang and Wang, Wenhui and Dong, Li and Hao, Yaru and Huang, Shaohan and Ma, Shuming and Ye, Qixiang and Wei, Furu},
  booktitle={The Twelfth International Conference on Learning Representations},
  year={2024}
}

@article{qin2025ui,
  title={Ui-tars: Pioneering automated gui interaction with native agents},
  author={Qin, Yujia and Ye, Yining and Fang, Junjie and Wang, Haoming and Liang, Shihao and Tian, Shizuo and Zhang, Junda and Li, Jiahao and Li, Yunxin and Huang, Shijue and others},
  journal={arXiv preprint arXiv:2501.12326},
  year={2025}
}

@inproceedings{chang2024mixed,
  title={Mixed text recognition with efficient parameter fine-tuning and transformer},
  author={Chang, Da and Li, Yu},
  booktitle={International Conference on Neural Information Processing},
  pages={17--31},
  year={2024},
  organization={Springer}
}

@inproceedings{yao2022react,
  title={React: Synergizing reasoning and acting in language models},
  author={Yao, Shunyu and Zhao, Jeffrey and Yu, Dian and Du, Nan and Shafran, Izhak and Narasimhan, Karthik R and Cao, Yuan},
  booktitle={The eleventh international conference on learning representations},
  year={2022}
}

@article{wei2022chain,
  title={Chain-of-thought prompting elicits reasoning in large language models},
  author={Wei, Jason and Wang, Xuezhi and Schuurmans, Dale and Bosma, Maarten and Xia, Fei and Chi, Ed and Le, Quoc V and Zhou, Denny and others},
  journal={Advances in neural information processing systems},
  volume={35},
  pages={24824--24837},
  year={2022}
}

@article{li2025inspo,
  title={Inspo: Unlocking intrinsic self-reflection for llm preference optimization},
  author={Li, Yu and Lan, Tian and Qi, Zhengling},
  journal={arXiv preprint arXiv:2512.23126},
  year={2025}
}

@article{yao2023tree,
  title={Tree of thoughts: Deliberate problem solving with large language models},
  author={Yao, Shunyu and Yu, Dian and Zhao, Jeffrey and Shafran, Izhak and Griffiths, Tom and Cao, Yuan and Narasimhan, Karthik},
  journal={Advances in neural information processing systems},
  volume={36},
  pages={11809--11822},
  year={2023}
}

@article{song2025coact,
  title={Coact-1: Computer-using agents with coding as actions},
  author={Song, Linxin and Dai, Yutong and Prabhu, Viraj and Zhang, Jieyu and Shi, Taiwei and Li, Li and Li, Junnan and Savarese, Silvio and Chen, Zeyuan and Zhao, Jieyu and others},
  journal={arXiv preprint arXiv:2508.03923},
  year={2025}
}

@article{yang2025gta1,
  title={Gta1: Gui test-time scaling agent},
  author={Yang, Yan and Li, Dongxu and Dai, Yutong and Yang, Yuhao and Luo, Ziyang and Zhao, Zirui and Hu, Zhiyuan and Huang, Junzhe and Saha, Amrita and Chen, Zeyuan and others},
  journal={arXiv preprint arXiv:2507.05791},
  year={2025}
}

@article{zhu2025scaling,
  title={Scaling Test-time Compute for LLM Agents},
  author={Zhu, King and Li, Hanhao and Wu, Siwei and Xing, Tianshun and Ma, Dehua and Tang, Xiangru and Liu, Minghao and Yang, Jian and Liu, Jiaheng and Jiang, Yuchen Eleanor and others},
  journal={arXiv preprint arXiv:2506.12928},
  year={2025}
}

@inproceedings{muennighoff2025s1,
  title={s1: Simple test-time scaling},
  author={Muennighoff, Niklas and Yang, Zitong and Shi, Weijia and Li, Xiang Lisa and Fei-Fei, Li and Hajishirzi, Hannaneh and Zettlemoyer, Luke and Liang, Percy and Cand{\`e}s, Emmanuel and Hashimoto, Tatsunori B},
  booktitle={Proceedings of the 2025 Conference on Empirical Methods in Natural Language Processing},
  pages={20286--20332},
  year={2025}
}

@article{tang2025malinzero,
  title={Malinzero: Efficient low-dimensional search for mastering complex multi-agent planning},
  author={Tang, Sizhe and Chen, Jiayu and Lan, Tian},
  journal={arXiv preprint arXiv:2511.06142},
  year={2025}
}

@inproceedings{uct,
  title={Bandit based monte-carlo planning},
  author={Kocsis, Levente and Szepesv{\'a}ri, Csaba},
  booktitle={European conference on machine learning},
  pages={282--293},
  year={2006},
  organization={Springer}
}

@inproceedings{yu2025look,
  title={Look-ahead robust network optimization with generative state predictions},
  author={Yu, Fei Xu and Zhang, Zuyuan and Grob, Emily and Adam, Gina and Coffey, Sean and Bastian, Nathaniel D and Lan, Tian},
  booktitle={AAAI 2025 Workshop on Artificial Intelligence for Wireless Communications and Networking (AI4WCN)}
}

@article{freedman1975tail,
  title={On tail probabilities for martingales},
  author={Freedman, David A},
  journal={the Annals of Probability},
  pages={100--118},
  year={1975},
  publisher={JSTOR}
}

@article{kato2020concentration,
  title={Concentration inequality using unconfirmed knowledge},
  author={Kato, Go},
  journal={arXiv preprint arXiv:2002.04357},
  year={2020}
}

@article{gonzalez2025unreasonable,
  title={The unreasonable effectiveness of scaling agents for computer use},
  author={Gonzalez-Pumariega, Gonzalo and Tu, Vincent and Lee, Chih-Lun and Yang, Jiachen and Li, Ang and Wang, Xin Eric},
  journal={arXiv preprint arXiv:2510.02250},
  year={2025}
}

@article{gu2024survey,
  title={A survey on llm-as-a-judge},
  author={Gu, Jiawei and Jiang, Xuhui and Shi, Zhichao and Tan, Hexiang and Zhai, Xuehao and Xu, Chengjin and Li, Wei and Shen, Yinghan and Ma, Shengjie and Liu, Honghao and others},
  journal={The Innovation},
  year={2024},
  publisher={Elsevier}
}

@article{lorentz2016using,
  title={Using evaluation functions in Monte-Carlo tree search},
  author={Lorentz, Richard},
  journal={Theoretical computer science},
  volume={644},
  pages={106--113},
  year={2016},
  publisher={Elsevier}
}

@article{yang2026symphony,
  title={OS-Symphony: A Holistic Framework for Robust and Generalist Computer-Using Agent},
  author={Yang, Bowen and Jin, Kaiming and Wu, Zhenyu and Liu, Zhaoyang and Sun, Qiushi and Li, Zehao and Xie, JingJing and Liu, Zhoumianze and Xu, Fangzhi and Cheng, Kanzhi and others},
  journal={arXiv preprint arXiv:2601.07779},
  year={2026}
}

@article{li2026acdzero,
  title={ACDZero: Graph-Embedding-Based Tree Search for Mastering Automated Cyber Defense},
  author={Li, Yu and Tang, Sizhe and Chen, Rongqian and Yu, Fei Xu and Jiang, Guangyu and Imani, Mahdi and Bastian, Nathaniel D and Lan, Tian},
  journal={arXiv preprint arXiv:2601.02196},
  year={2026}
}

@article{chen2025neurosymbolic,
  title={A neurosymbolic framework for interpretable cognitive attack detection in augmented reality},
  author={Chen, Rongqian and Andreyev, Allison and Xiu, Yanming and Chilukuri, Joshua and Sen, Shunav and Imani, Mahdi and Li, Bin and Gorlatova, Maria and Tan, Gang and Lan, Tian},
  journal={arXiv preprint arXiv:2508.09185},
  year={2025}
}

@inproceedings{chen2025perception,
  title={Perception graph for cognitive attack reasoning in augmented reality},
  author={Chen, Rongqian and Hong, Shu and Islam, Rifatul and Imani, Mahdi and Tan, Gang and Lan, Tian},
  booktitle={Proceedings of the Twenty-sixth International Symposium on Theory, Algorithmic Foundations, and Protocol Design for Mobile Networks and Mobile Computing},
  pages={505--506},
  year={2025}
}

@article{yu2025optimizing,
  title={Optimizing prompt sequences using monte carlo tree search for llm-based optimization},
  author={Yu, Fei Xu and Adam, Gina and Bastian, Nathaniel D and Lan, Tian},
  journal={arXiv preprint arXiv:2508.05995},
  year={2025}
}
\bibliographystyle{unsrtnat}

\newpage
\clearpage
\onecolumn
\appendix

\section{Theoretical Proofs}
\label{app:proofs}

In this appendix, we provide the detailed derivation of the regret bound for Agent Alpha (Theorem \ref{thm:regret_bound_kato}). Our analysis relies on the concentration inequalities for martingales, explicitly leveraging the \textit{unconfirmed knowledge} framework introduced by \citet{kato2020concentration} to quantify the variance reduction effect of the reflection mechanism.

\textbf{Alignment with Methodology:} As discussed in Section \ref{sec:theoretical_analysis}, while Agent Alpha utilizes a robust max-UCB selection criterion (Eq. \ref{eq:alpha_uct_max}) for engineering stability, the core efficiency gain stems from the reflection and evaluation mechanism reducing the effective variance of the value estimation. The following proof quantifies this information-theoretic advantage.

\subsection{Preliminaries and Definitions}

Let $\mathcal{K} = \{1, \dots, K\}$ be the set of actions (arms) at a given expansion step. We denote the true expected value of action $a$ as $\mu_a = \mathbb{E}[X \mid a]$. Let $a^* = \arg\max_{a \in \mathcal{K}} \mu_a$ be the optimal action with mean $\mu^*$, and for any suboptimal action $a \neq a^*$, let $\Delta_a = \mu^* - \mu_a$ be the optimality gap.

We assume the evaluation scores $X_t \in [0, 1]$. Agent Alpha generates a reflection prior $\hat{\theta}_t$ for each step, serving as a predictor for $X_t$. The uncertainty of the estimation is governed by the conditional residual variance:
\begin{equation}
    \sigma_{\text{res}, a}^2 = \mathbb{E} \left[ (X_t - \hat{\theta}_t)^2 \mid a_t = a, \mathcal{F}_{t-1} \right].
\end{equation}

\subsection{Proof of Theorem \ref{thm:regret_bound_kato}}

The proof proceeds in three steps: establishing the concentration bound via Freedman's Inequality, deriving the sample complexity for suboptimal arms, and summing the regret.

\subsubsection{Step 1: Concentration with Residual Variance}

Let $S_{n, a} = \sum_{k=1}^n (X_{k, a} - \mu_a)$ be the sum of martingale differences for arm $a$ after $n$ visits. Note that $|X_{k,a} - \mu_a| \le 1$.
Let $V_{n, a}$ be the cumulative conditional variance. Following \citet{kato2020concentration}, the predictable quadratic variation process is bounded by the residual variance due to the presence of the predictor $\hat{\theta}$:
\begin{equation}
    V_{n, a} = \sum_{k=1}^n \text{Var}(X_{k,a} \mid \mathcal{F}_{k-1}) \le \sum_{k=1}^n \mathbb{E}[(X_{k, a} - \hat{\theta}_{k, a})^2 \mid \mathcal{F}_{k-1}] \approx n \sigma_{\text{res}, a}^2.
\end{equation}

\textbf{Lemma 1 (Freedman's Inequality \cite{freedman1975tail}).} 
For any real numbers $\epsilon > 0$ and $v > 0$,
\begin{equation}
    \mathbb{P} \left( S_{n,a} \ge \epsilon \cap V_{n,a} \le v \right) \le \exp \left( - \frac{\epsilon^2}{2v + 2\epsilon/3} \right).
\end{equation}

We seek a confidence width $C_{n,a}$ such that the probability of deviation is bounded by a small $\delta$. Setting the RHS to $\delta$:
\begin{equation}
    \exp \left( - \frac{\epsilon^2}{2v + 2\epsilon/3} \right) = \delta \implies \frac{\epsilon^2}{2v + 2\epsilon/3} = \ln(1/\delta).
\end{equation}

Rearranging terms leads to a quadratic inequality in $\epsilon$:
\begin{equation}
    \epsilon^2 - \frac{2}{3}\ln(1/\delta)\epsilon - 2v \ln(1/\delta) = 0.
\end{equation}

Solving for the positive root $\epsilon$ using the quadratic formula:
\begin{equation}
    \epsilon = \frac{\frac{2}{3}\ln(1/\delta) + \sqrt{\frac{4}{9}(\ln(1/\delta))^2 + 8v \ln(1/\delta)}}{2} = \frac{\ln(1/\delta)}{3} + \sqrt{\frac{(\ln(1/\delta))^2}{9} + 2v \ln(1/\delta)}.
\end{equation}
Using the inequality $\sqrt{A+B} \le \sqrt{A} + \sqrt{B}$, we bound $\epsilon$:
\begin{equation}
    \epsilon \le \frac{\ln(1/\delta)}{3} + \frac{\ln(1/\delta)}{3} + \sqrt{2v \ln(1/\delta)} = \sqrt{2v \ln(1/\delta)} + \frac{2}{3}\ln(1/\delta).
\end{equation}

Substituting $v \approx n \sigma_{\text{res}, a}^2$ and dividing by $n$ to convert the sum $S_{n,a}$ to the mean $\bar{X}_{n,a}$, we obtain the confidence radius $C_{n, a}(\delta) = \epsilon / n$:
\begin{equation}
\label{eq:confidence_radius_derived}
    C_{n, a}(\delta) = \sqrt{\frac{2 \sigma_{\text{res}, a}^2 \ln(1/\delta)}{n}} + \frac{2 \ln(1/\delta)}{3n}.
\end{equation}
By setting $\delta = T^{-2}$, we establish that with high probability, $|\bar{X}_{n, a} - \mu_a| \le C_{n,a}(T^{-2})$.

\subsubsection{Step 2: Bounding the Number of Suboptimal Selections}

Consider a suboptimal action $a$ with optimality gap $\Delta_a > 0$. The algorithm selects action $a$ at time $t$ only if its Upper Confidence Bound exceeds that of the optimal action $a^*$. This implies:
\begin{equation}
    \bar{X}_{T_a(t), a} + C_{T_a(t), a} \ge \bar{X}_{T_{a^*}(t), a^*} + C_{T_{a^*}(t), a^*}.
\end{equation}
This condition holds only if at least one of the following three failure events occurs (standard UCT decomposition \cite{uct}):
\begin{align}
    E_1: & \quad \bar{X}_{T_{a^*}(t), a^*} \le \mu^* - C_{T_{a^*}(t), a^*} \quad (\text{Optimal arm under-estimated}) \\
    E_2: & \quad \bar{X}_{T_a(t), a} \ge \mu_a + C_{T_a(t), a} \quad (\text{Suboptimal arm over-estimated}) \\
    E_3: & \quad \mu^* < \mu_a + 2 C_{T_a(t), a} \quad (\text{True means are close relative to confidence width})
\end{align}
The probabilities of $E_1$ and $E_2$ are bounded by $\delta = T^{-2}$ via Step 1. The dominant contribution to regret comes from $E_3$, which determines the necessary samples $N_a$ to resolve the gap. $E_3$ implies:
\begin{equation}
    \Delta_a < 2 C_{N_a, a} = 2 \sqrt{\frac{2 \sigma_{\text{res}, a}^2 \ln T}{N_a}} + \frac{4 \ln T}{3 N_a}.
\end{equation}
Solving for $\sqrt{N_a}$, we treat this as a quadratic inequality $\Delta_a (\sqrt{N_a})^2 - \alpha \sqrt{N_a} - \beta < 0$. For asymptotic behavior ($T \to \infty$), the first order term dominates:
\begin{equation}
    \sqrt{N_a} < \frac{2 \sqrt{2 \sigma_{\text{res}, a}^2 \ln T}}{\Delta_a}.
\end{equation}
Squaring provides the bound on the number of pulls:
\begin{equation}
\label{eq:num_draws}
    N_a(T) \le \frac{8 \sigma_{\text{res}, a}^2 \ln T}{\Delta_a^2} + \Psi(T),
\end{equation}
where $\Psi(T)$ represents lower-order terms $O(\ln T / \Delta_a)$.

\subsubsection{Step 3: Cumulative Regret}

The cumulative regret $R_T$ is defined as the sum of gaps weighted by expected pulls:
\begin{equation}
    R_T = \sum_{a \neq a^*} \Delta_a \mathbb{E}[N_a(T)].
\end{equation}

Substituting Eq. (\ref{eq:num_draws}):
\begin{equation}
    R_T \le \sum_{a \neq a^*} \Delta_a \left( \frac{8 \sigma_{\text{res}, a}^2 \ln T}{\Delta_a^2} + \frac{16 \ln T}{3 \Delta_a} + 2 \right).
\end{equation}
(Note: The constant 2 accounts for the finite summation of probabilities of $E_1, E_2$).
Simplifying, we obtain the final bound:
\begin{equation}
    R_T \le \sum_{a \neq a^*} \left( \frac{8 \sigma_{\text{res}, a}^2 \ln T}{\Delta_a} + \frac{16 \ln T}{3} + 2\Delta_a \right).
\end{equation}
\hfill \qed

\subsection{Proof of Corollary \ref{cor:complexity_order} (The Order of Regret Bound)}

To derive the asymptotic complexity order, we analyze the dominant terms in the regret bound established in Theorem \ref{thm:regret_bound_kato}.

Recall the bound:
\begin{equation}
    R_T \le \sum_{a \in \mathcal{K}, a \neq a^*} \left( \frac{8 \sigma_{\text{res}, a}^2 \ln T}{\Delta_a} + \frac{16 \ln T}{3} + 2\Delta_a \right).
\end{equation}

Let $K = |\mathcal{K}|$ be the total number of arms (branching factor). There are $K-1$ suboptimal arms in the summation.
We define the minimal optimality gap $\Delta_{\min} = \min_{a \neq a^*} \Delta_a > 0$ and the maximum expected residual variance $\sigma_{\max}^2 = \max_{a} \sigma_{\text{res}, a}^2$.

We proceed by bounding each term in the summation:

1.  \textbf{Dominant Term (Variance-dependent):}
    \begin{equation}
        \sum_{a \neq a^*} \frac{8 \sigma_{\text{res}, a}^2 \ln T}{\Delta_a} \le (K-1) \frac{8 \sigma_{\max}^2 \ln T}{\Delta_{\min}}.
    \end{equation}

2.  \textbf{Secondary Term (Bernstein Correction):}
    \begin{equation}
        \sum_{a \neq a^*} \frac{16 \ln T}{3} = (K-1) \frac{16 \ln T}{3}.
    \end{equation}

3.  \textbf{Constant Term:}
    \begin{equation}
        \sum_{a \neq a^*} 2\Delta_a \le 2(K-1) \Delta_{\max}, \quad \text{where } \Delta_{\max} \le 1.
    \end{equation}

Combining these, the regret is bounded by:
\begin{equation}
    R_T \le (K-1) \ln T \left( \frac{8 \sigma_{\max}^2}{\Delta_{\min}} + \frac{16}{3} \right) + 2(K-1).
\end{equation}

We treat the gap parameters $\Delta_{\min}$ as problem-dependent constants. We focus on the scaling with respect to the horizon $T$, the branching factor $K$, and the core variance characteristic $\sigma_{\text{res}}^2$ (represented here by $\sigma_{\max}^2$).

Dropping the lower-order constant terms and absorbing the numerical constants ($8/\Delta_{\min}, 16/3$) into the asymptotic notation, we obtain:
\begin{equation}
    R_T = \mathcal{O}\left( K \cdot \sigma_{\text{res}}^2 \cdot \ln T \right).
\end{equation}

This confirms that the regret grows logarithmically with time $T$, linearly with the action space size $K$, and linearly with the residual variance $\sigma_{\text{res}}^2$. Crucially, if $\sigma_{\text{res}}^2 \to 0$ (perfect reflection), the constant factor pre-multiplying $\ln T$ vanishes, significantly reducing the regret compared to standard UCT where $\sigma^2$ is a fixed constant.

\hfill \qed

\subsection{Proof of Corollary \ref{cor:efficiency_gain} (Efficiency Gain)}

We explicitly compare the asymptotic regret terms.
Let $R_T^{\text{UCT}}$ be the regret of standard UCT using Hoeffding's inequality, which assumes a variance proxy $\sigma_{\text{raw}}^2 = 1/4$ (bounded range $[0, 1]$):
\begin{equation}
    R_T^{\text{UCT}} \approx \sum_{a \neq a^*} \frac{8 \sigma_{\text{raw}}^2 \ln T}{\Delta_a}.
\end{equation}
Let $R_T^{\text{Alpha}}$ be the regret derived above using residual variance $\sigma_{\text{res}}^2$:
\begin{equation}
    R_T^{\text{Alpha}} \approx \sum_{a \neq a^*} \frac{8 \sigma_{\text{res}, a}^2 \ln T}{\Delta_a}.
\end{equation}
The efficiency ratio is:
\begin{equation}
    \eta = \frac{R_T^{\text{Alpha}}}{R_T^{\text{UCT}}} \approx \frac{\sigma_{\text{res}}^2}{\sigma_{\text{raw}}^2} = \frac{\mathbb{E}[(X - \hat{\theta})^2]}{\text{Var}(X)}.
\end{equation}
Since the reflection $\hat{\theta}$ correlates with $X$, $\sigma_{\text{res}}^2 < \sigma_{\text{raw}}^2$, proving $\eta < 1$.
\qed

\clearpage
\section{Case Studies}
\subsection{Task1: Can you make Bing the main search engine when I look stuff up on the internet? }

\begin{figure}[h!]
  \centering
  \resizebox{\linewidth}{!}{%
    \begin{forest}
      for tree={
        parent anchor=south,
        child anchor=north,
        l sep=1.2cm,    
        s sep=0.5cm,    
        align=center,
        font=\sffamily\large\bfseries, 
        edge={thick, -{Stealth[]}},    
        inner sep=3pt,
      }
      [
        {\includegraphics[width=7cm]{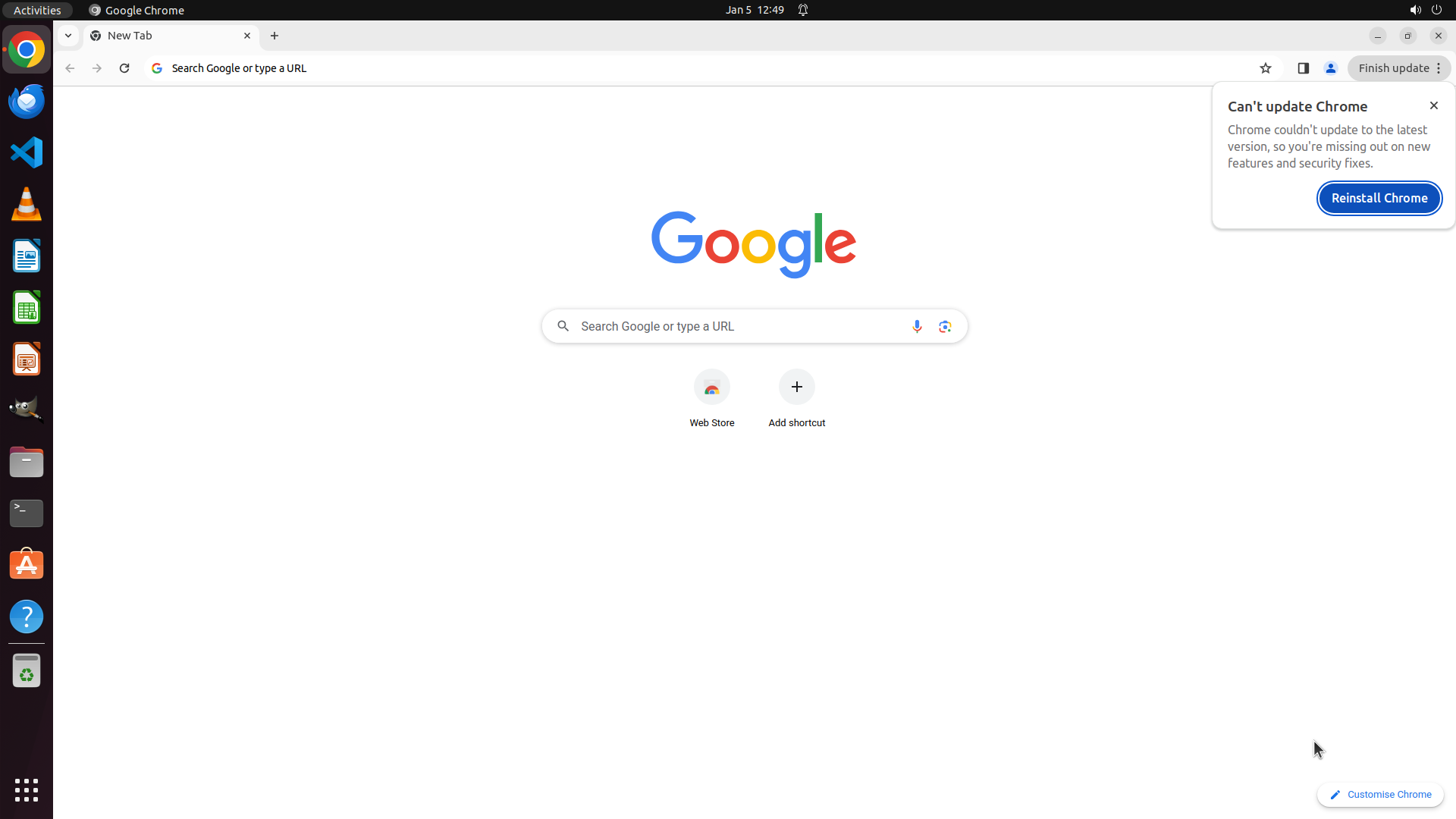}\\ \textbf{Root}}, name=root
        [
          {\includegraphics[width=7cm]{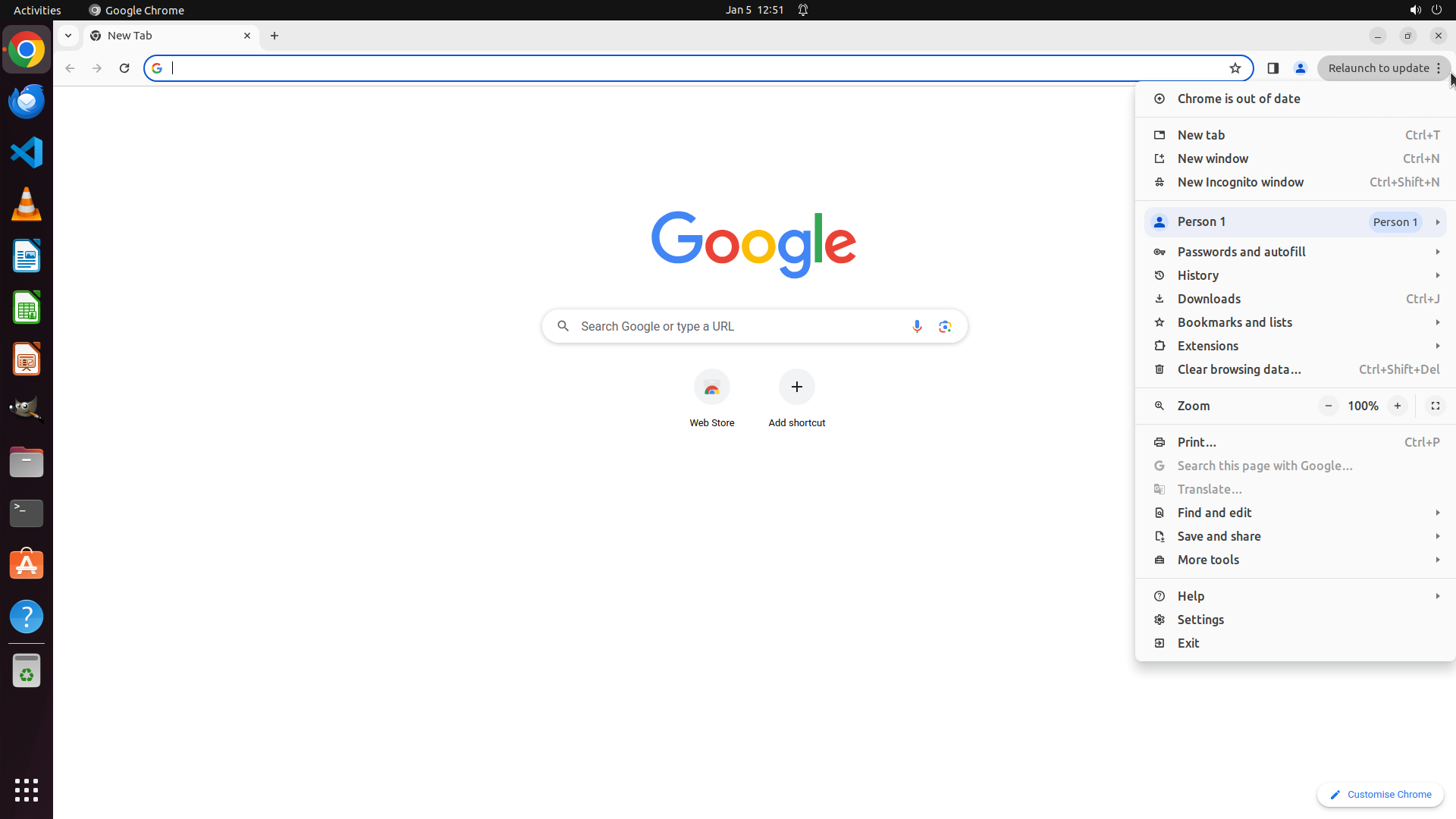}\\ \textbf{Node 1.1}\\ UCB: 0.729 \\ \texttt{\small Action: click(1908, 90)}}, name=n11
          [
            {\includegraphics[width=7cm]{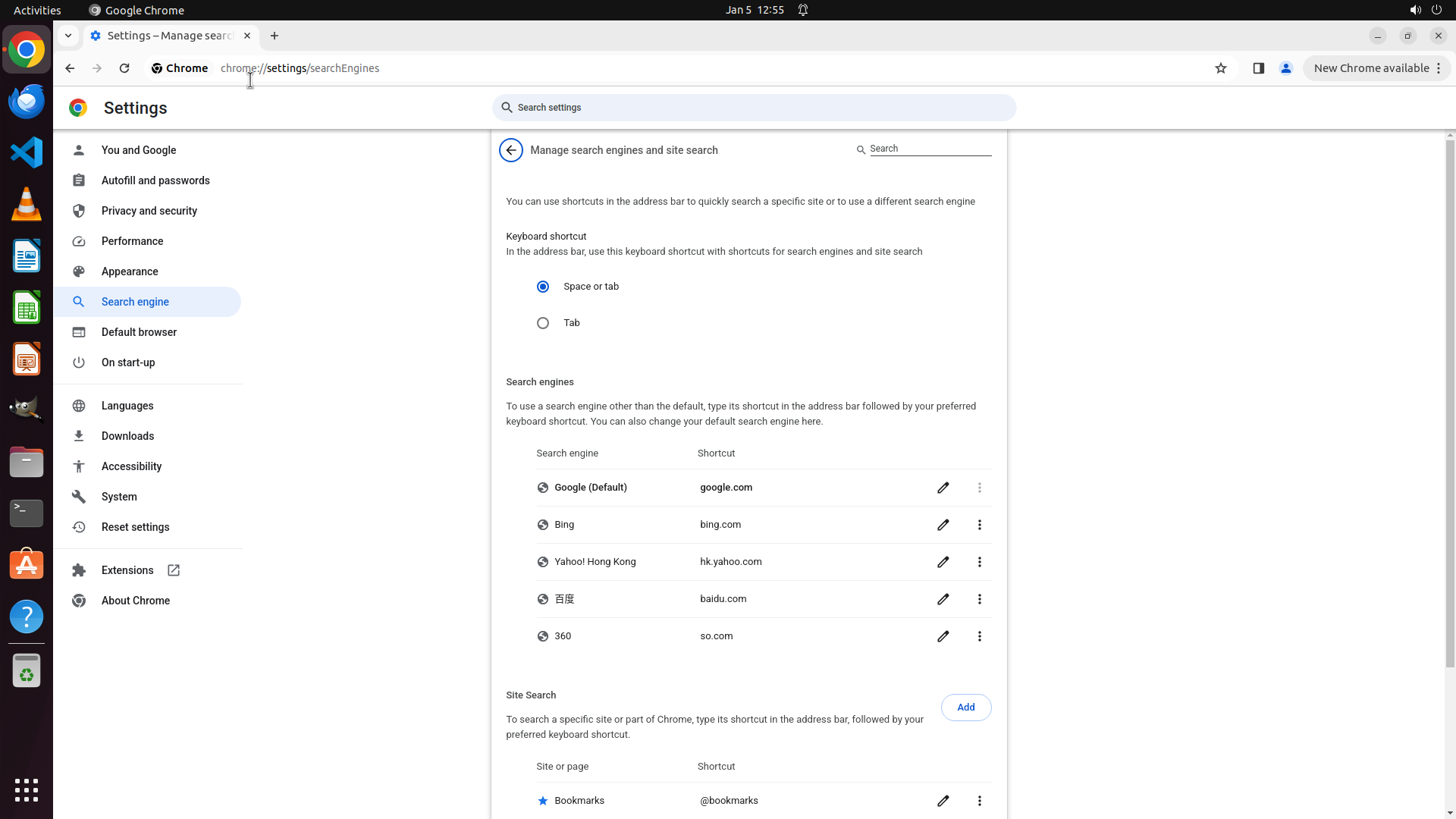}\\ \textbf{Node 1.1.3}\\ UCB: 0.731 \\ \texttt{\small Action: write('chrome://settings...')}}, name=n113
            [
              {\includegraphics[width=7cm]{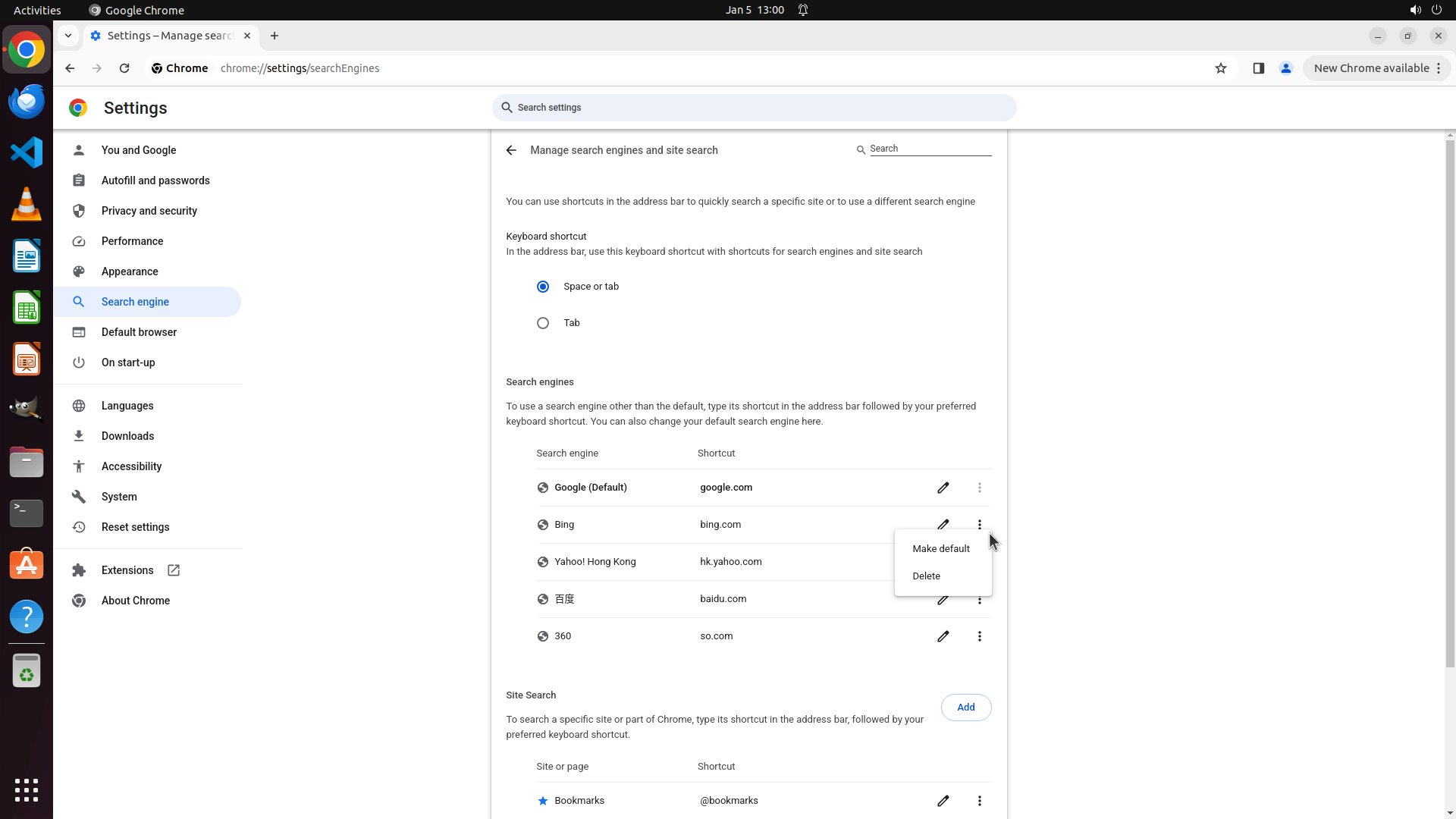}\\ \textbf{Node 1.1.3.1}\\ UCB: 0.539 \\ \texttt{\small Action: click(1300, 698)}}, name=n1131
              [{\includegraphics[width=7cm]{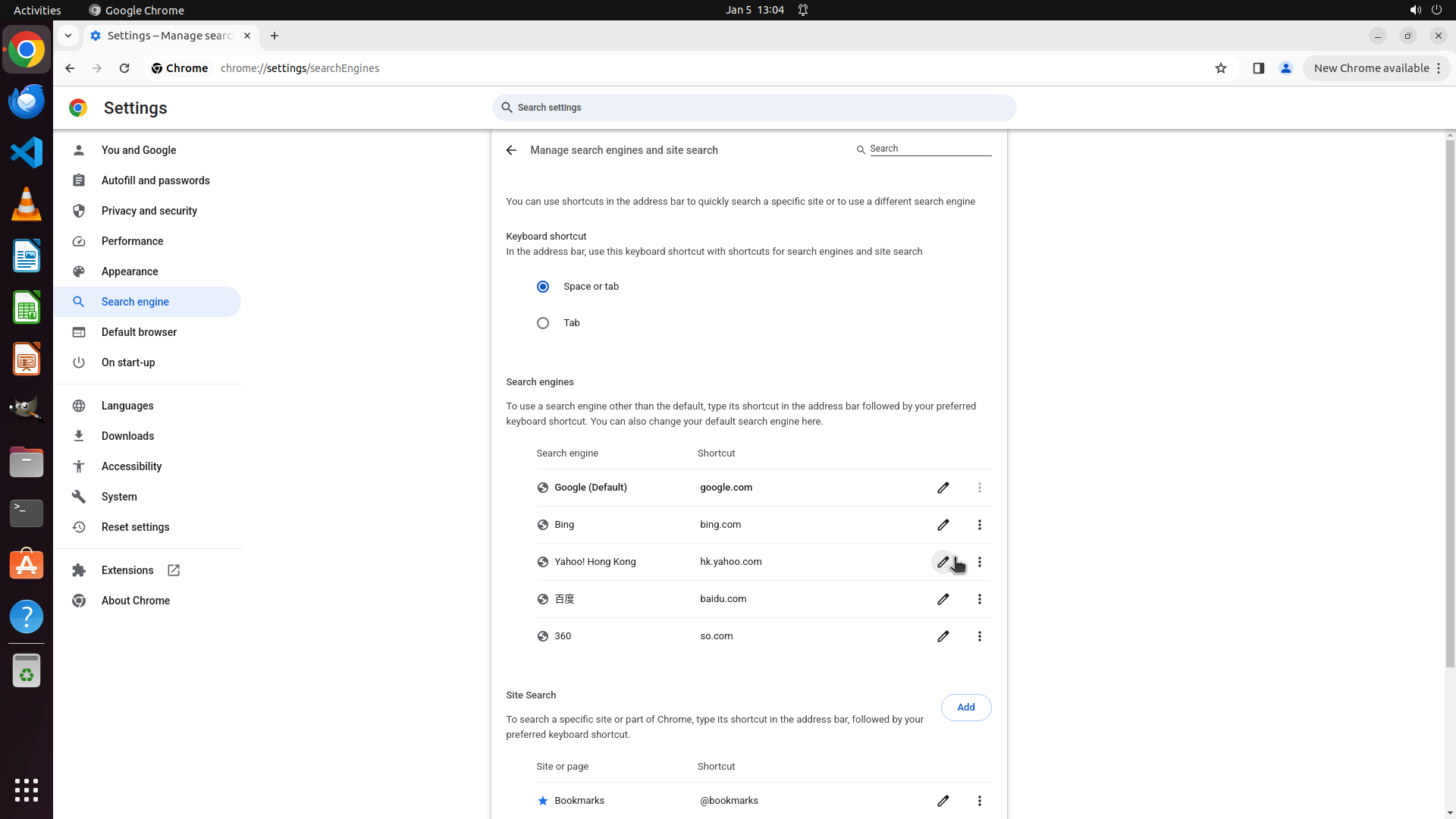}\\ \textbf{Node 1.1.3.1.1}\\ Evaluation: Success \\ \texttt{\small Action: click(1300, 750)}}]
            ]
            [
              {\includegraphics[width=7cm]{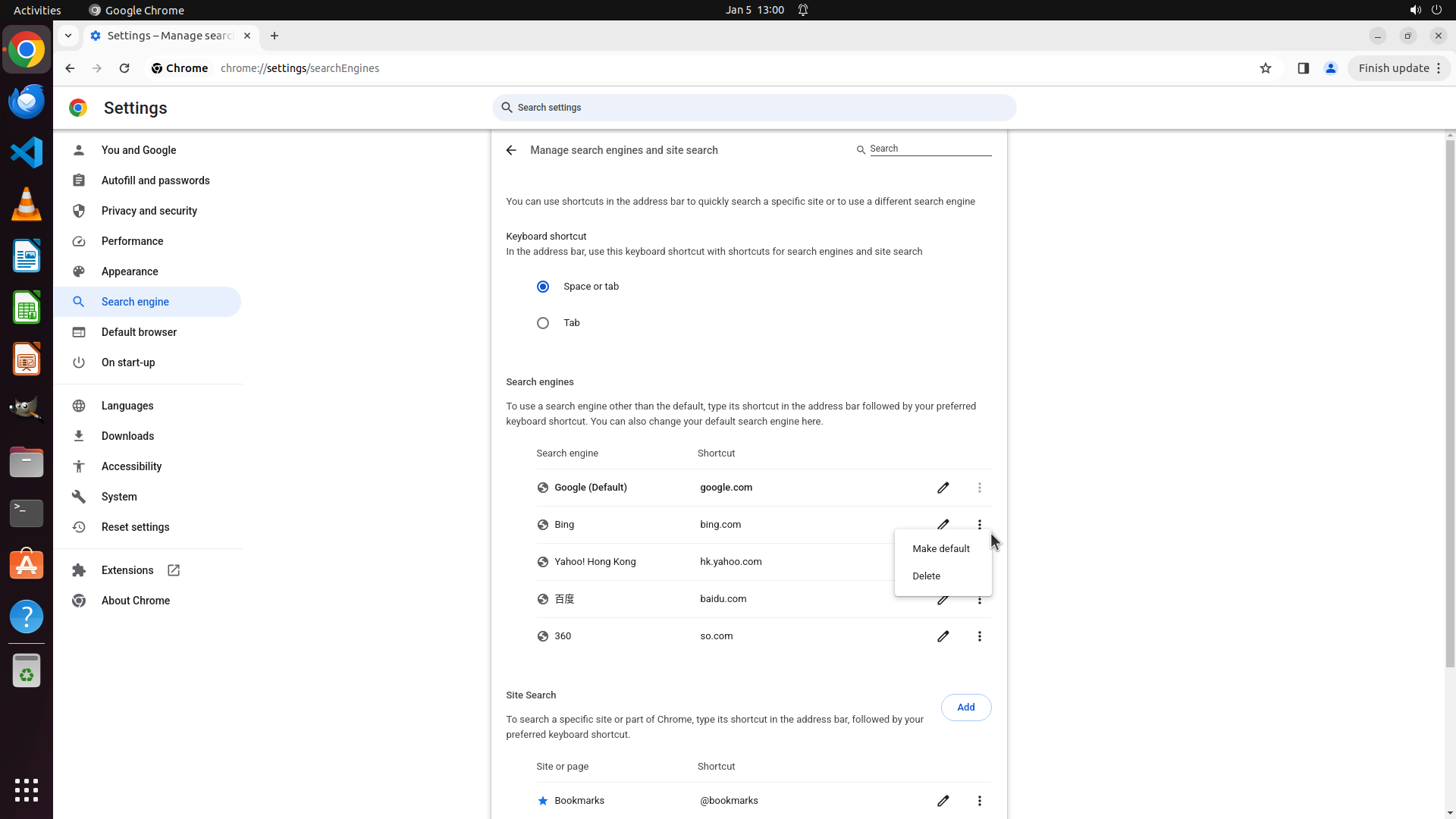}\\ \textbf{Node 1.1.3.2}\\ UCB: 0.491 \\ \texttt{\small Action: click(1302, 698)}}
            ]
          ]
        ]
        [
          {\includegraphics[width=7cm]{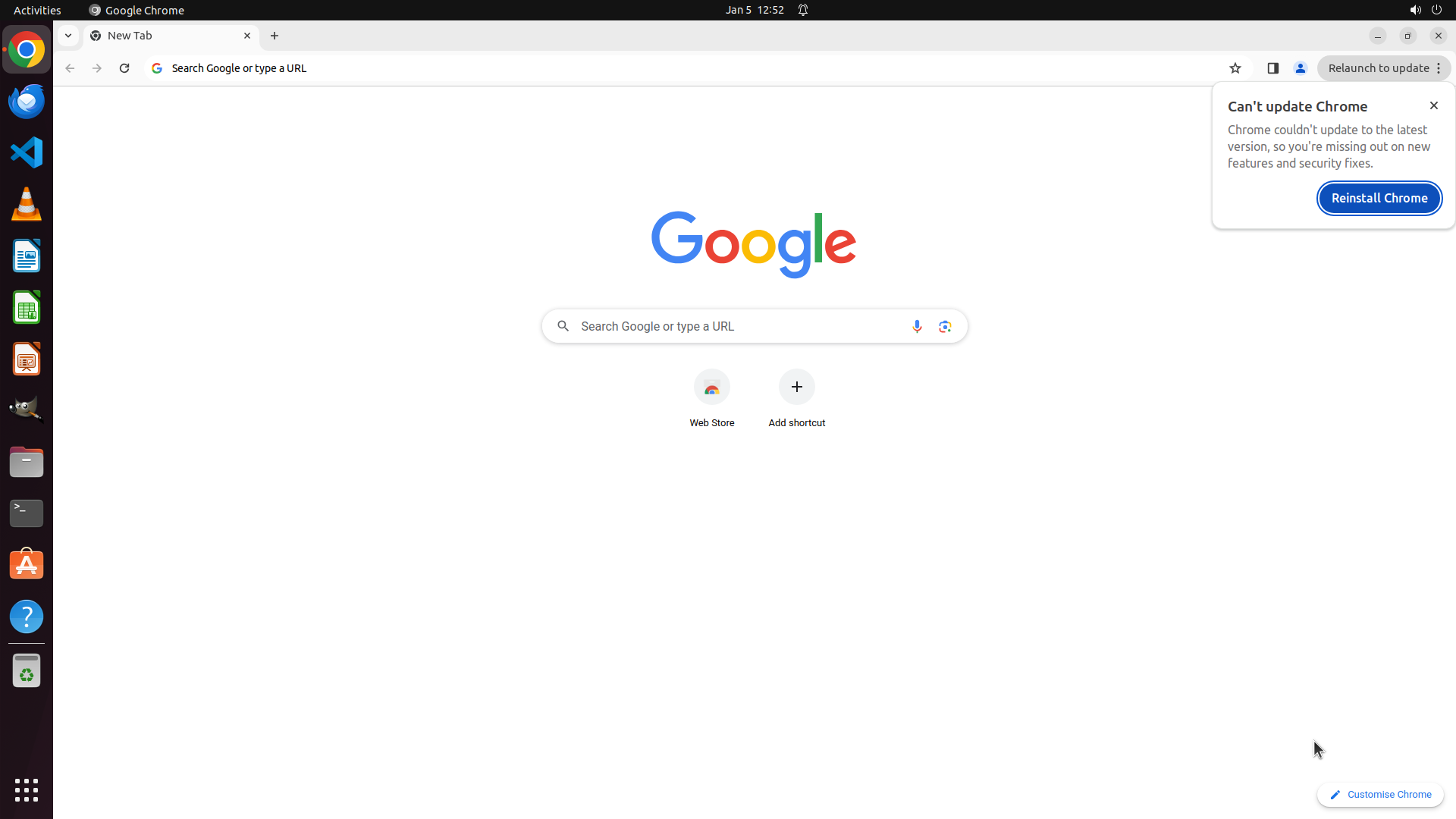}\\ \textbf{Node 1.2}\\ UCB: 0.131 \\ \texttt{\small Action: hotkey('ctrl', ',')}}, name=n12
          [
            {\includegraphics[width=7cm]{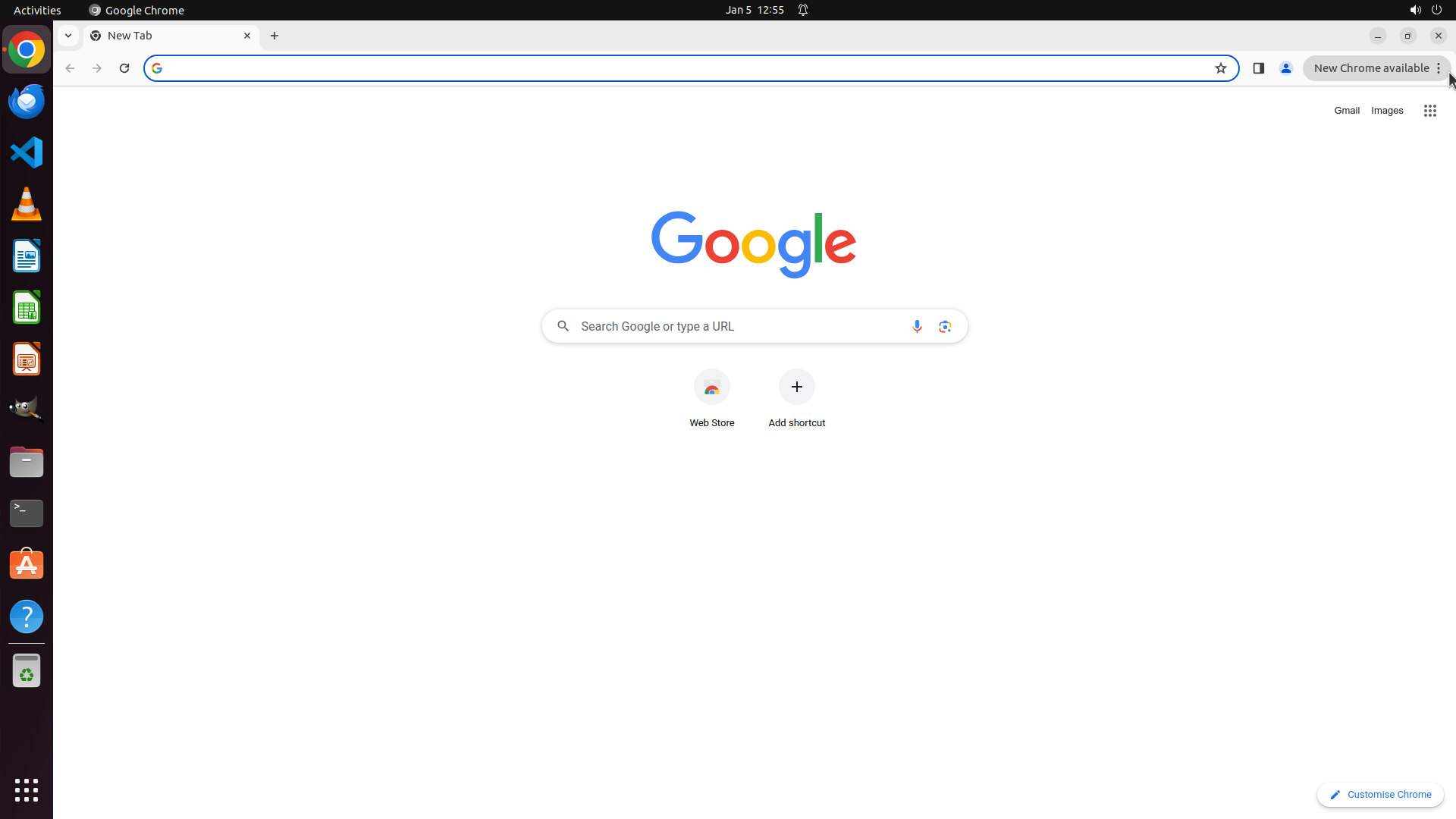}\\ \textbf{Node 1.1.1}\\ UCB: 0.101 \\ \texttt{\small Action: click(1906, 90)}}, name=n111, edge={draw=none}
          ]
        ]
        [
          {\includegraphics[width=7cm]{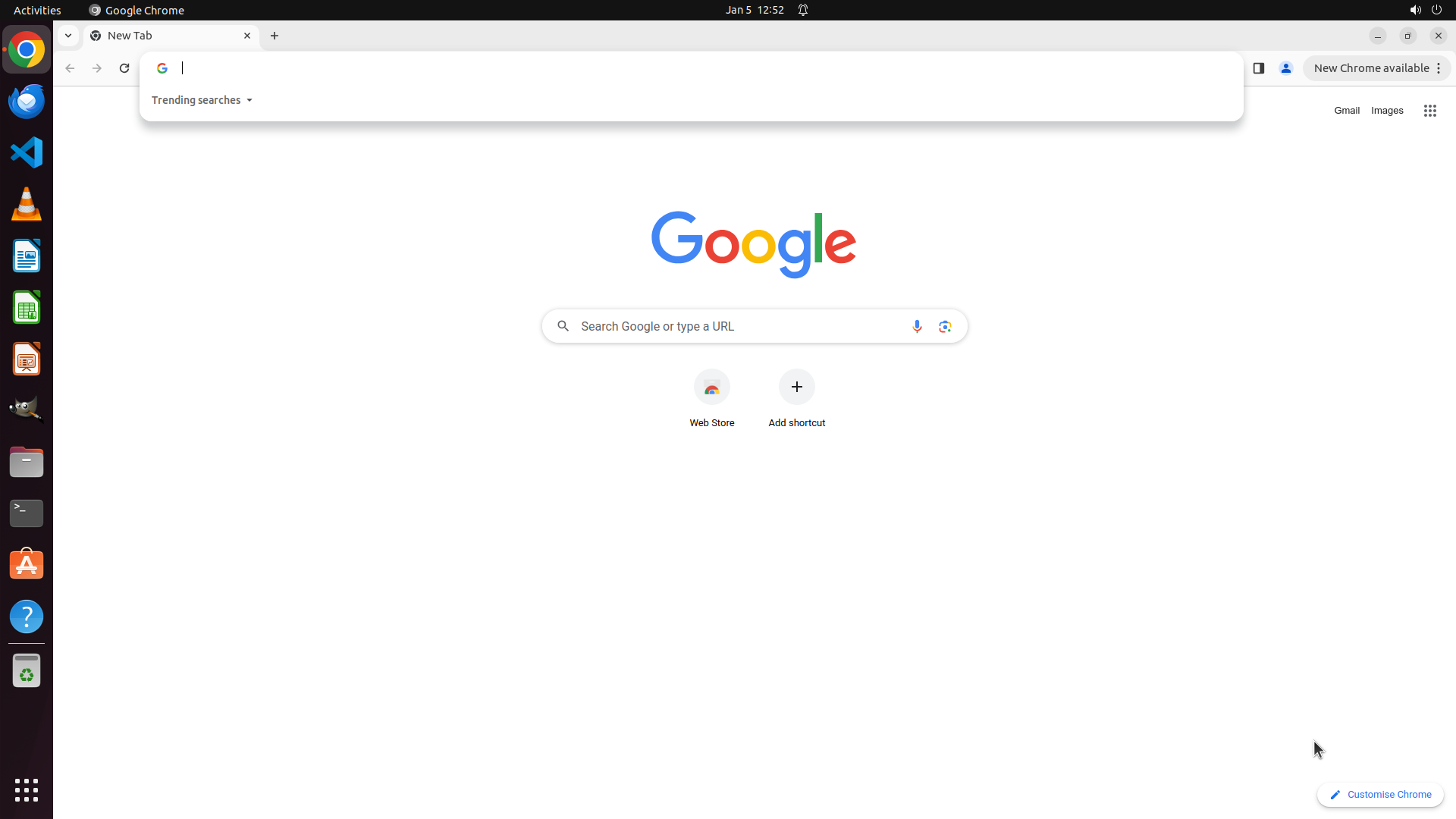}\\ \textbf{Node 1.3}\\ UCB: 0.289 \\ \texttt{\small Action: hotkey('ctrl', 'l')}}, name=n13
          [
            {\includegraphics[width=7cm]{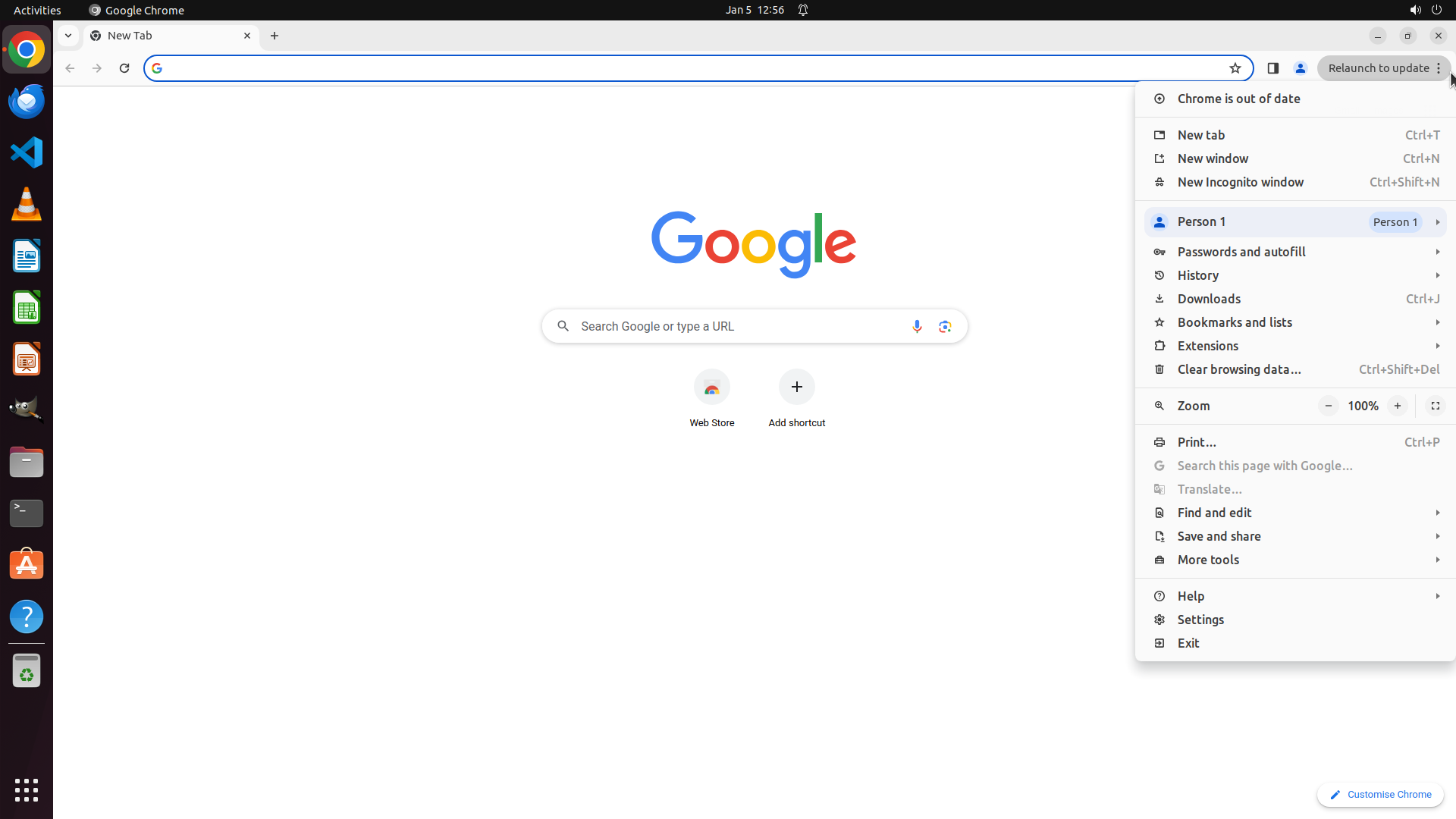}\\ \textbf{Node 1.1.2}\\ UCB: 0.221 \\ \texttt{\small Action: hotkey('ctrl', 'l')}}, name=n112, edge={draw=none}
          ]
        ]
      ]
      \draw[thick, -{Stealth[]}] (n11.south) -- (n111.north);
      \draw[thick, -{Stealth[]}] (n11.south) -- (n112.north);
    \end{forest}
  }
\end{figure}

For Task 1 (a relatively simple task in the Chrome domain), the hyperparameters were configured as follows: max MCTS iterations = 20, expansion factor = 3, and action chunking = 1. The search process successfully located the solution at depth 5.
\vspace{0.5cm}

\subsection{Task2: Help me export the first image from the doc file attached in the most recent email in Notes folder, and set this image as the new desktop background. }

In Task 2, a long-horizon task involving multi-app operations, we set the action chunking = 5, the maximum MCTS iterations = 15, and the expansion factor = 5. The generated search tree is illustrated in the following figure, while the subsequent figure displays screenshots of the successful path.

\tikzset{
    optimal_node/.style={fill=red!10, draw=red, thick},
    optimal_edge/.style={draw=red, line width=2pt},
    standard_node/.style={draw=black, thin}
}

\begin{figure}[htbp]
\centering
\resizebox{0.95\linewidth}{!}{
\begin{forest}
  for tree={
    grow=south,
    edge path={
      \noexpand\path[\forestoption{edge}]
        (!u.parent anchor) -- +(0,-15pt) -| (.child anchor)\forestoption{edge label};
    },
    parent anchor=south,
    child anchor=north,
    l sep=1cm,
    s sep=0.5cm,
    draw, 
    rounded corners,
    align=center,
    font=\sffamily\small,
    inner sep=4pt
  }
[\textbf{1}\\UCB: N/A, align=center, optimal_node [\textbf{1.1}\\UCB: 0.855, align=center, optimal_node, edge={optimal_edge} [\textbf{1.1.1}\\UCB: 0.140, align=center] [\textbf{1.1.2}\\UCB: 0.440, align=center] [\textbf{1.1.3}\\UCB: 0.040, align=center] [\textbf{1.1.4}\\UCB: 0.856, align=center, optimal_node, edge={optimal_edge} [\textbf{1.1.4.1}\\UCB: -0.063, align=center] [\textbf{1.1.4.2}\\UCB: 0.237, align=center] [\textbf{1.1.4.3}\\UCB: 0.337, align=center] [\textbf{1.1.4.4}\\UCB: -0.063, align=center] [\textbf{1.1.4.5}\\UCB: 0.856, align=center, optimal_node, edge={optimal_edge} [\textbf{1.1.4.5.1}\\UCB: 0.334, align=center] [\textbf{1.1.4.5.2}\\UCB: -0.166, align=center] [\textbf{1.1.4.5.3}\\UCB: 0.857, align=center, optimal_node, edge={optimal_edge} [\textbf{1.1.4.5.3.1}\\UCB: 0.857, align=center, optimal_node, edge={optimal_edge} [\textbf{1.1.4.5.3.1.1}\\UCB: 0.826, align=center] [\textbf{1.1.4.5.3.1.2}\\UCB: 0.426, align=center] [\textbf{1.1.4.5.3.1.3}\\UCB: 0.859, align=center, optimal_node, edge={optimal_edge} [\textbf{1.1.4.5.3.1.3.1}\\UCB: 0.421, align=center] [\textbf{1.1.4.5.3.1.3.2}\\UCB: 0.811, align=center, optimal_node, edge={optimal_edge} [\textbf{1.1.4.5.3.1.3.2.1}\\UCB: 0.815\\SUCCEED, align=center, optimal_node, edge={optimal_edge}] [\textbf{1.1.4.5.3.1.3.2.2}\\UCB: 0.315, align=center]]]] [\textbf{1.1.4.5.3.2}\\UCB: -0.170, align=center]]]]] [\textbf{1.2}\\UCB: 0.342, align=center]]
\end{forest}
}
\caption{MCTS Tree Structure for Task2(Optimal Path Highlighted)}
\end{figure}
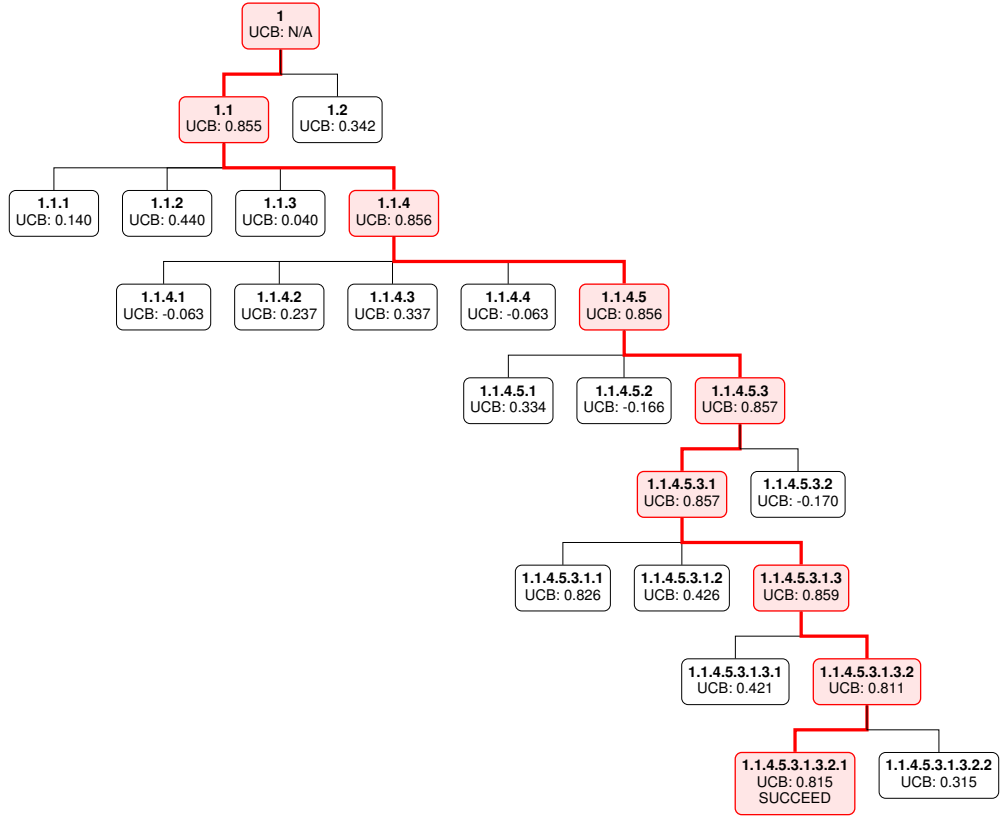

\begin{figure}[htbp]
  \centering
  
  \textbf{Node 1} \\
  \begin{subfigure}[t]{0.19\linewidth}
    \includegraphics[width=\linewidth]{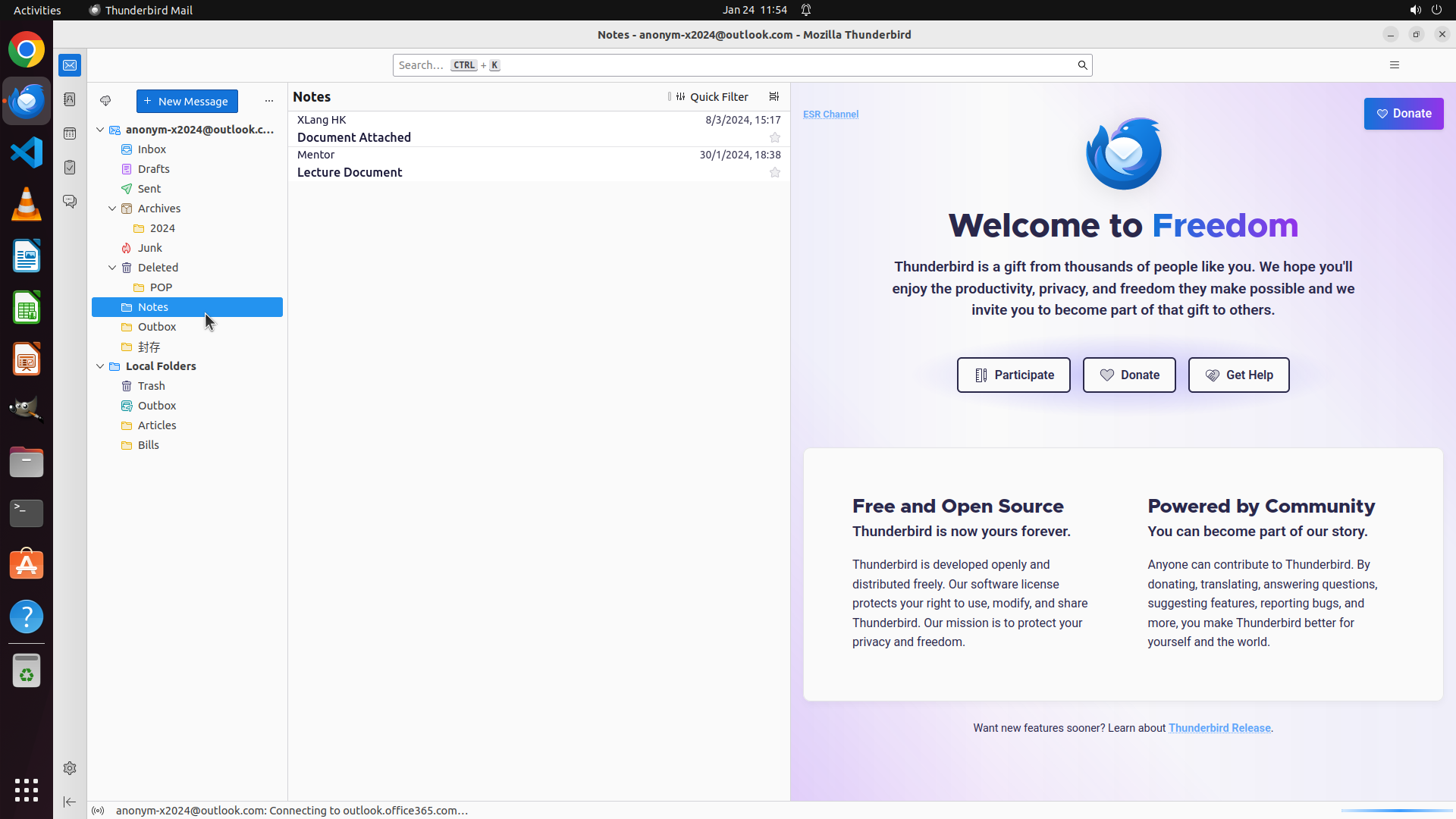}
    \\ \centering \texttt{click(266, 408)}
  \end{subfigure}%
 \hfill   \begin{subfigure}[t]{0.19\linewidth}
    \includegraphics[width=\linewidth]{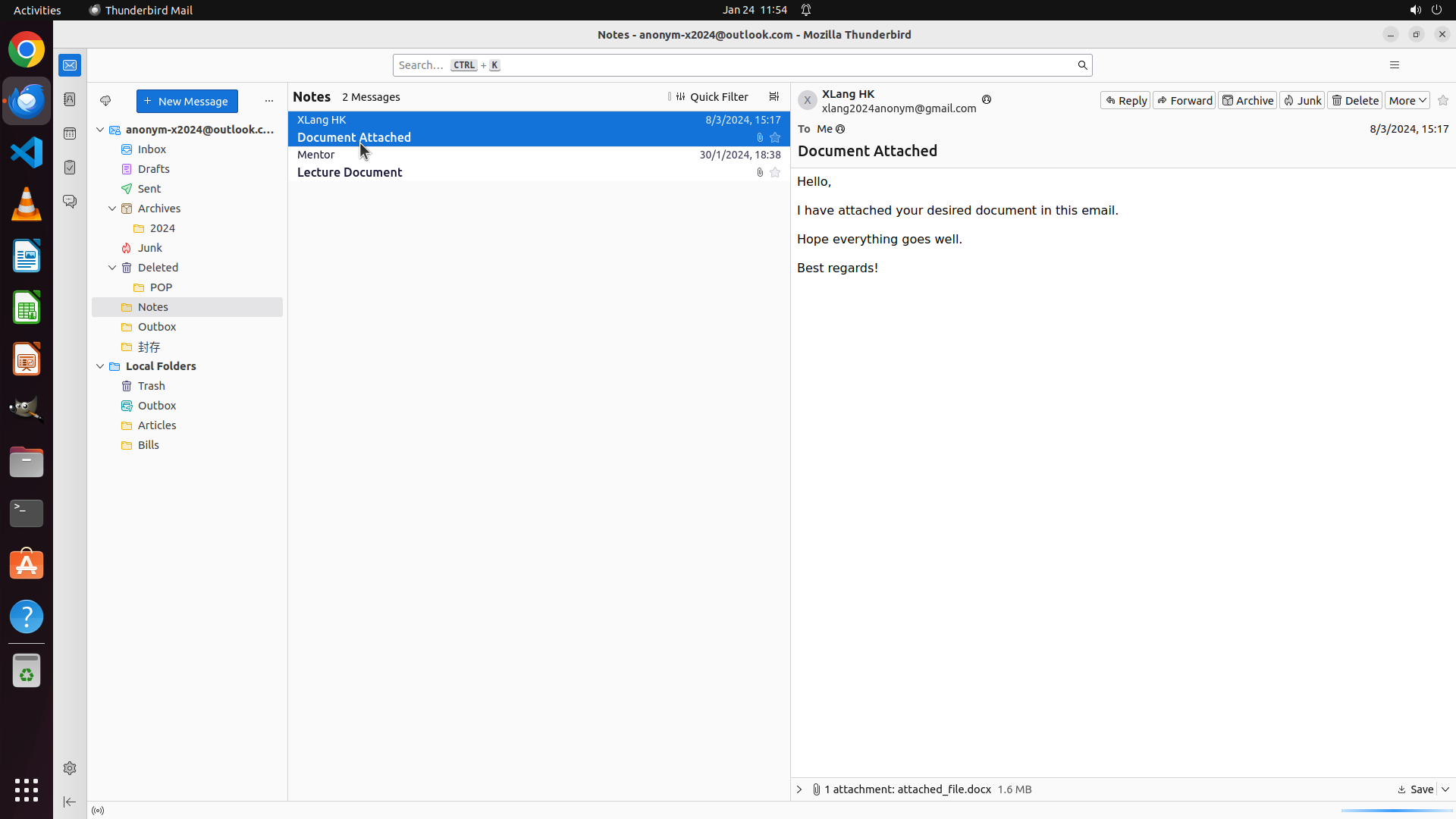}
    \\ \centering \texttt{click(470, 183)}
  \end{subfigure}%
 \hfill   \begin{subfigure}[t]{0.19\linewidth}
    \includegraphics[width=\linewidth]{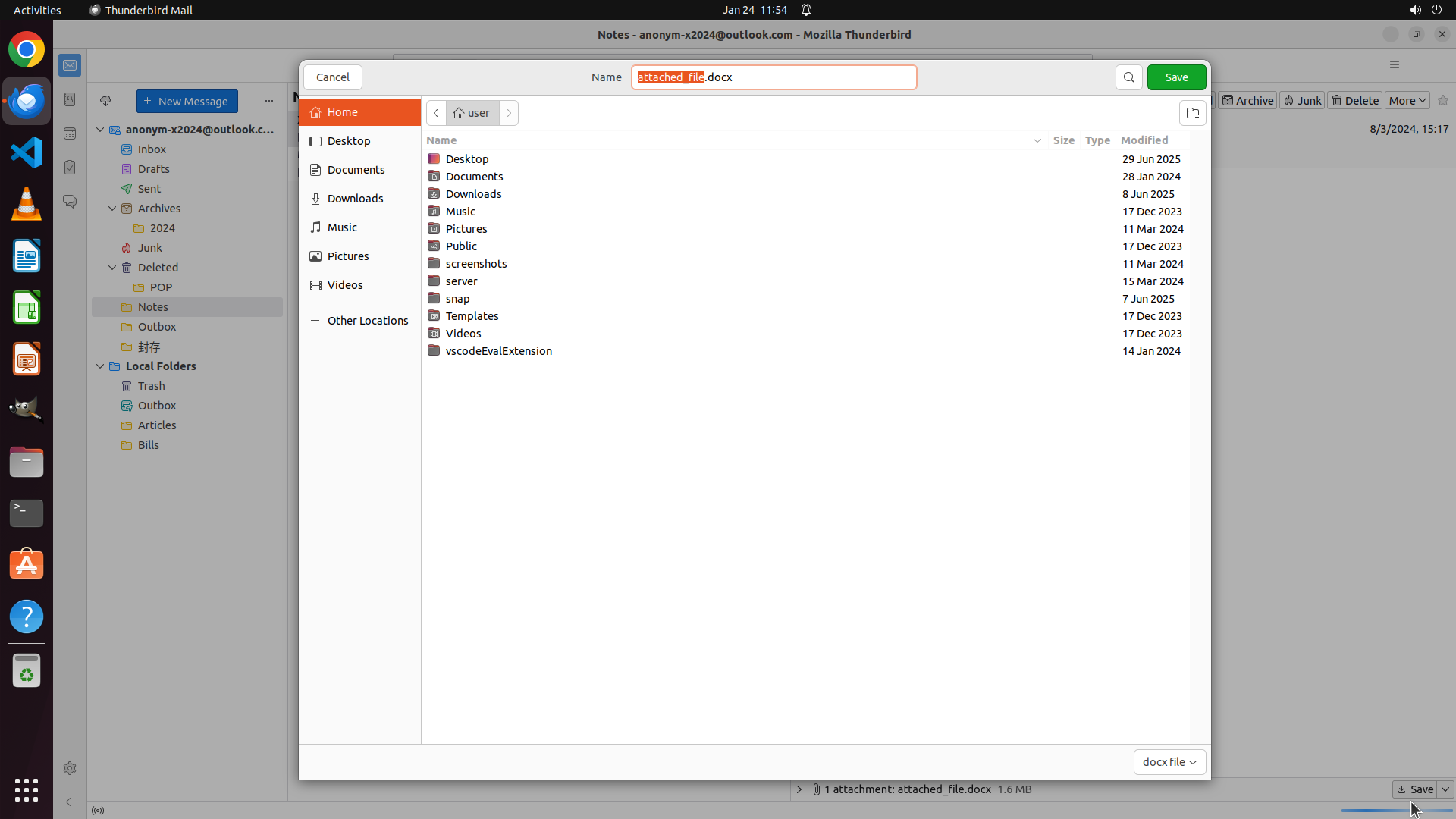}
    \\ \centering \texttt{click(1856, 1052)}
  \end{subfigure}%
 \hfill   \begin{subfigure}[t]{0.19\linewidth}
    \includegraphics[width=\linewidth]{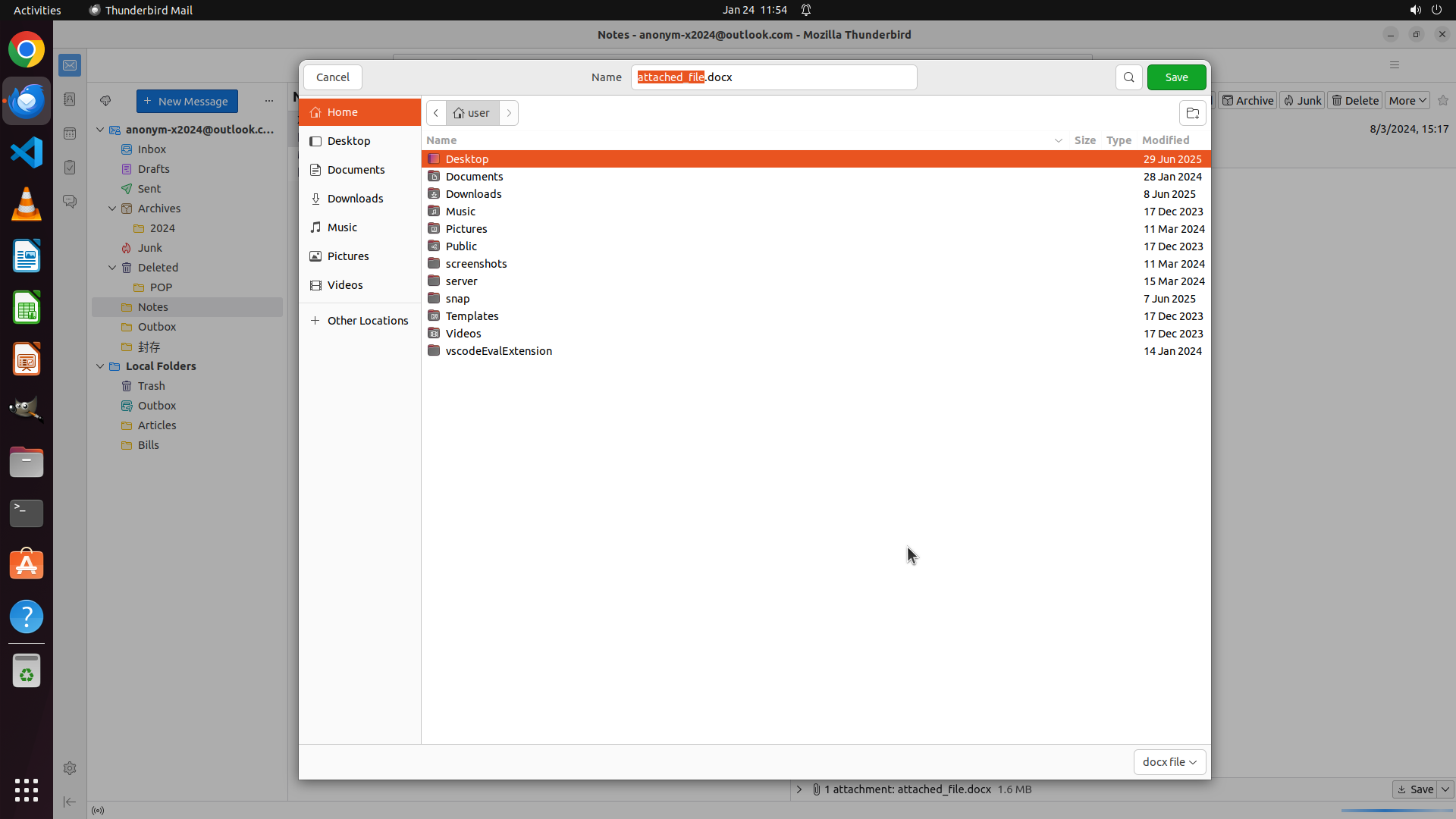}
    \\ \centering \texttt{click(1192, 716)}
  \end{subfigure}%
 \hfill   \begin{subfigure}[t]{0.19\linewidth}
    \includegraphics[width=\linewidth]{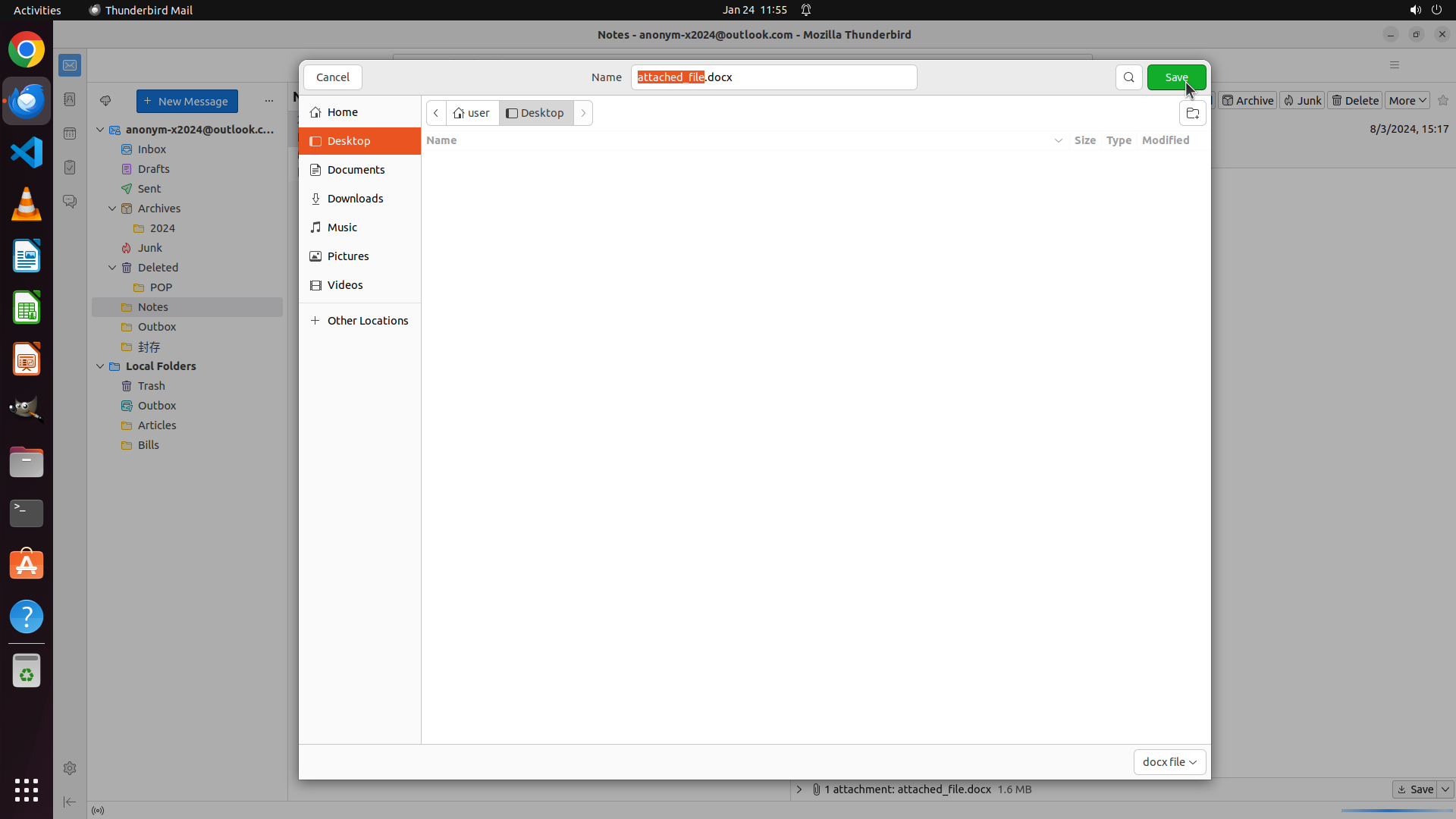}
    \\ \centering \texttt{click(1559, 103)}
  \end{subfigure}%

  \par\vspace{1em}
  \textbf{Node 1.1} \\
  \begin{subfigure}[t]{0.19\linewidth}
    \includegraphics[width=\linewidth]{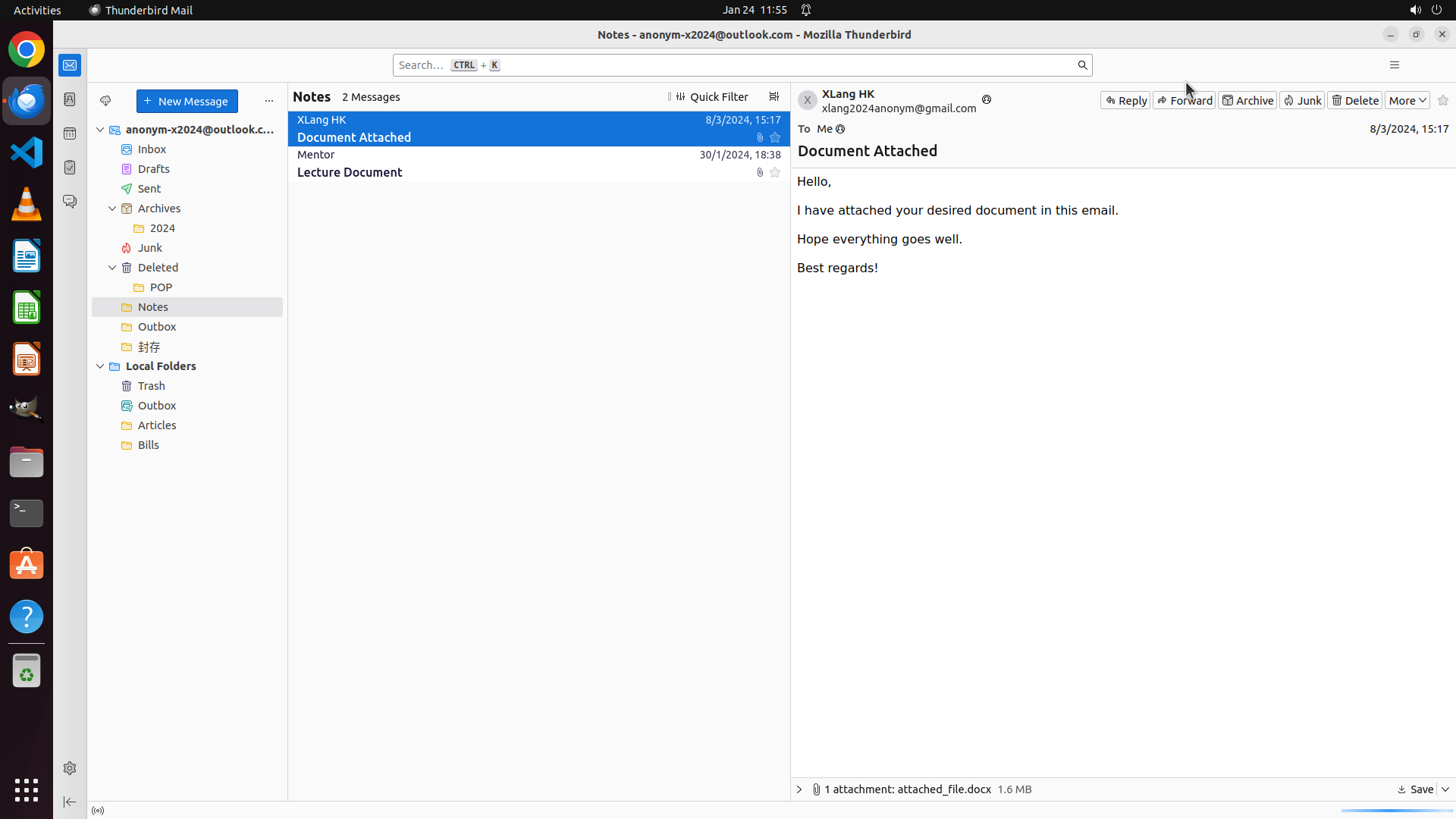}
    \\ \centering \texttt{hotkey(enter)}
  \end{subfigure}%
 \hfill   \begin{subfigure}[t]{0.19\linewidth}
    \includegraphics[width=\linewidth]{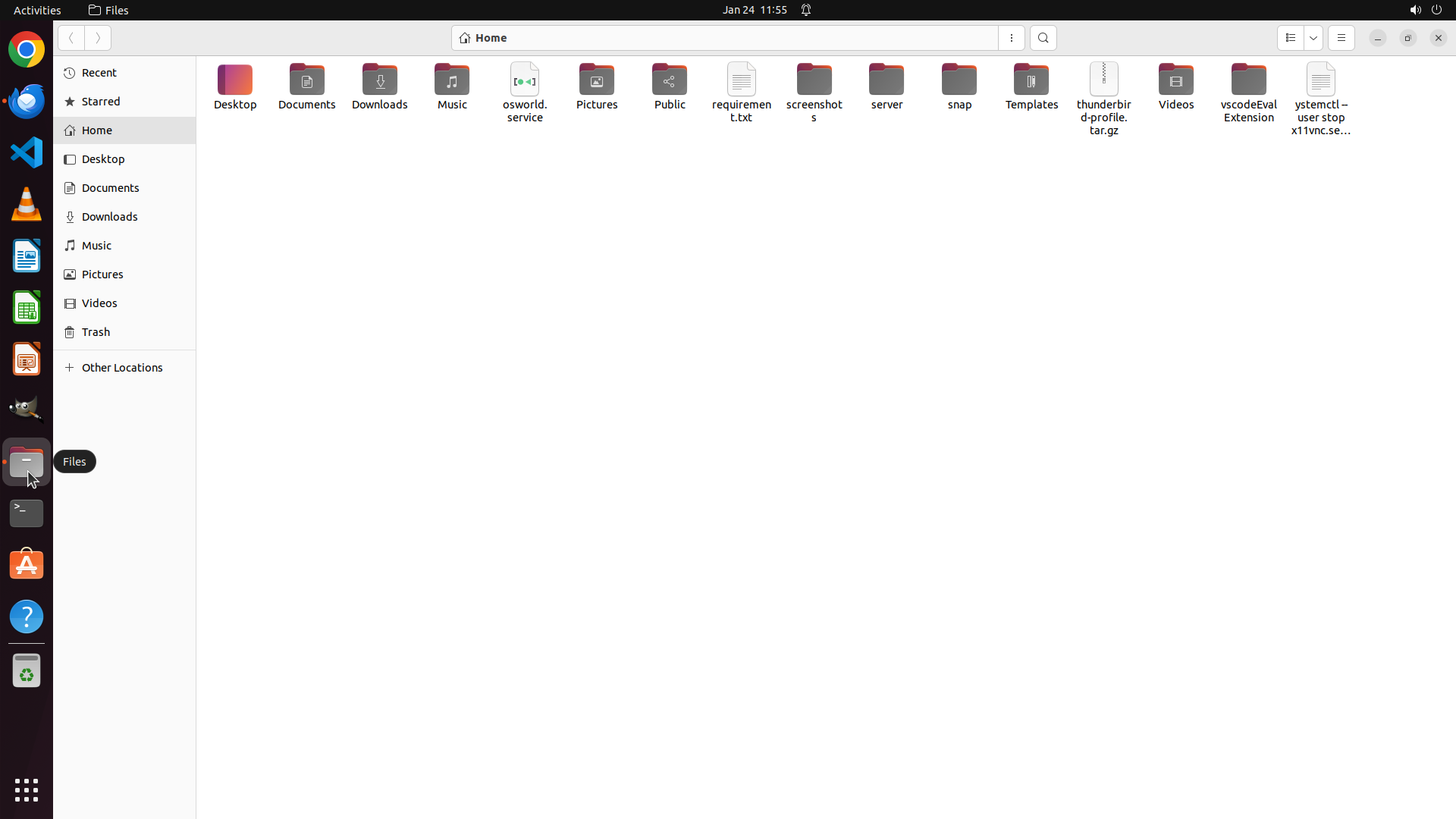}
    \\ \centering \texttt{click(32, 616)}
  \end{subfigure}%
 \hfill   \begin{subfigure}[t]{0.19\linewidth}
    \includegraphics[width=\linewidth]{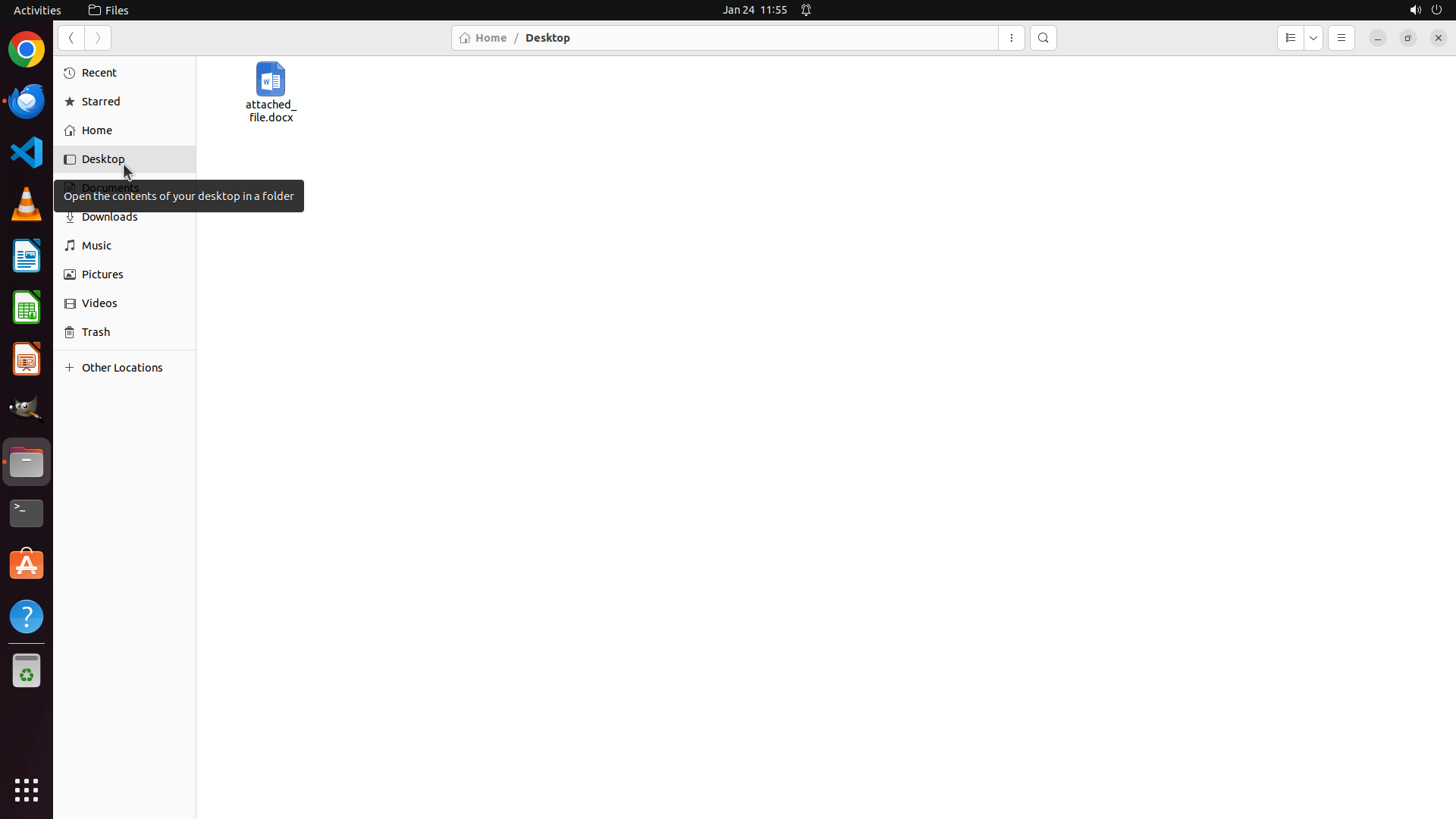}
    \\ \centering \texttt{click(158, 211)}
  \end{subfigure}%
 \hfill   \begin{subfigure}[t]{0.19\linewidth}
    \includegraphics[width=\linewidth]{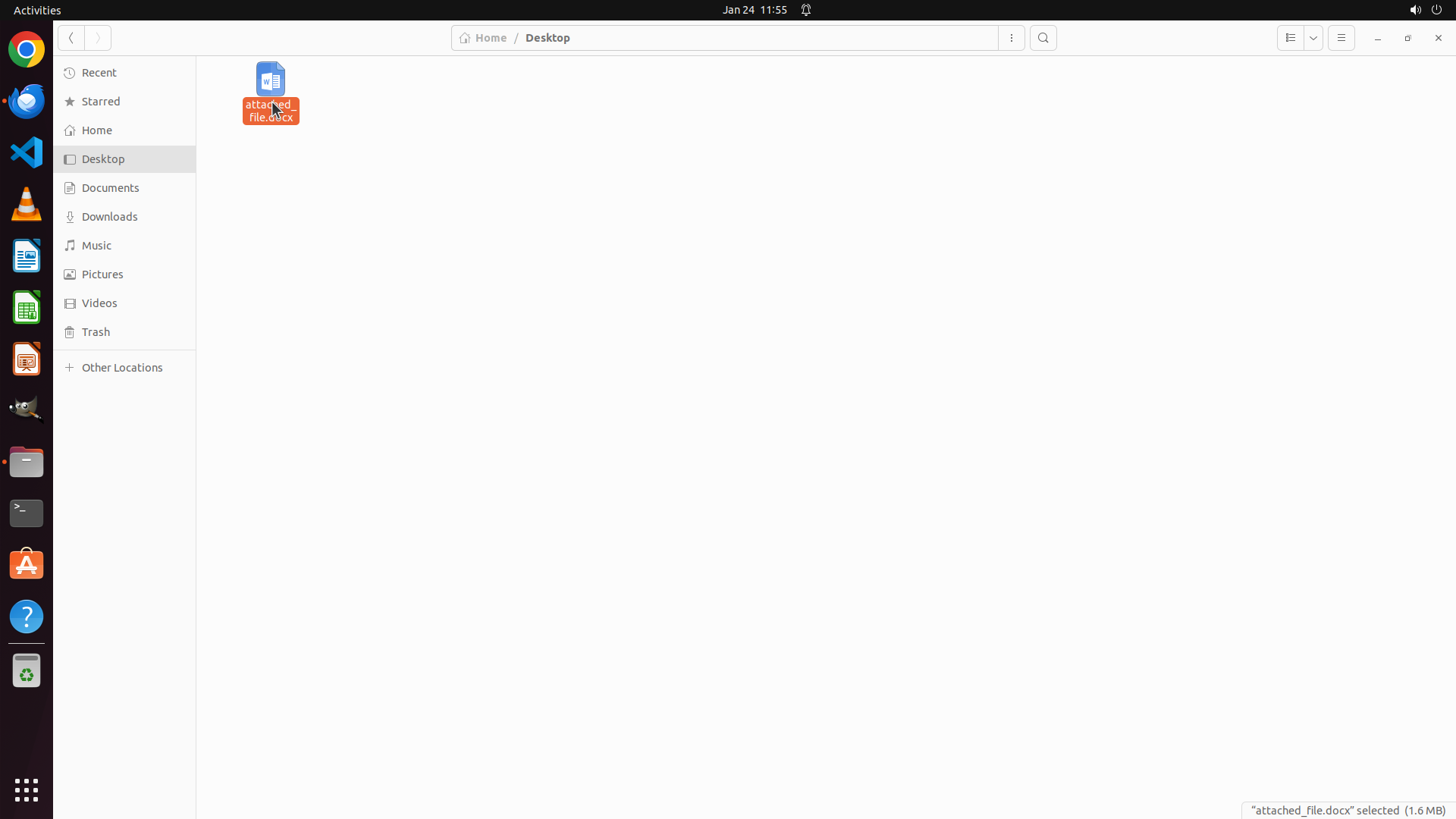}
    \\ \centering \texttt{click(354, 129)}
  \end{subfigure}%
 \hfill   \begin{subfigure}[t]{0.19\linewidth}
    \includegraphics[width=\linewidth]{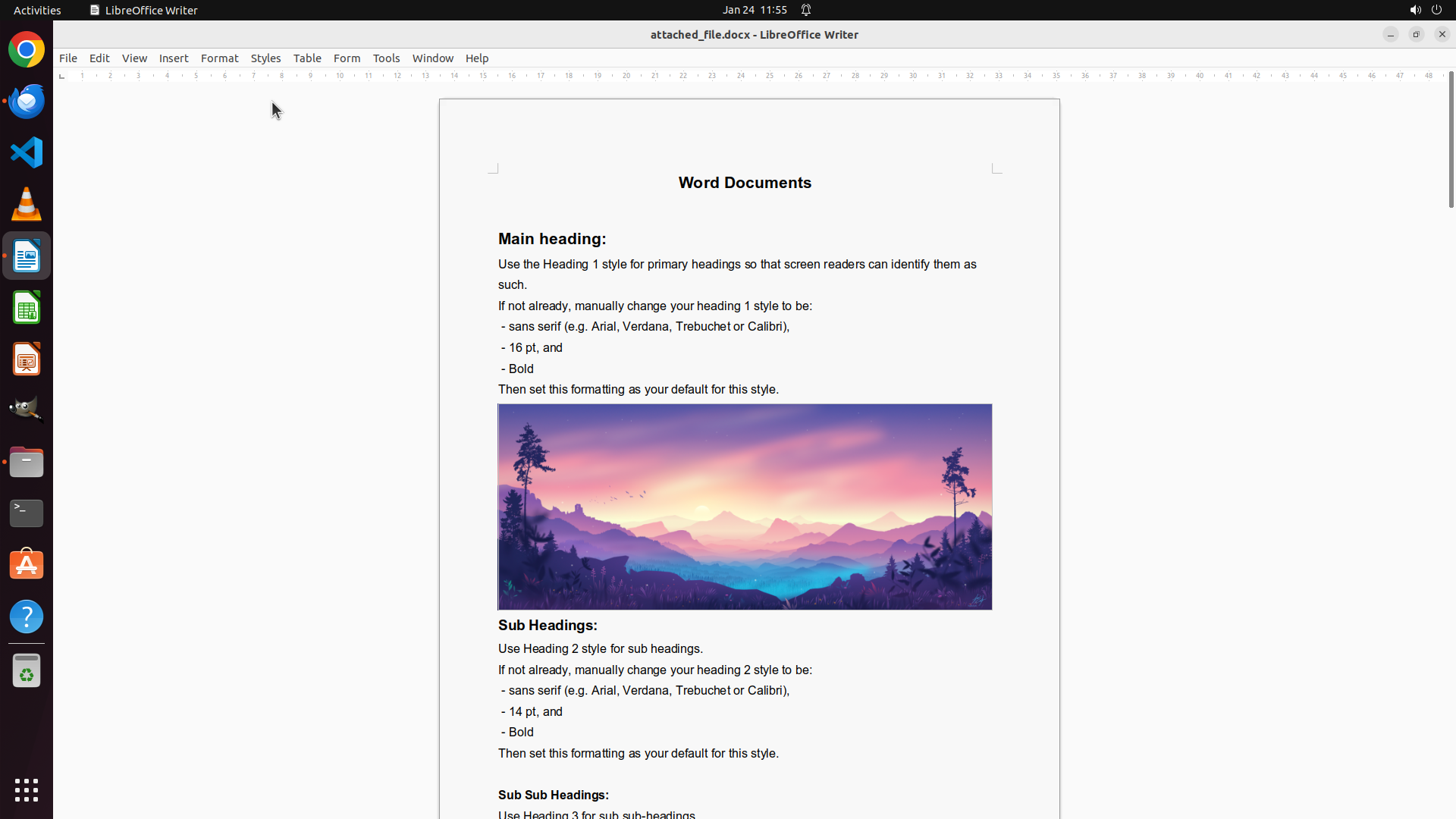}
    \\ \centering \texttt{time.sleep(1.333)}
  \end{subfigure}%

  \par\vspace{1em}
  \textbf{Node 1.1.4} \\
  \begin{subfigure}[t]{0.19\linewidth}
    \includegraphics[width=\linewidth]{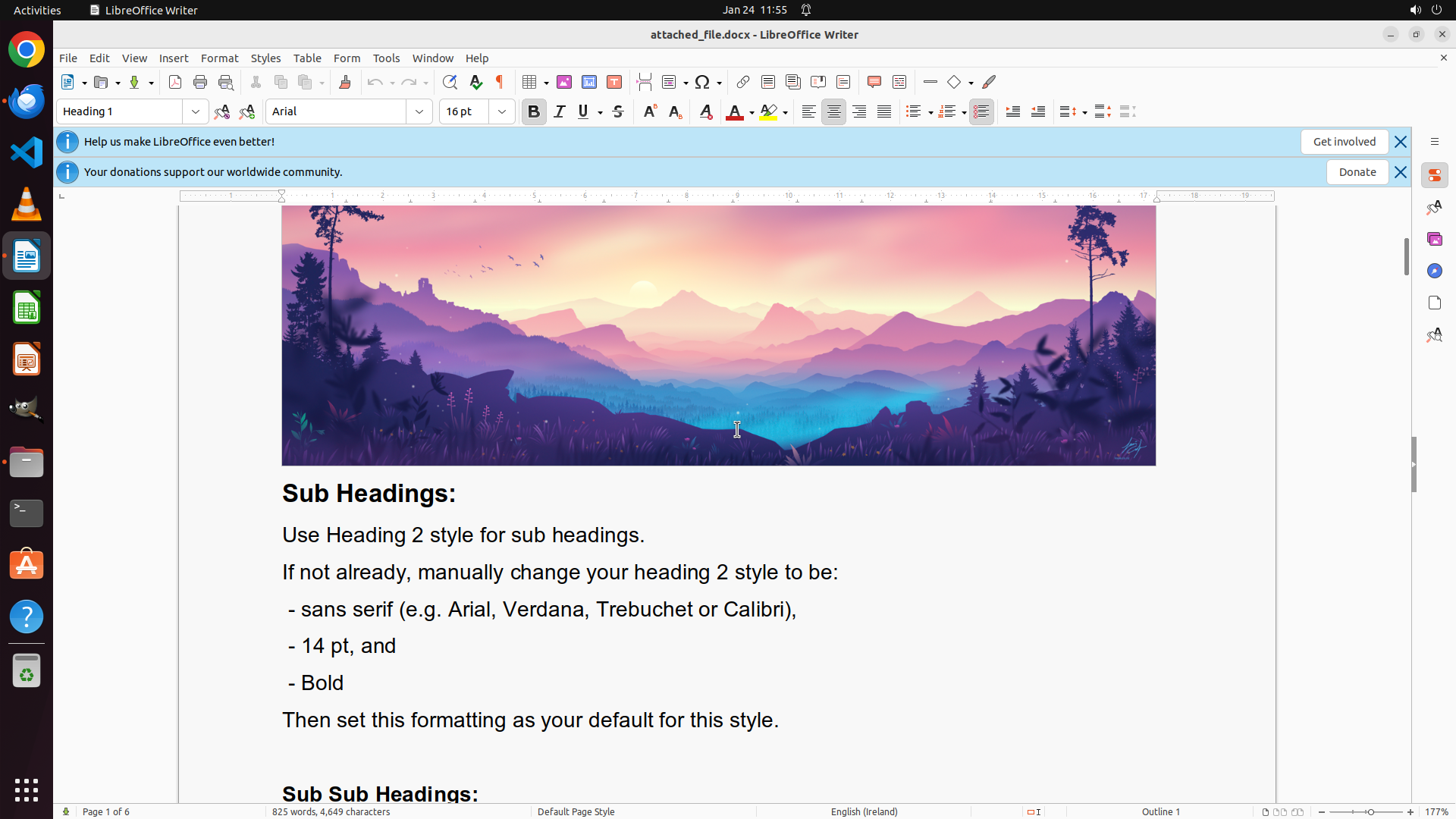}
    \\ \centering \texttt{scroll(-10)}
  \end{subfigure}%
 \hfill   \begin{subfigure}[t]{0.19\linewidth}
    \includegraphics[width=\linewidth]{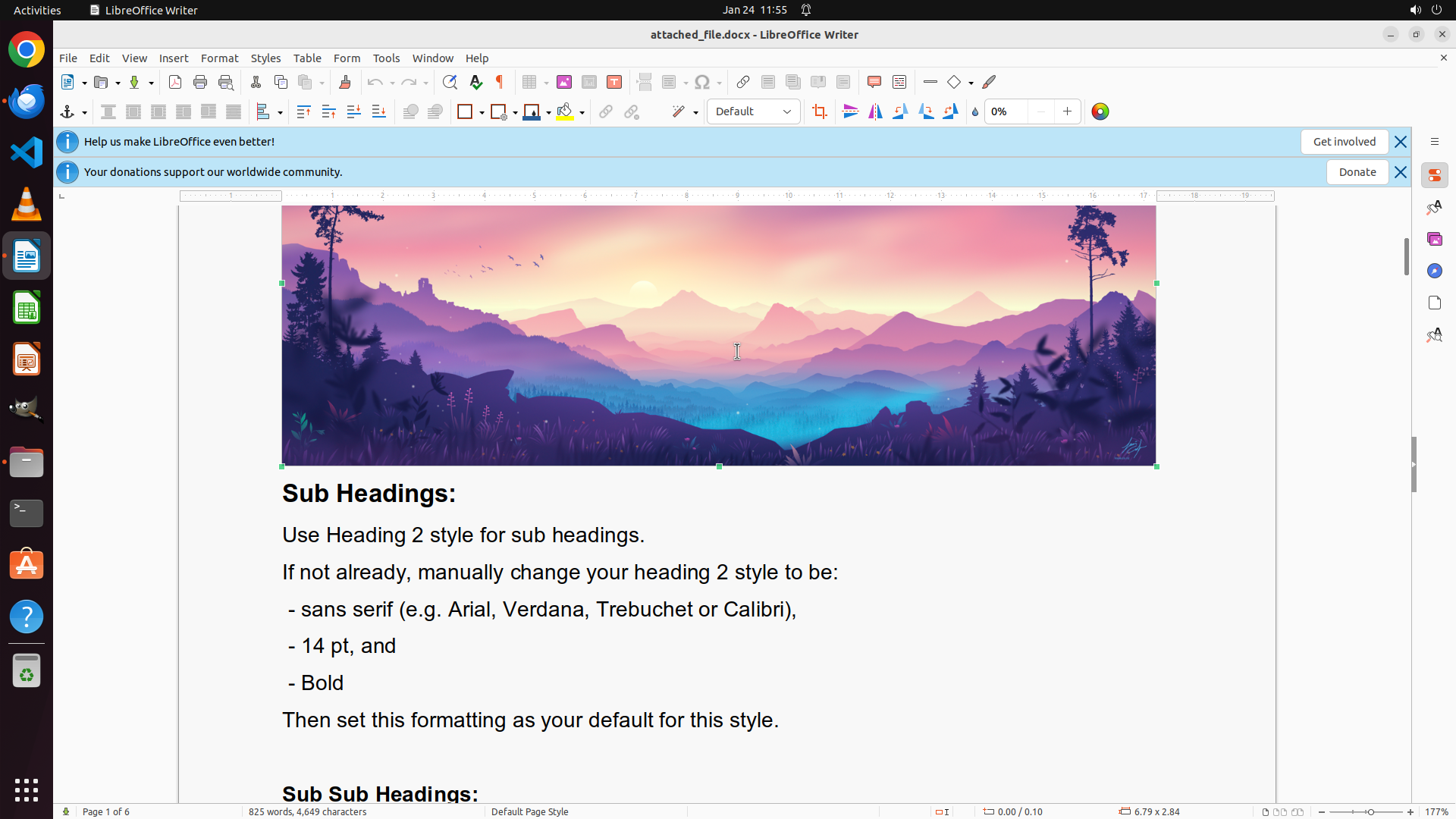}
    \\ \centering \texttt{click(958, 448)}
  \end{subfigure}%
 \hfill   \begin{subfigure}[t]{0.19\linewidth}
    \includegraphics[width=\linewidth]{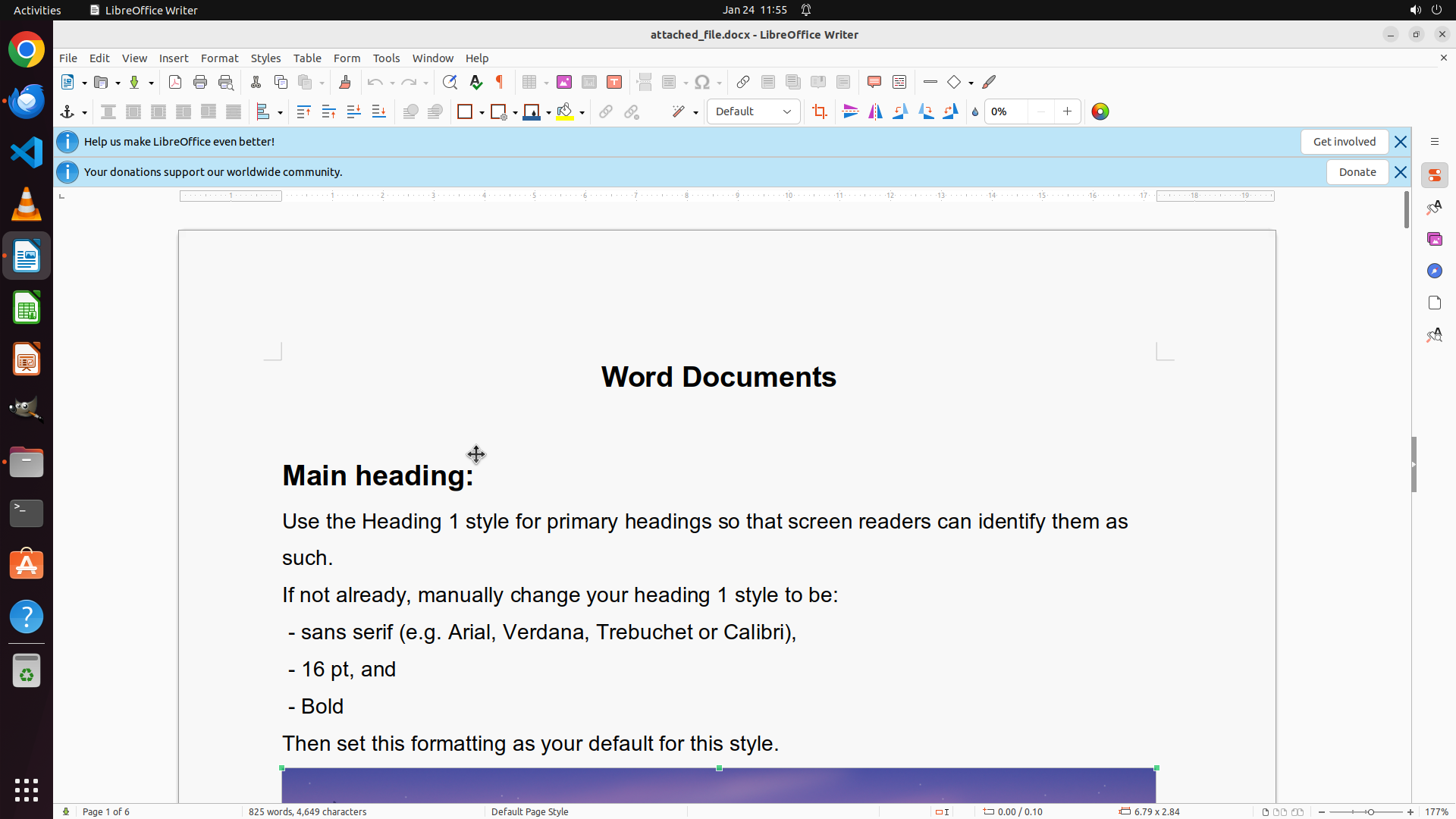}
    \\ \centering \texttt{scroll(20)}
  \end{subfigure}%
 \hfill   \begin{subfigure}[t]{0.19\linewidth}
    \includegraphics[width=\linewidth]{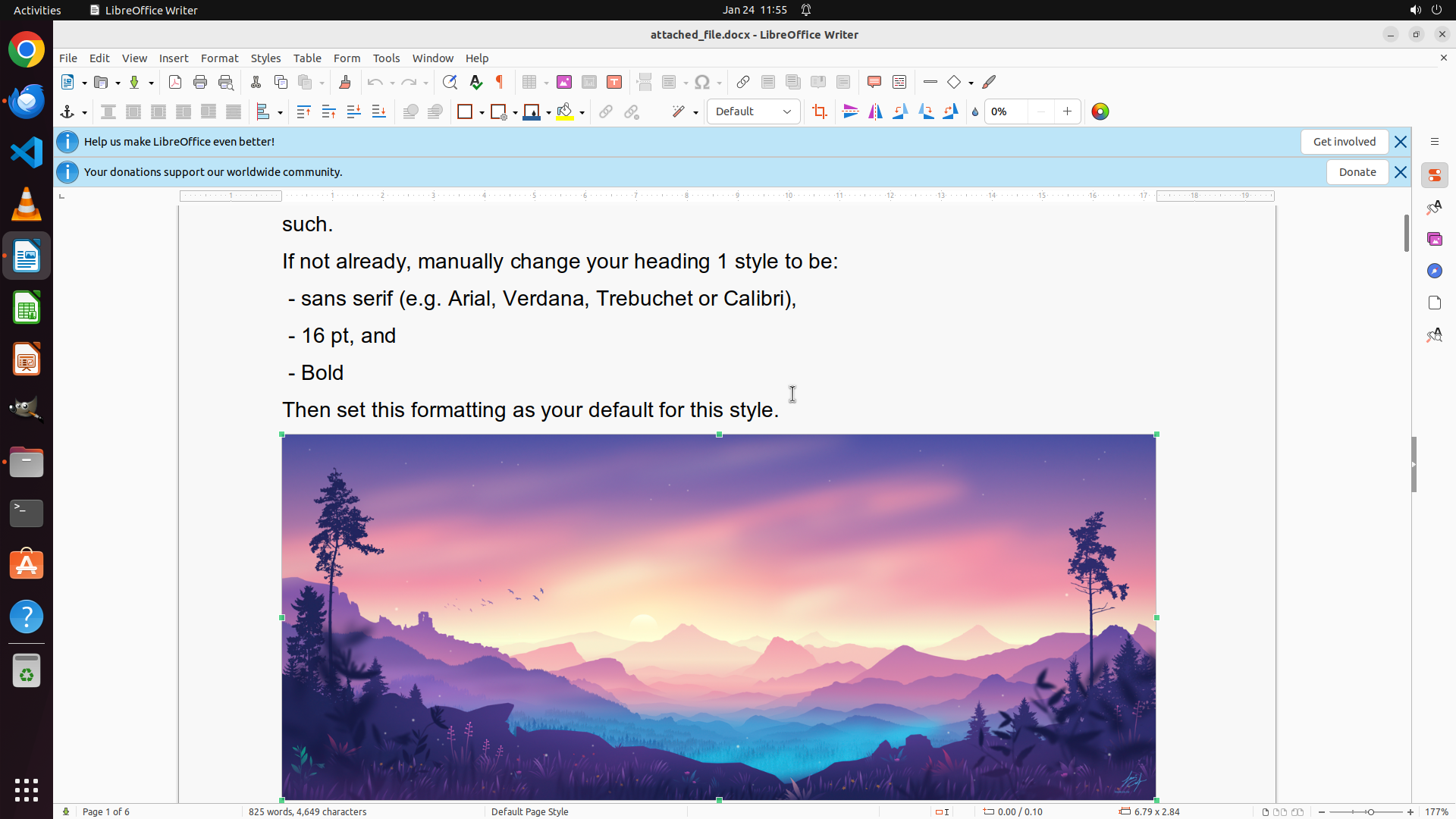}
    \\ \centering \texttt{scroll(-5)}
  \end{subfigure}%
 \hfill   \begin{subfigure}[t]{0.19\linewidth}
    \includegraphics[width=\linewidth]{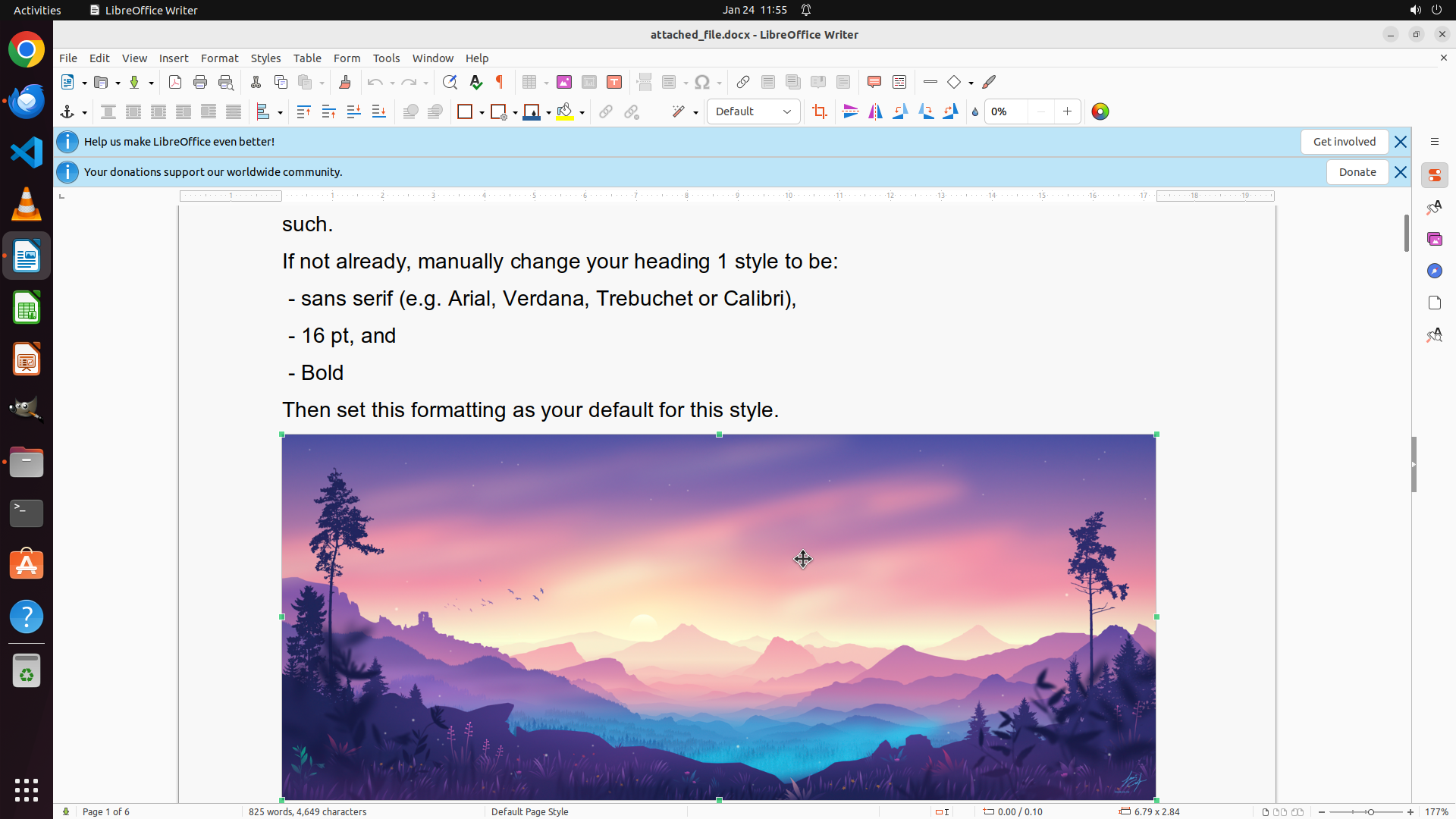}
    \\ \centering \texttt{click(1043, 721)}
  \end{subfigure}%

  \par\vspace{1em}
  \textbf{Node 1.1.4.5} \\
  \begin{subfigure}[t]{0.19\linewidth}
    \includegraphics[width=\linewidth]{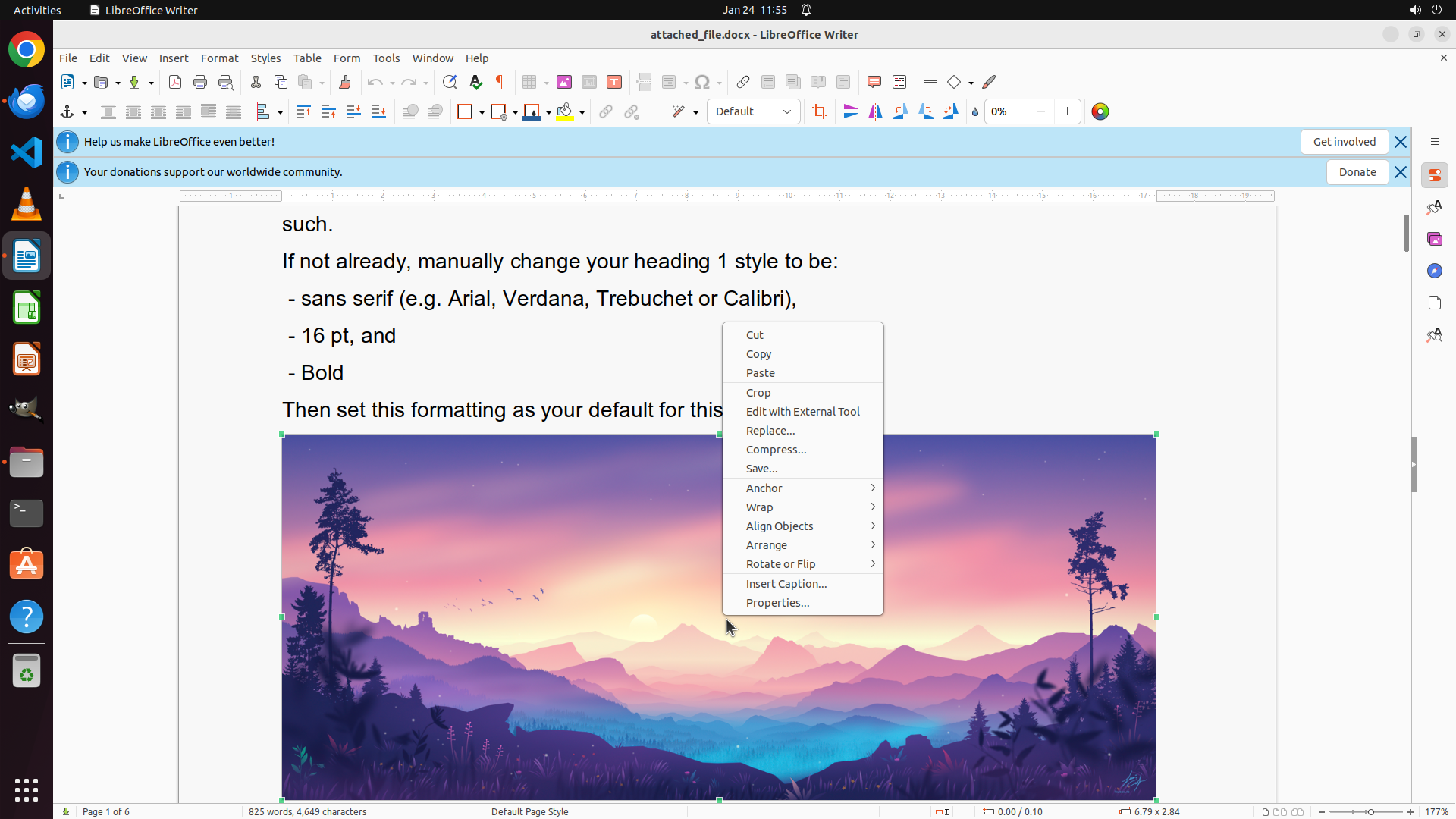}
    \\ \centering \texttt{click(953, 811)}
  \end{subfigure}%
 \hfill   \begin{subfigure}[t]{0.19\linewidth}
    \includegraphics[width=\linewidth]{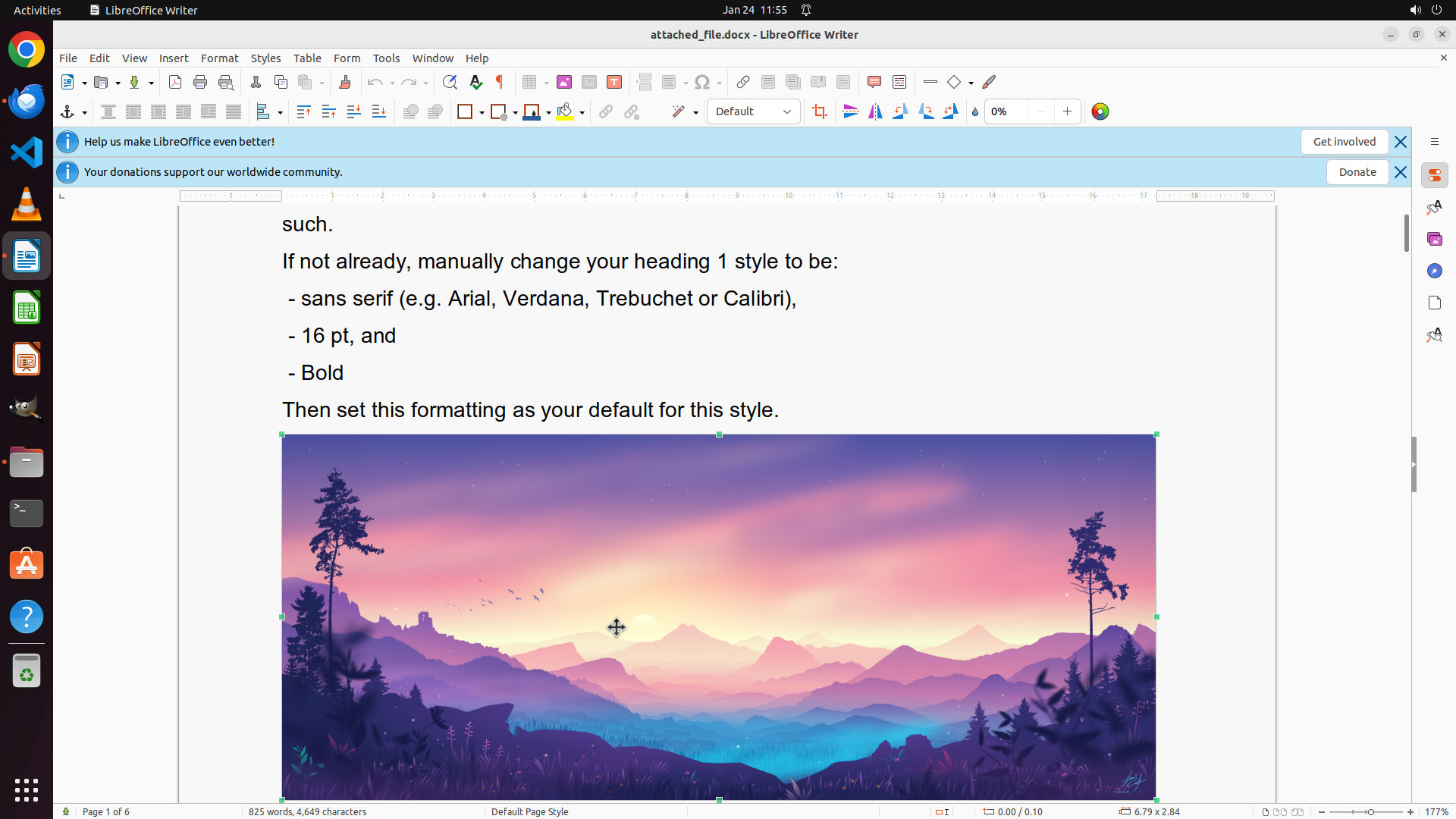}
    \\ \centering \texttt{click(797, 811)}
  \end{subfigure}%
 \hfill   \begin{subfigure}[t]{0.19\linewidth}
    \includegraphics[width=\linewidth]{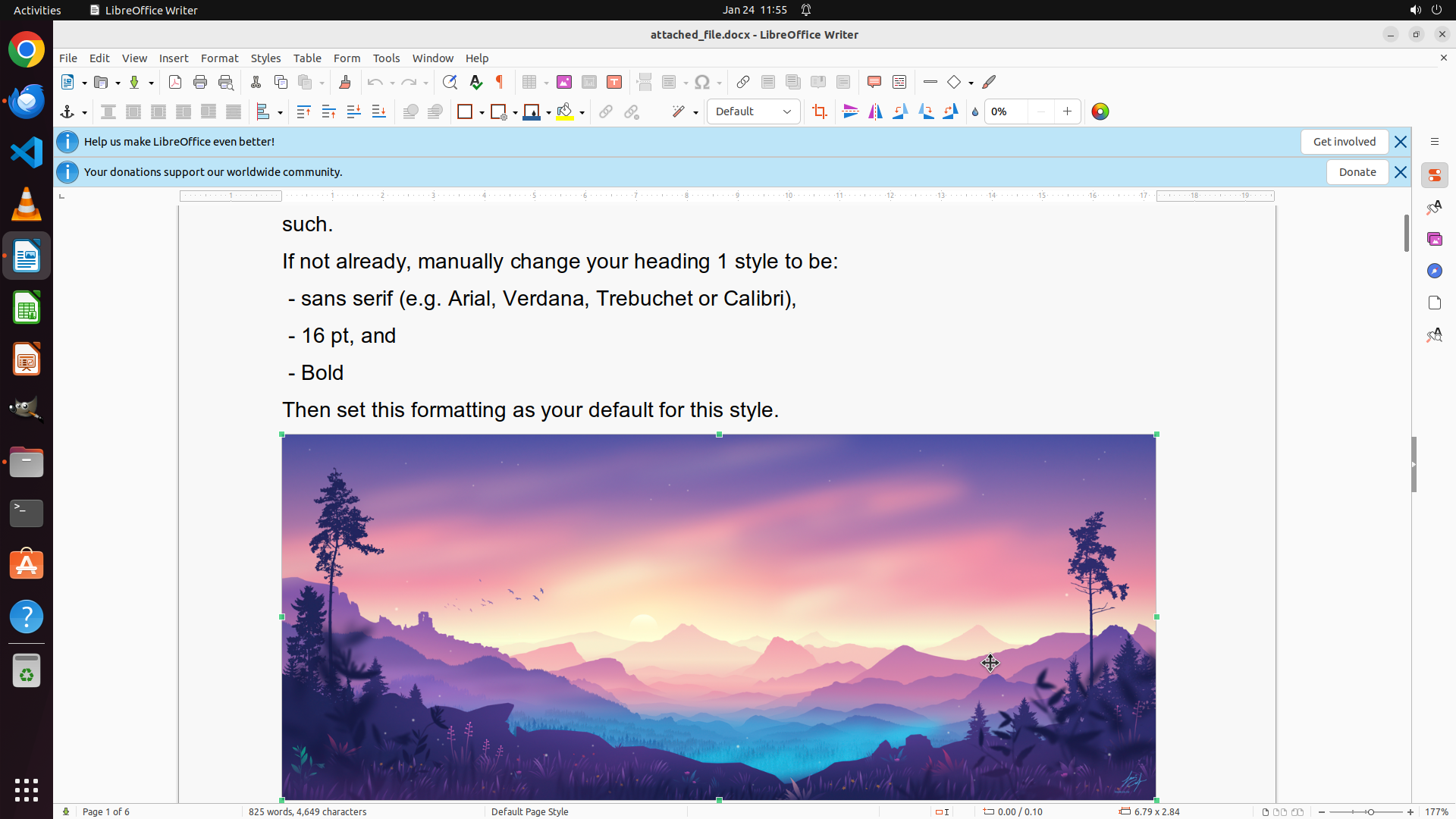}
    \\ \centering \texttt{click(1290, 858)}
  \end{subfigure}%
 \hfill   \begin{subfigure}[t]{0.19\linewidth}
    \includegraphics[width=\linewidth]{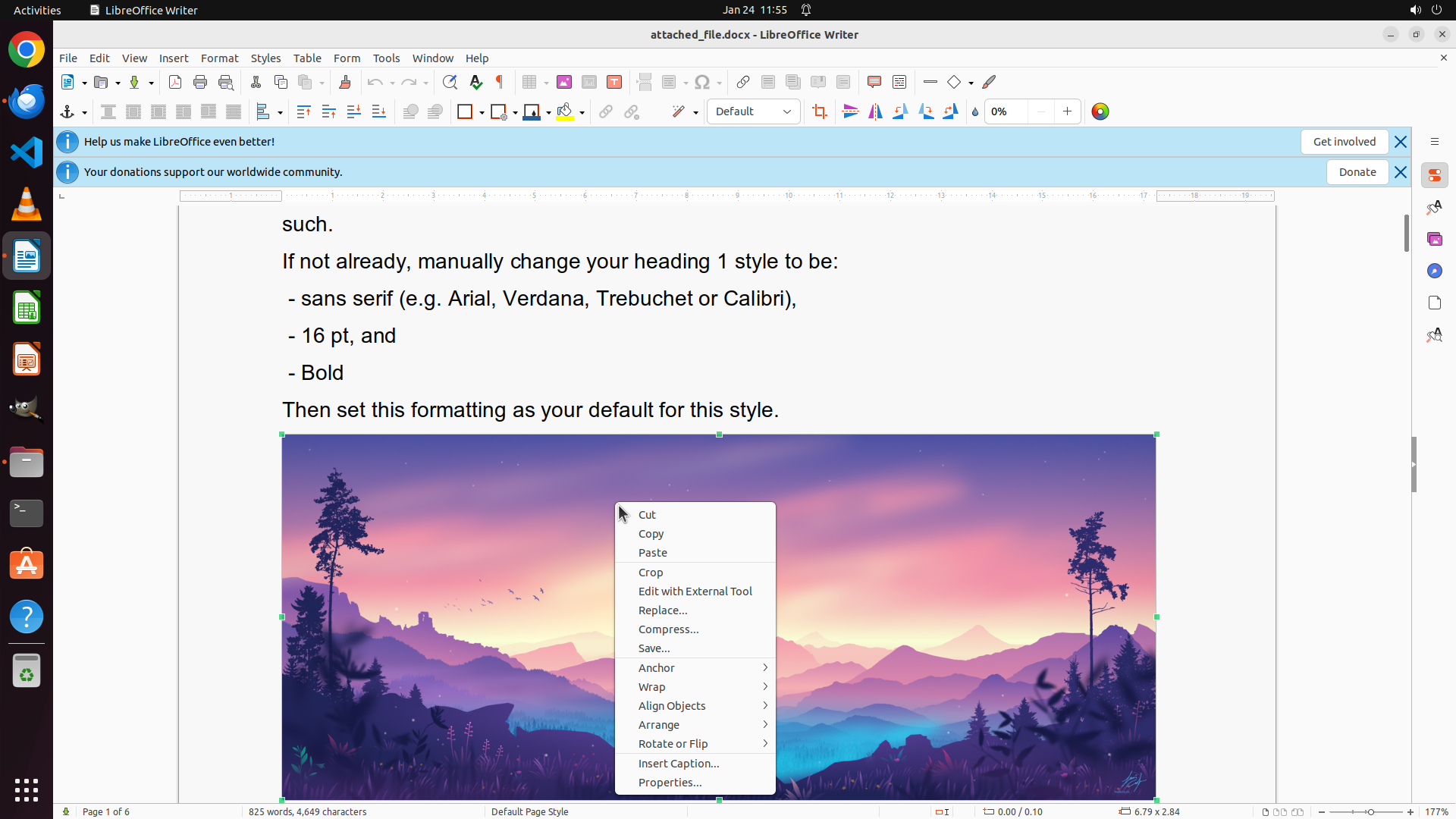}
    \\ \centering \texttt{click(811, 661)}
  \end{subfigure}%
 \hfill   \begin{subfigure}[t]{0.19\linewidth}
    \includegraphics[width=\linewidth]{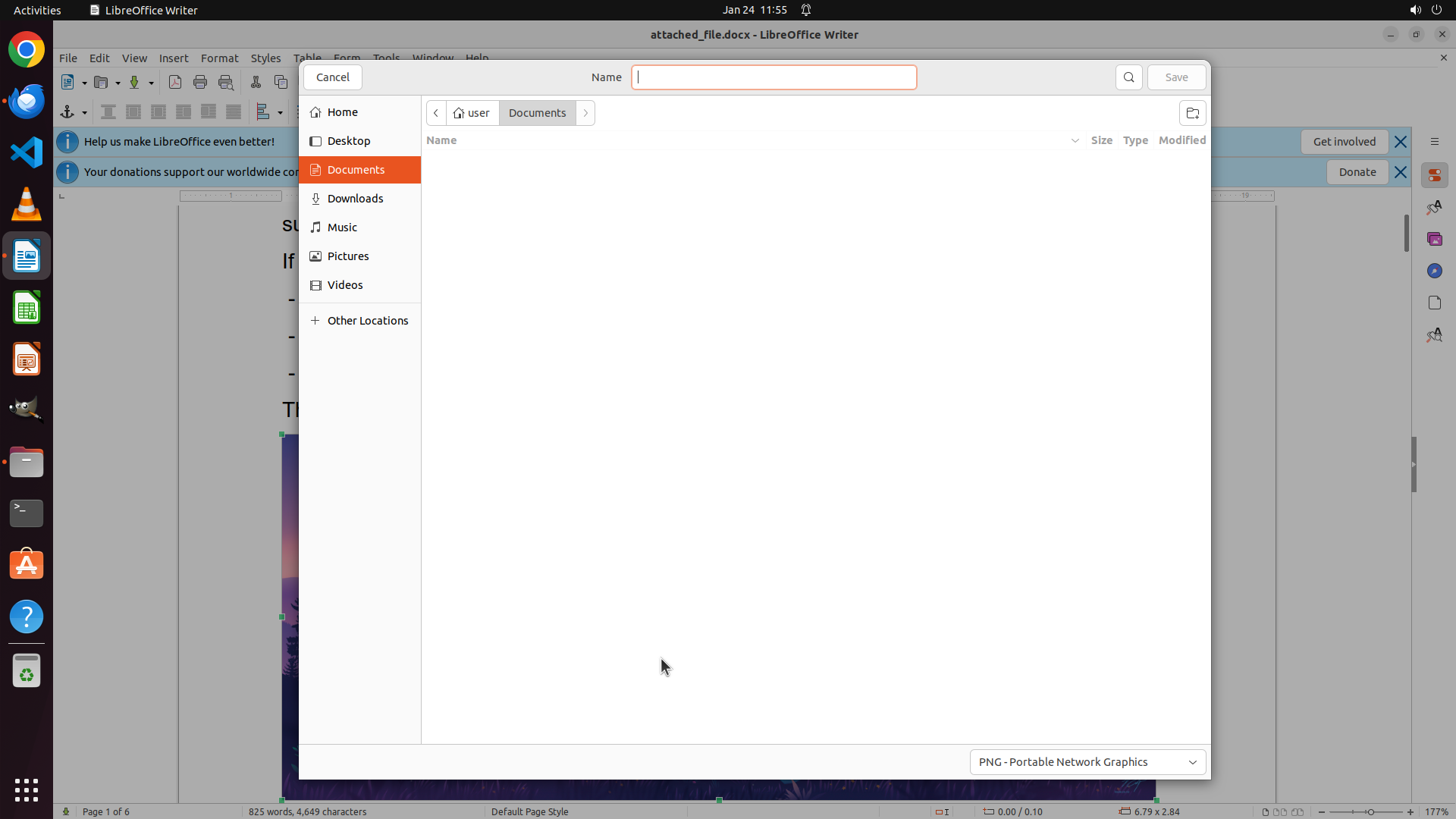}
    \\ \centering \texttt{click(867, 863)}
  \end{subfigure}%

  \par\vspace{1em}
  \textbf{Node 1.1.4.5.3} \\
  \begin{subfigure}[t]{0.19\linewidth}
    \includegraphics[width=\linewidth]{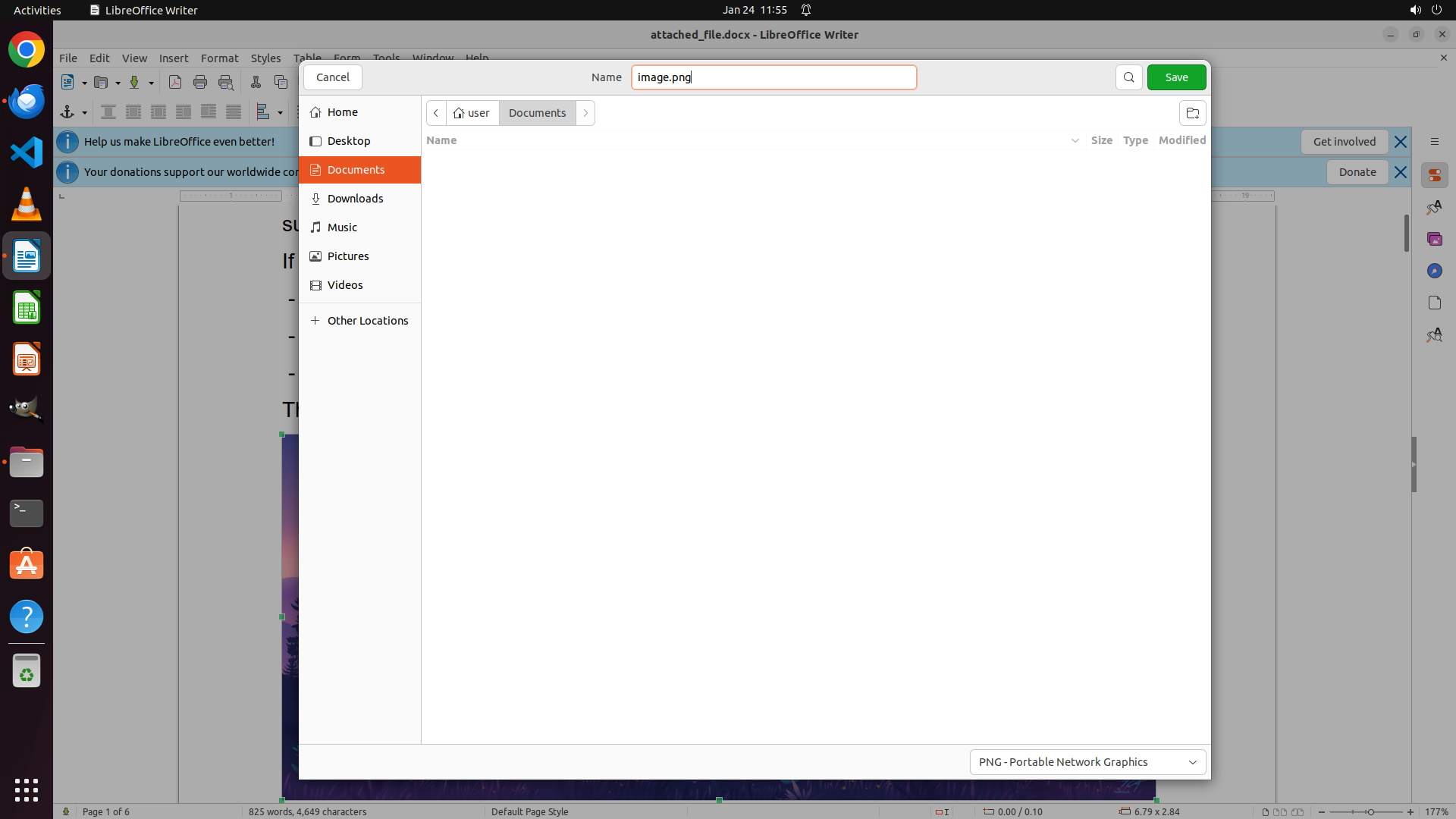}
    \\ \centering \texttt{click(1028, 103)}
  \end{subfigure}%
 \hfill   \begin{subfigure}[t]{0.19\linewidth}
    \includegraphics[width=\linewidth]{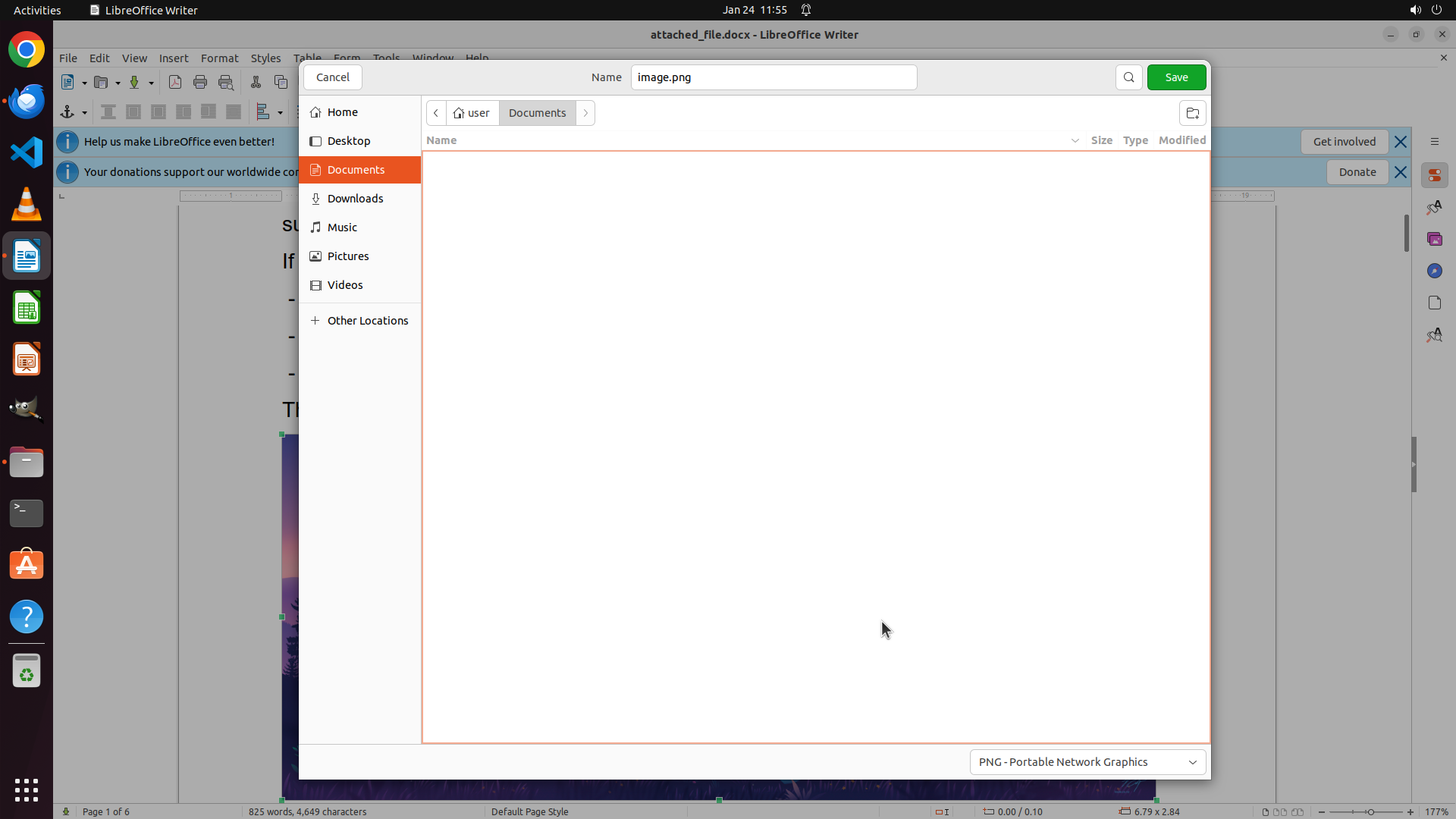}
    \\ \centering \texttt{click(1158, 814)}
  \end{subfigure}%
 \hfill   \begin{subfigure}[t]{0.19\linewidth}
    \includegraphics[width=\linewidth]{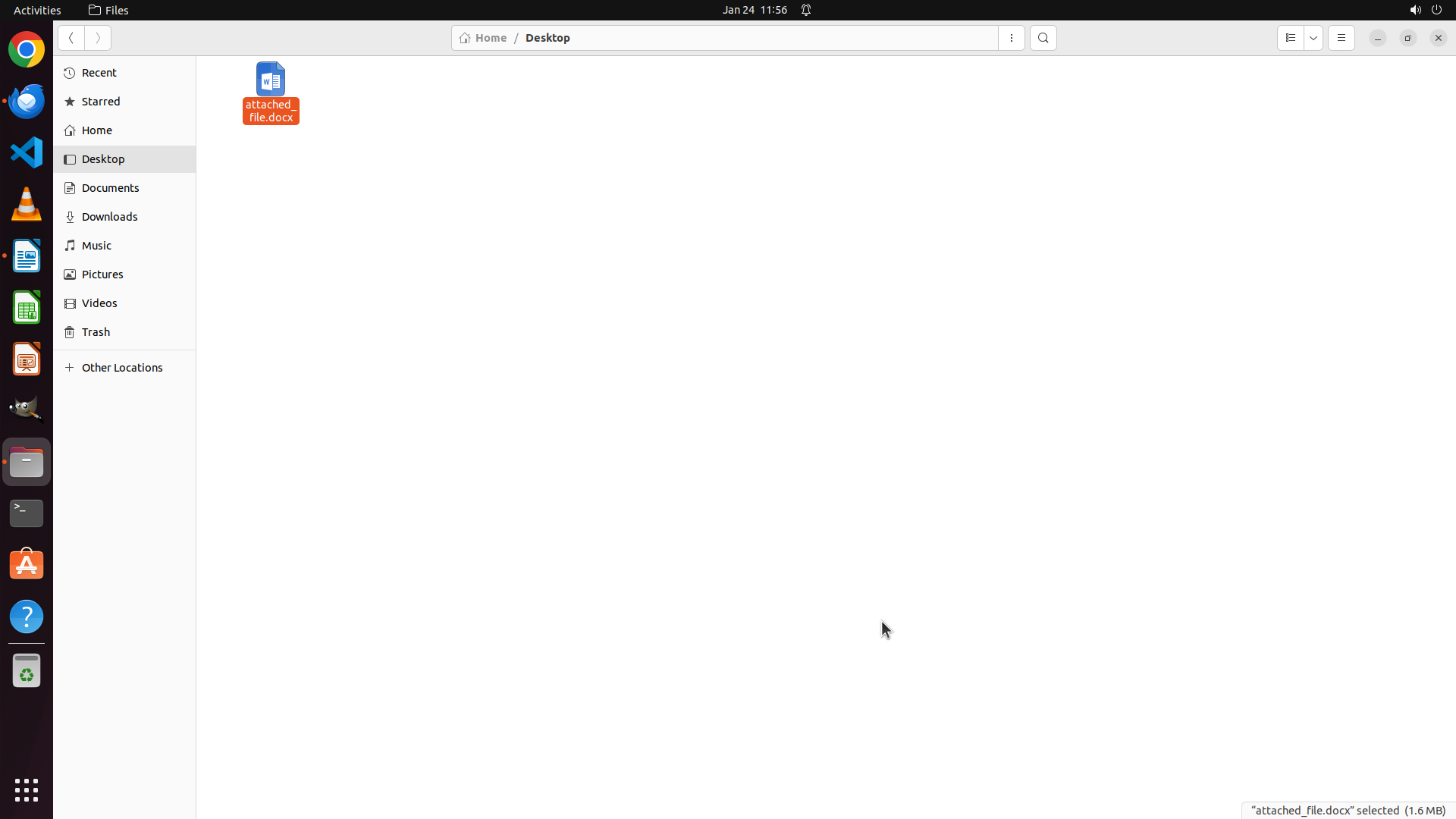}
    \\ \centering \texttt{hotkey(win)}
  \end{subfigure}%
 \hfill   \begin{subfigure}[t]{0.19\linewidth}
    \includegraphics[width=\linewidth]{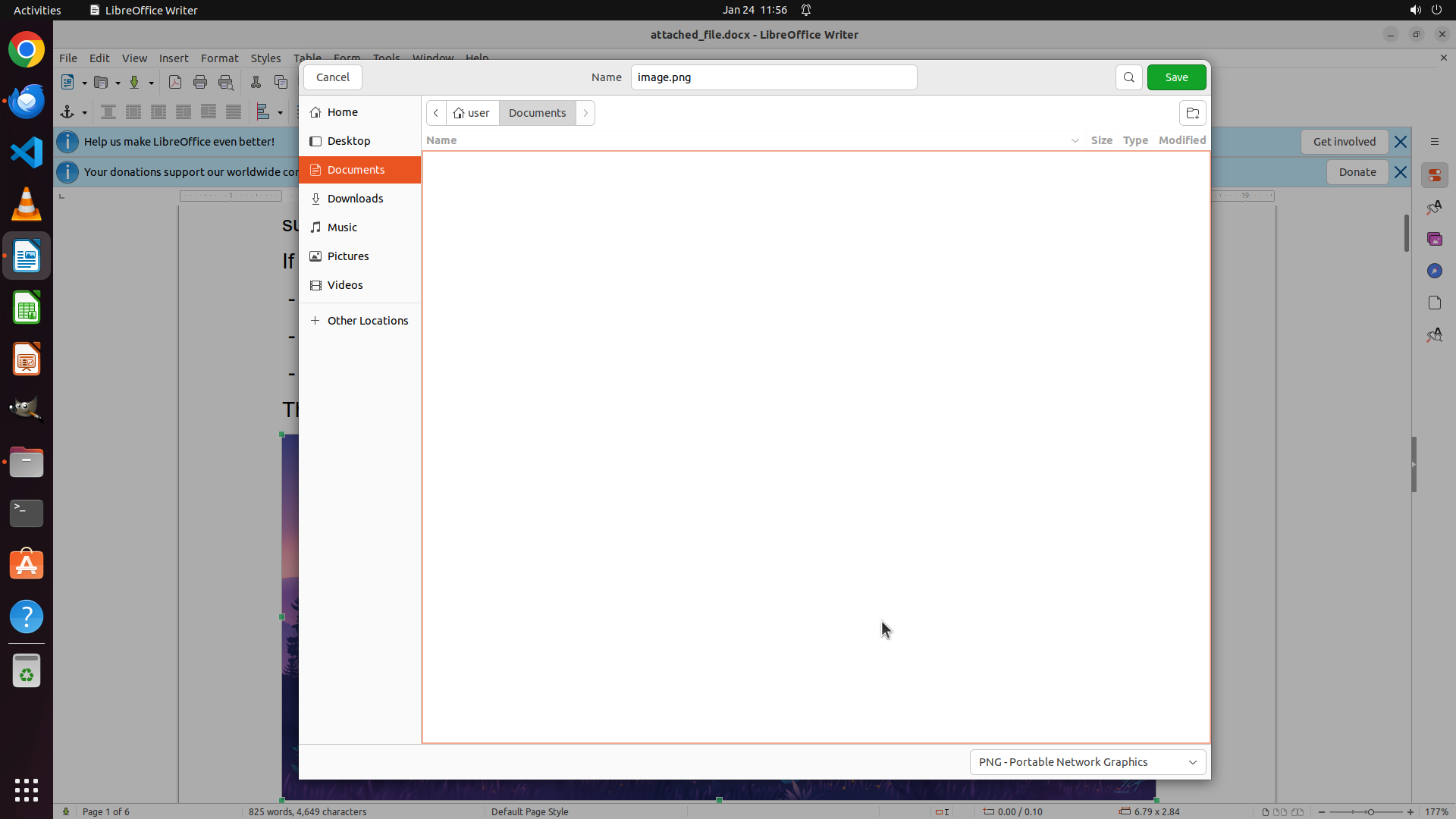}
    \\ \centering \texttt{press(escape)}
  \end{subfigure}%
 \hfill   \begin{subfigure}[t]{0.19\linewidth}
    \includegraphics[width=\linewidth]{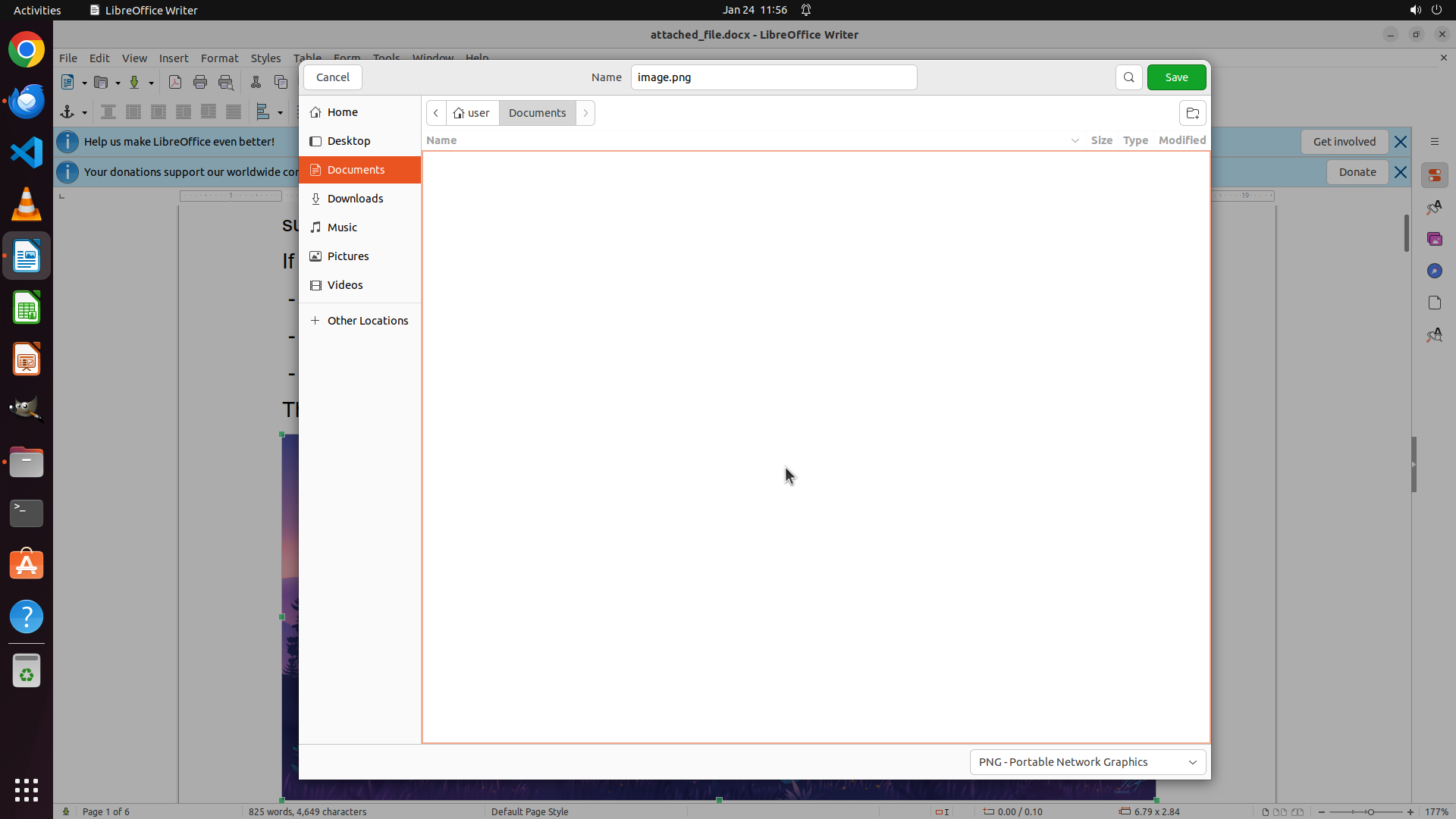}
    \\ \centering \texttt{scroll(10)}
  \end{subfigure}%

  \par\vspace{1em}
  \textbf{Node 1.1.4.5.3.1} \\
  \begin{subfigure}[t]{0.19\linewidth}
    \includegraphics[width=\linewidth]{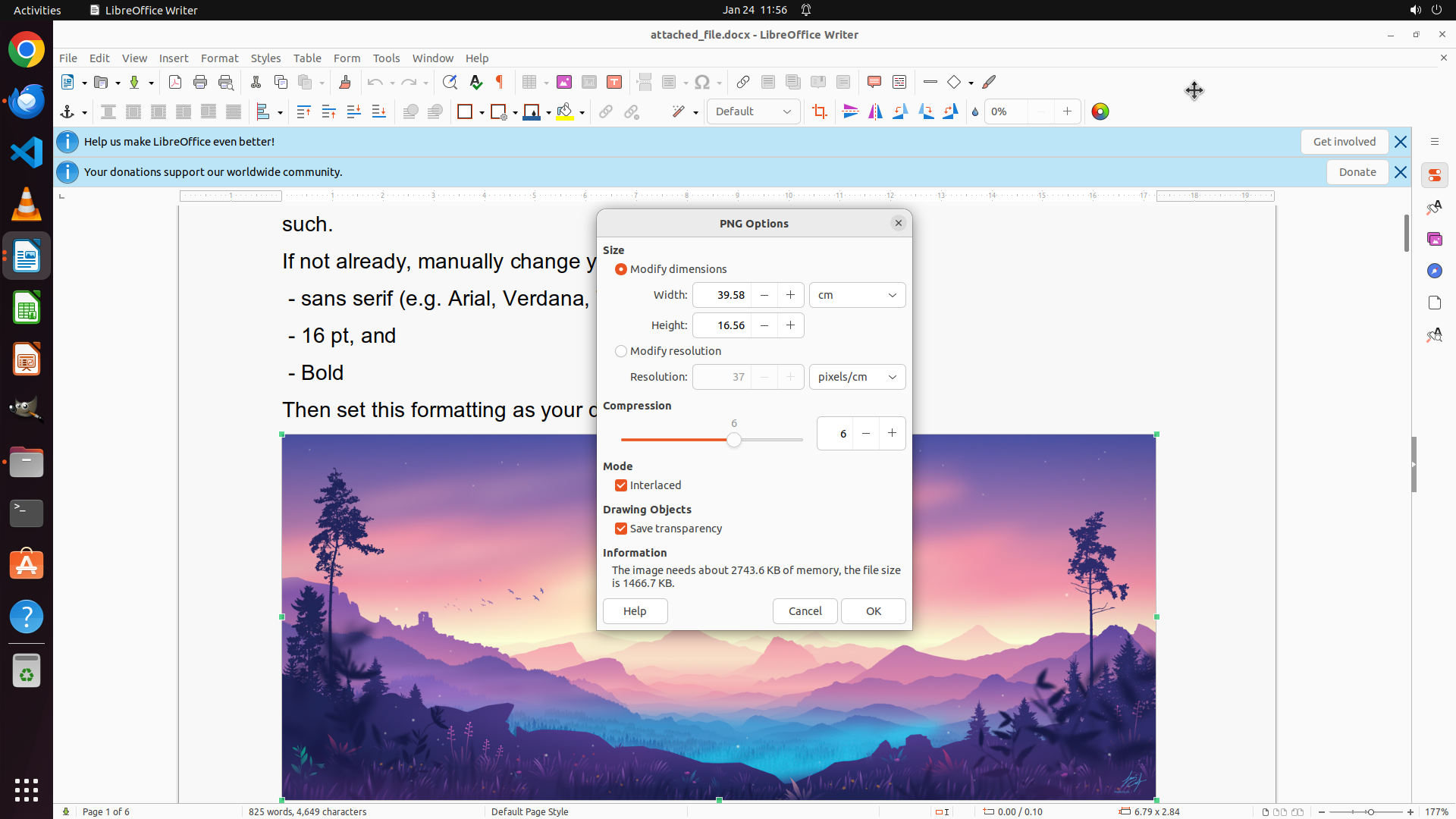}
    \\ \centering \texttt{click(1559, 103)}
  \end{subfigure}%
 \hfill   \begin{subfigure}[t]{0.19\linewidth}
    \includegraphics[width=\linewidth]{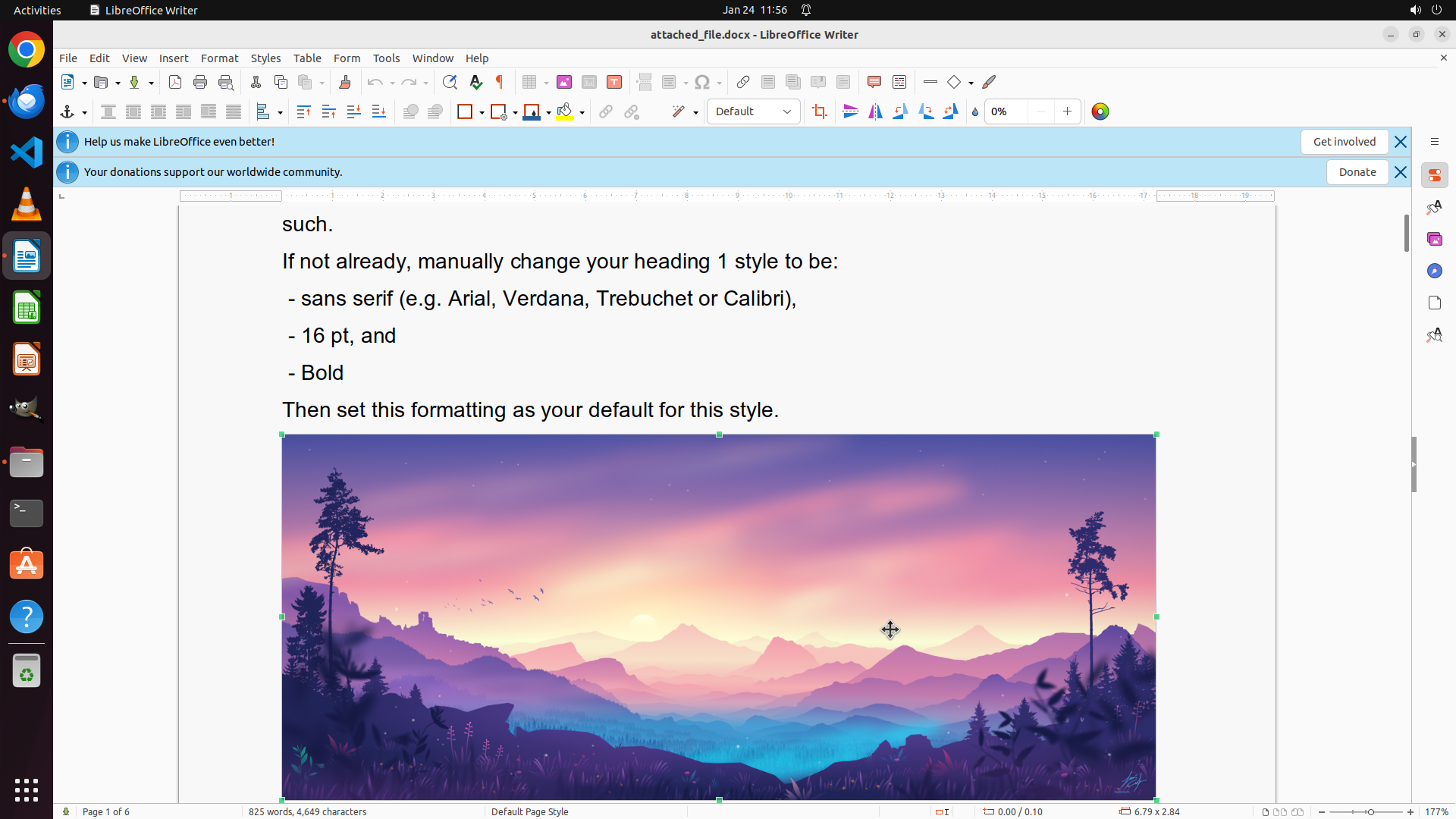}
    \\ \centering \texttt{click(1158, 814)}
  \end{subfigure}%
 \hfill   \begin{subfigure}[t]{0.19\linewidth}
    \includegraphics[width=\linewidth]{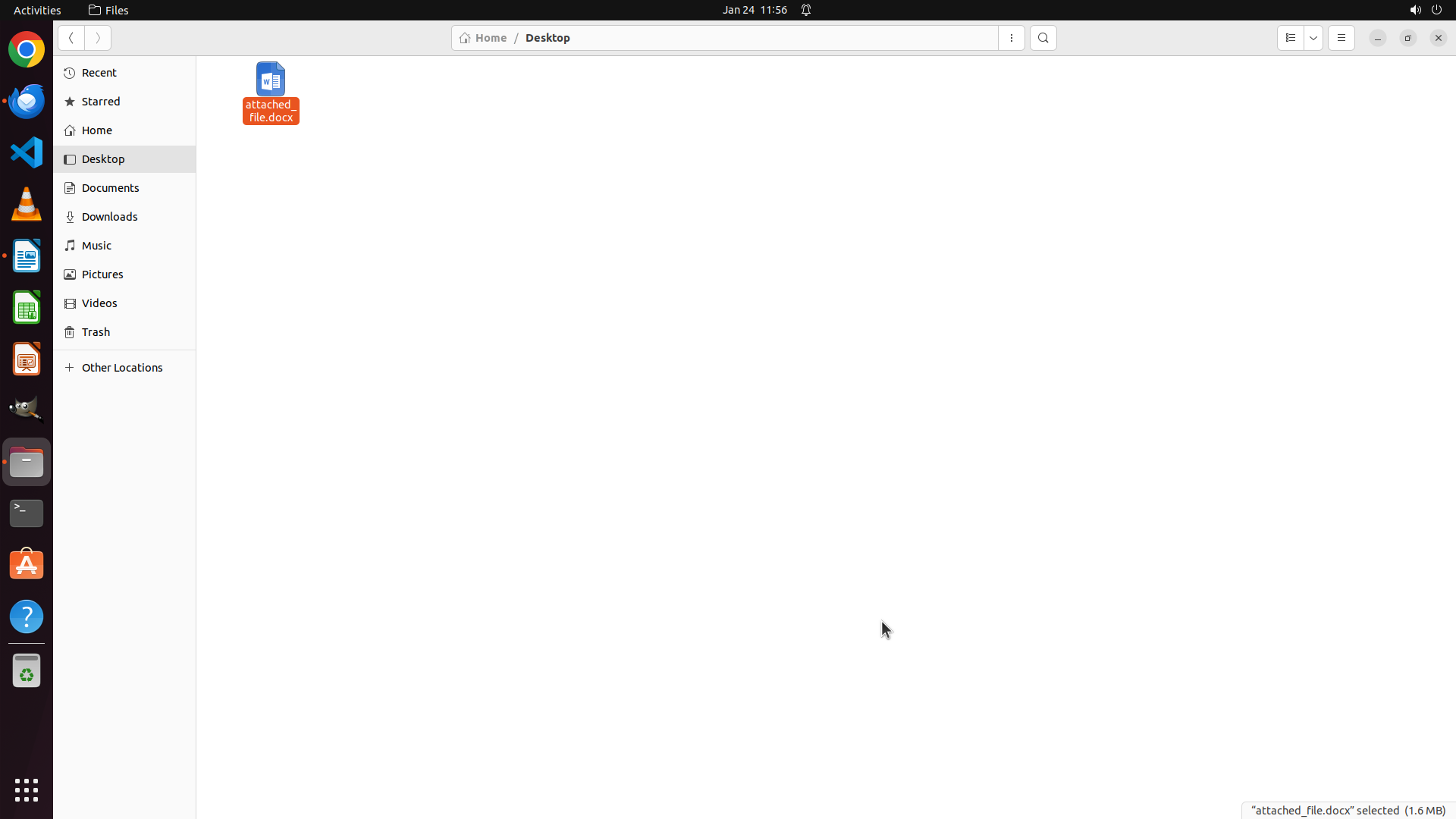}
    \\ \centering \texttt{hotkey(win)}
  \end{subfigure}%
 \hfill   \begin{subfigure}[t]{0.19\linewidth}
    \includegraphics[width=\linewidth]{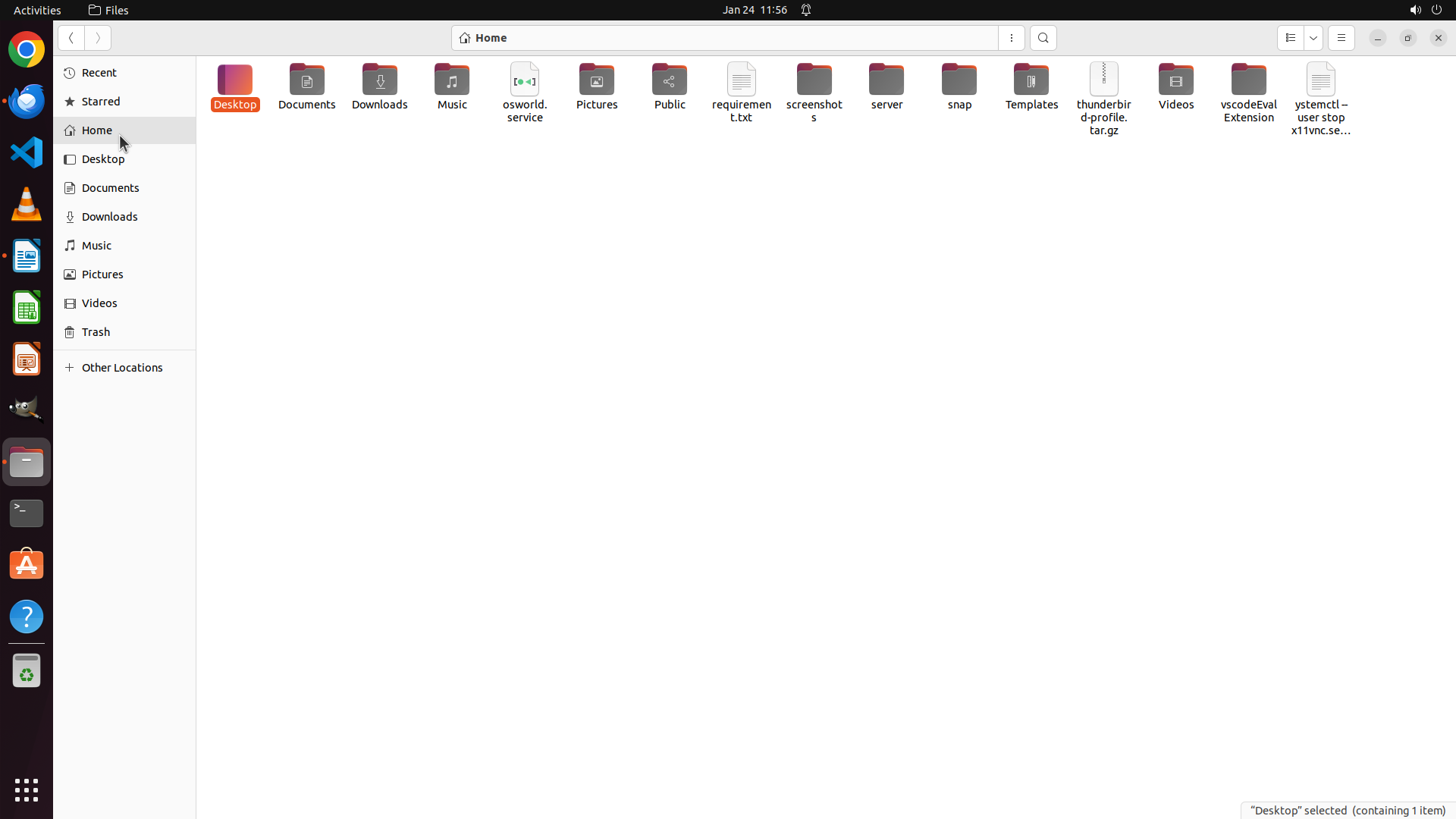}
    \\ \centering \texttt{click(153, 173)}
  \end{subfigure}%
 \hfill   \begin{subfigure}[t]{0.19\linewidth}
    \includegraphics[width=\linewidth]{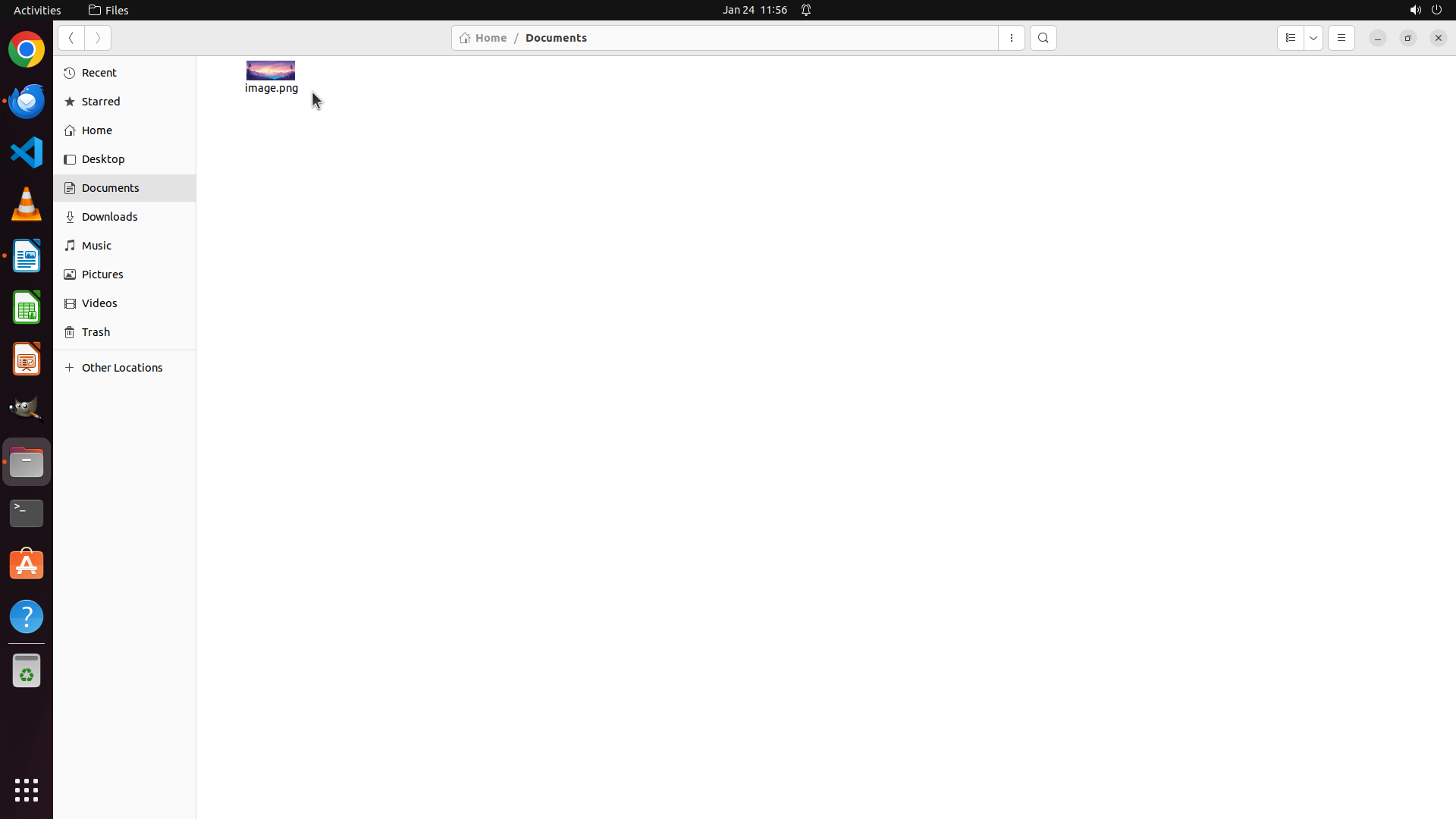}
    \\ \centering \texttt{click(407, 116)}
  \end{subfigure}%

  \par\vspace{1em}
  \textbf{Node 1.1.4.5.3.1.3} \\
  \begin{subfigure}[t]{0.19\linewidth}
    \includegraphics[width=\linewidth]{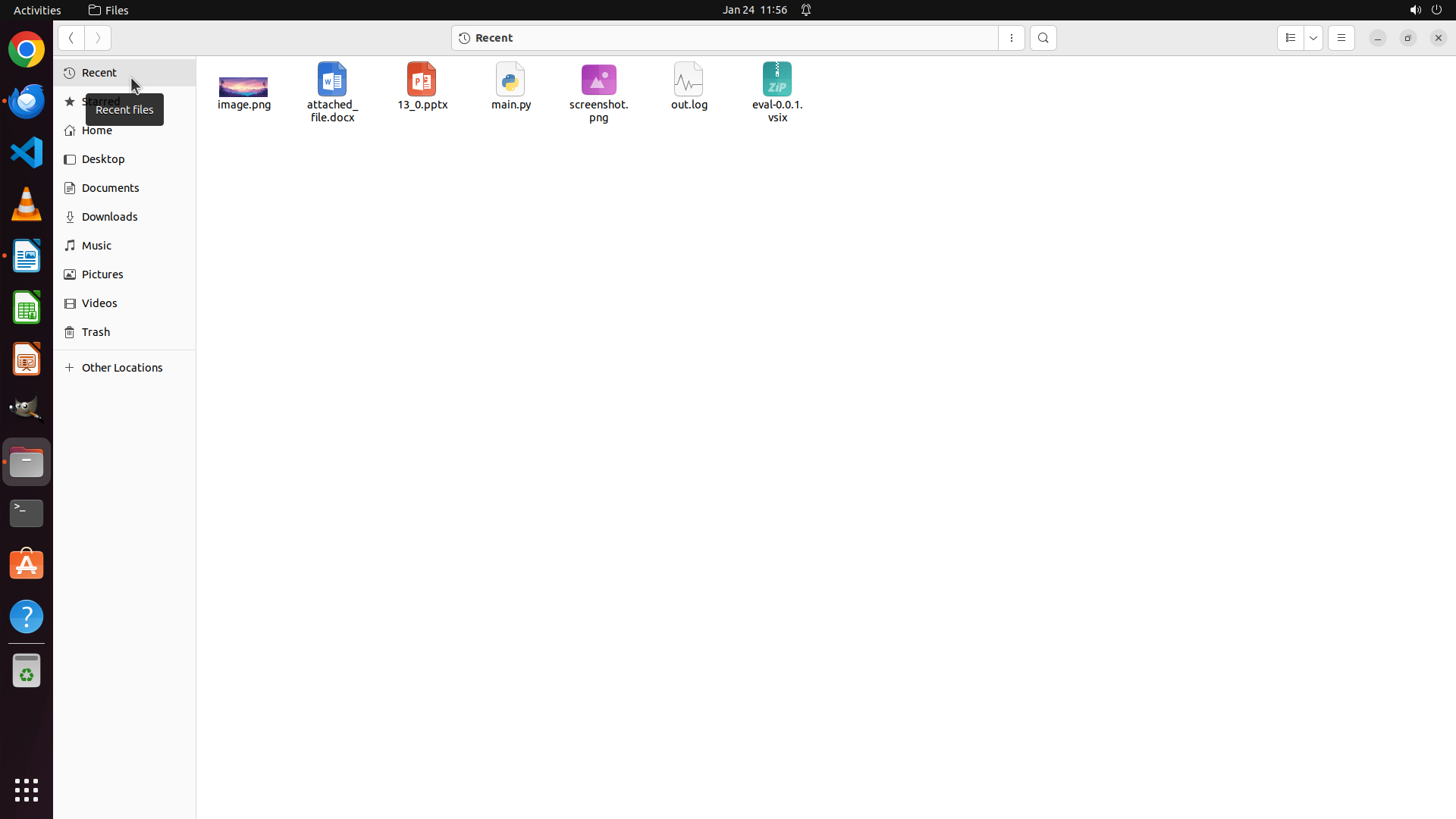}
    \\ \centering \texttt{click(168, 97)}
  \end{subfigure}%
 \hfill   \begin{subfigure}[t]{0.19\linewidth}
    \includegraphics[width=\linewidth]{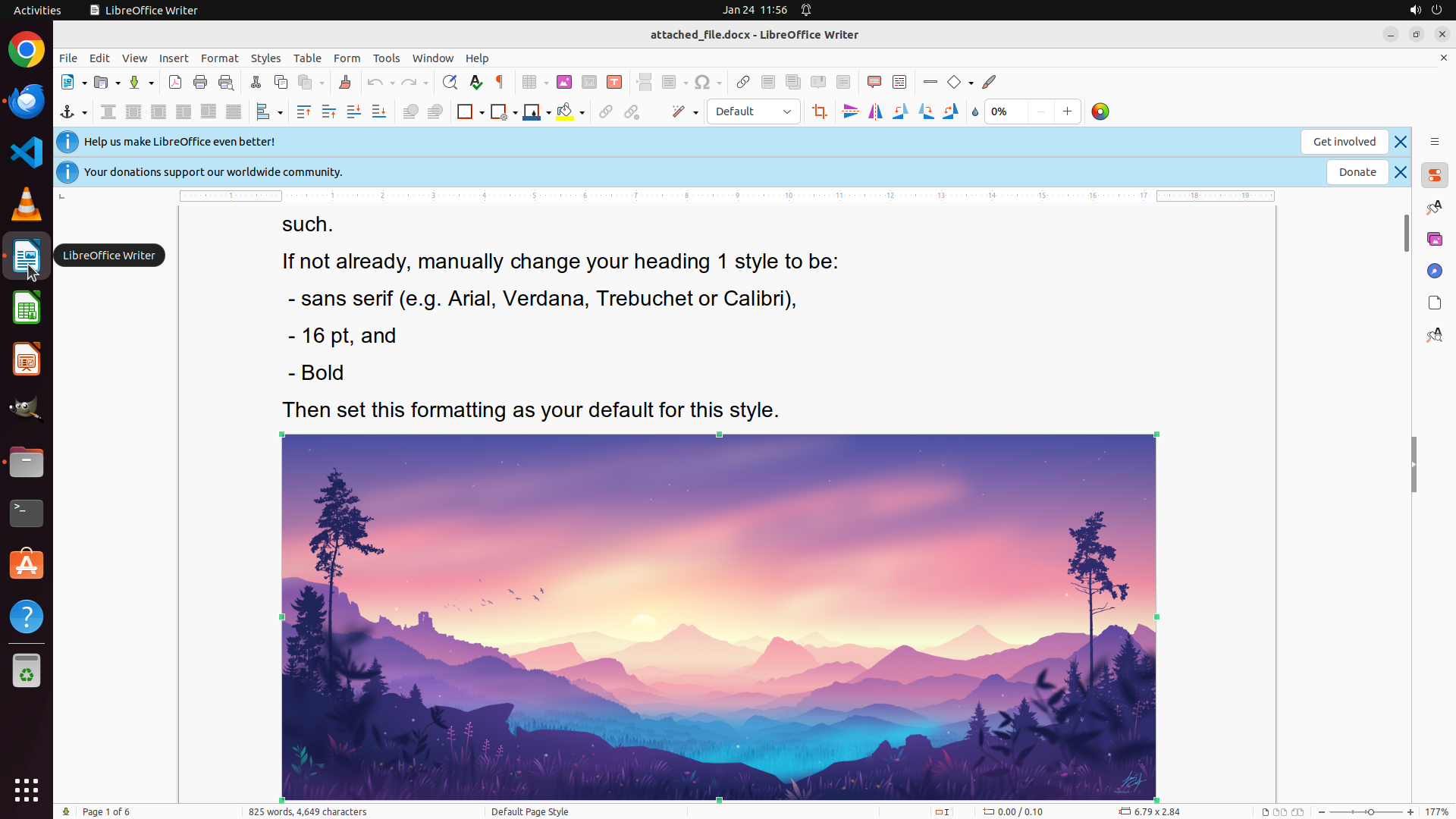}
    \\ \centering \texttt{click(32, 343)}
  \end{subfigure}%
 \hfill   \begin{subfigure}[t]{0.19\linewidth}
    \includegraphics[width=\linewidth]{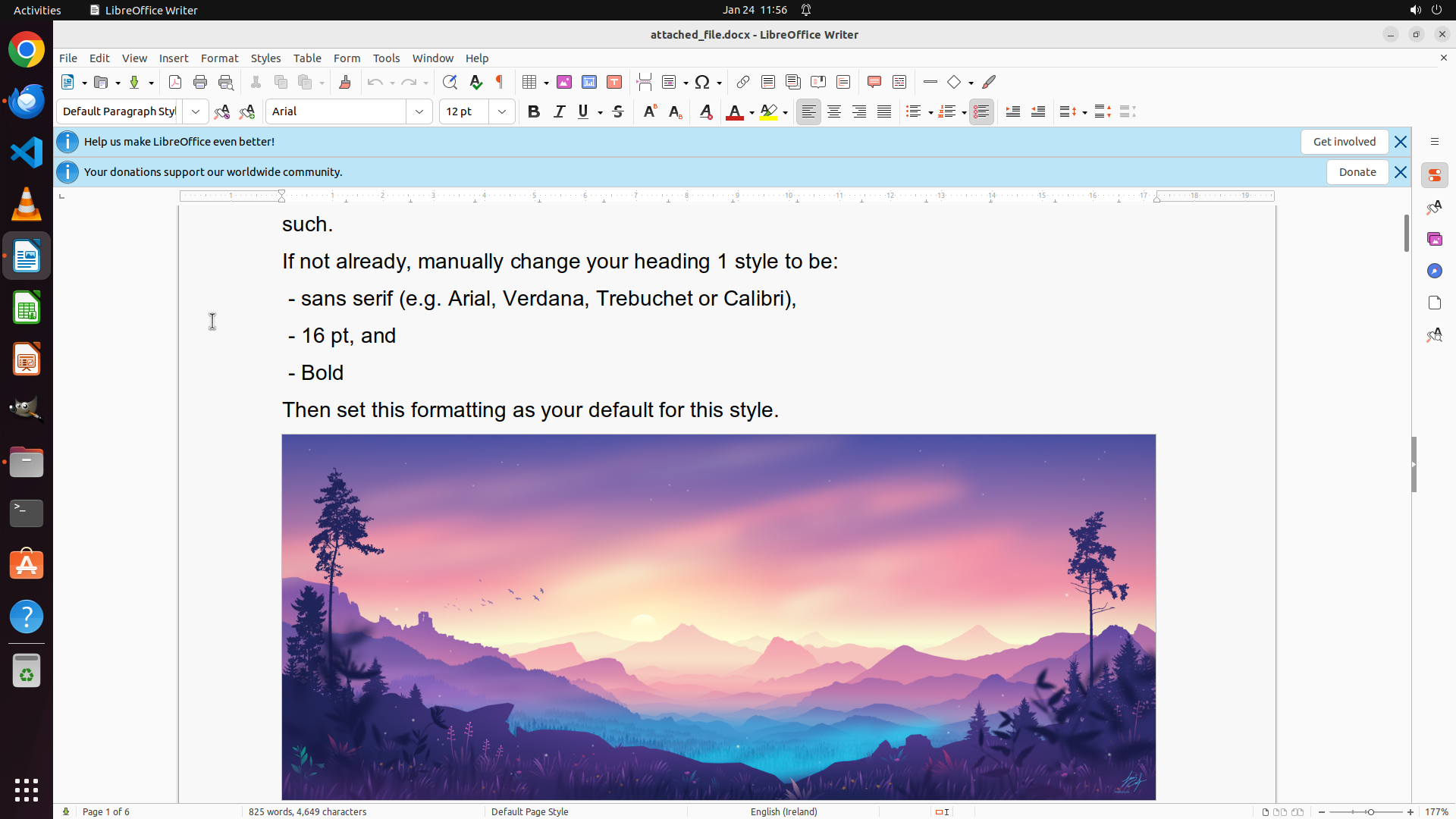}
    \\ \centering \texttt{click(266, 408)}
  \end{subfigure}%
 \hfill   \begin{subfigure}[t]{0.19\linewidth}
    \includegraphics[width=\linewidth]{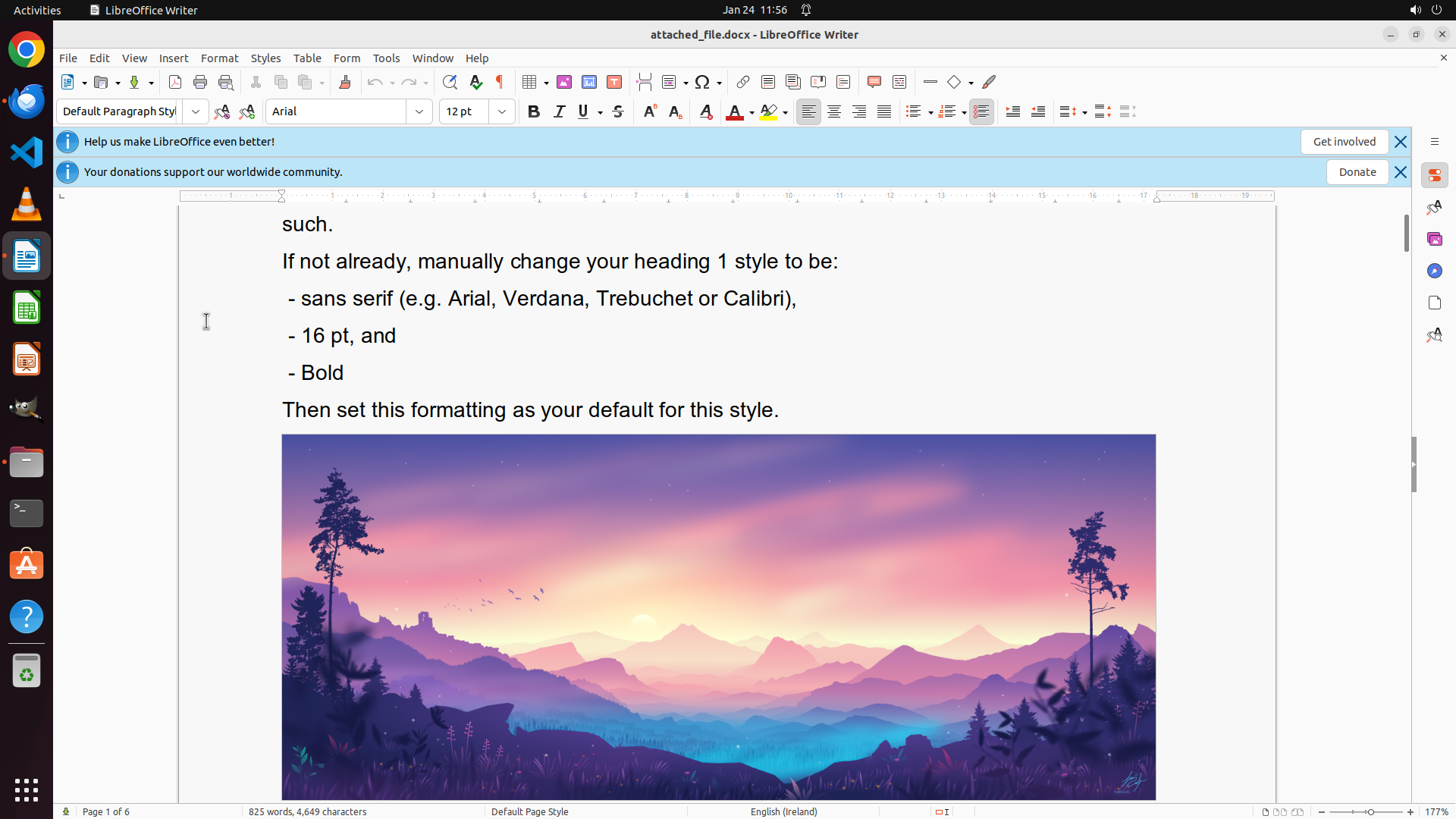}
    \\ \centering \texttt{click(258, 408)}
  \end{subfigure}%
 \hfill   \begin{subfigure}[t]{0.19\linewidth}
    \includegraphics[width=\linewidth]{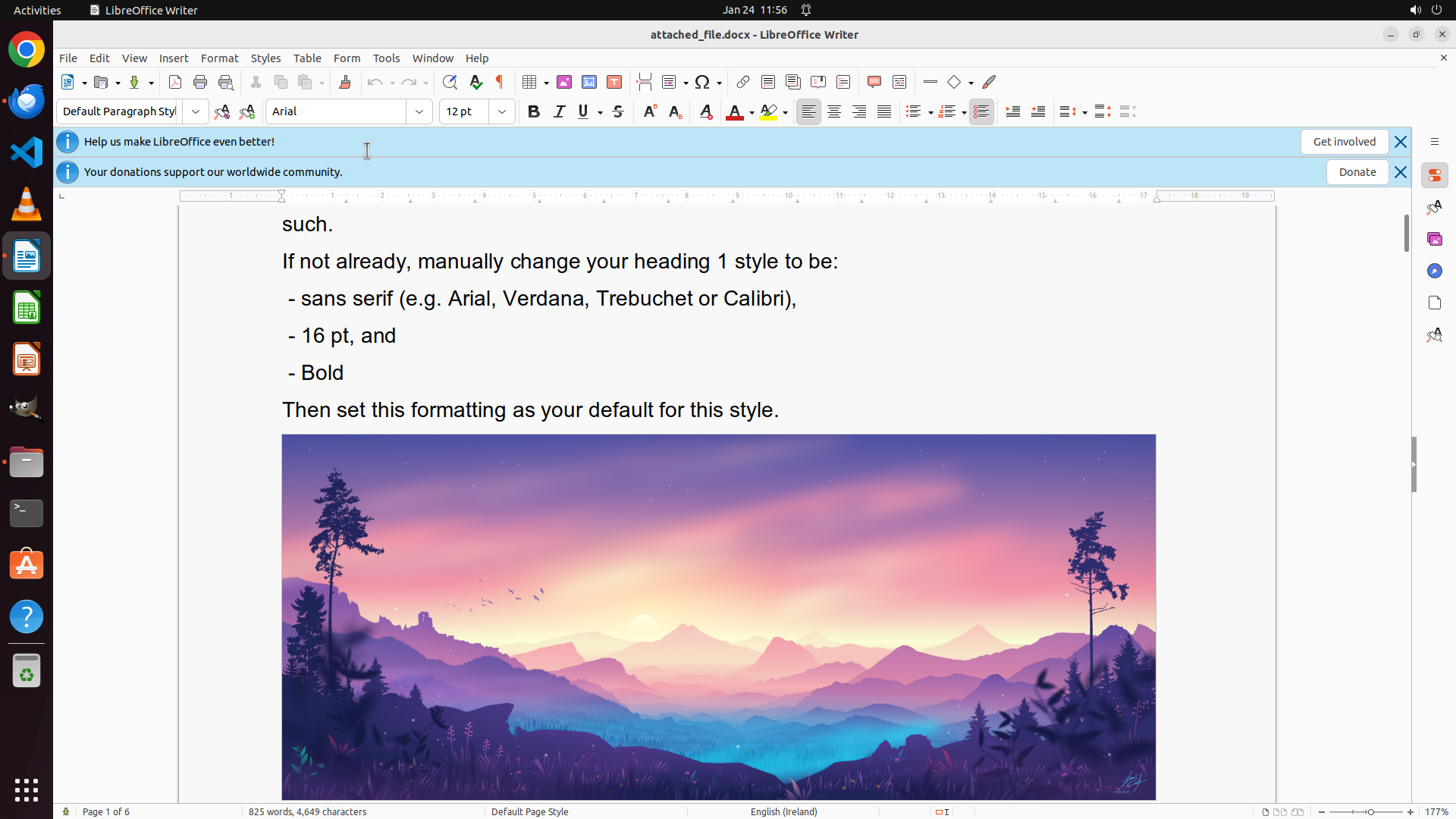}
    \\ \centering \texttt{click(470, 183)}
  \end{subfigure}%

  \caption{Trajectory Visualization of Task2: Part 1}
\end{figure}

\clearpage

\begin{figure}[htbp]
  \centering
  
  \textbf{Node 1.1.4.5.3.1.3.2} \\
  \begin{subfigure}[t]{0.19\linewidth}
    \includegraphics[width=\linewidth]{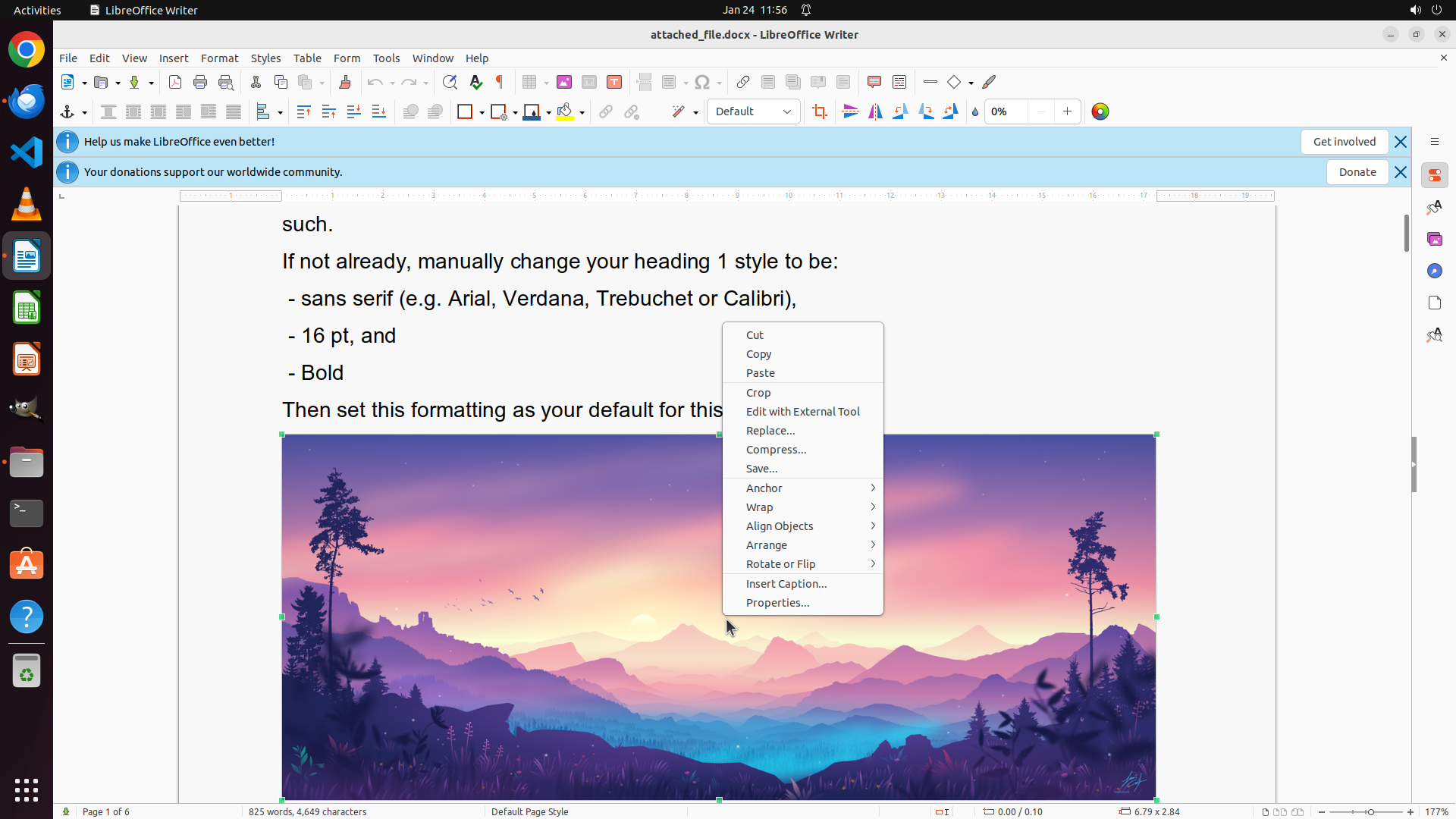}
    \\ \centering \texttt{click(953, 811)}
  \end{subfigure}%
 \hfill   \begin{subfigure}[t]{0.19\linewidth}
    \includegraphics[width=\linewidth]{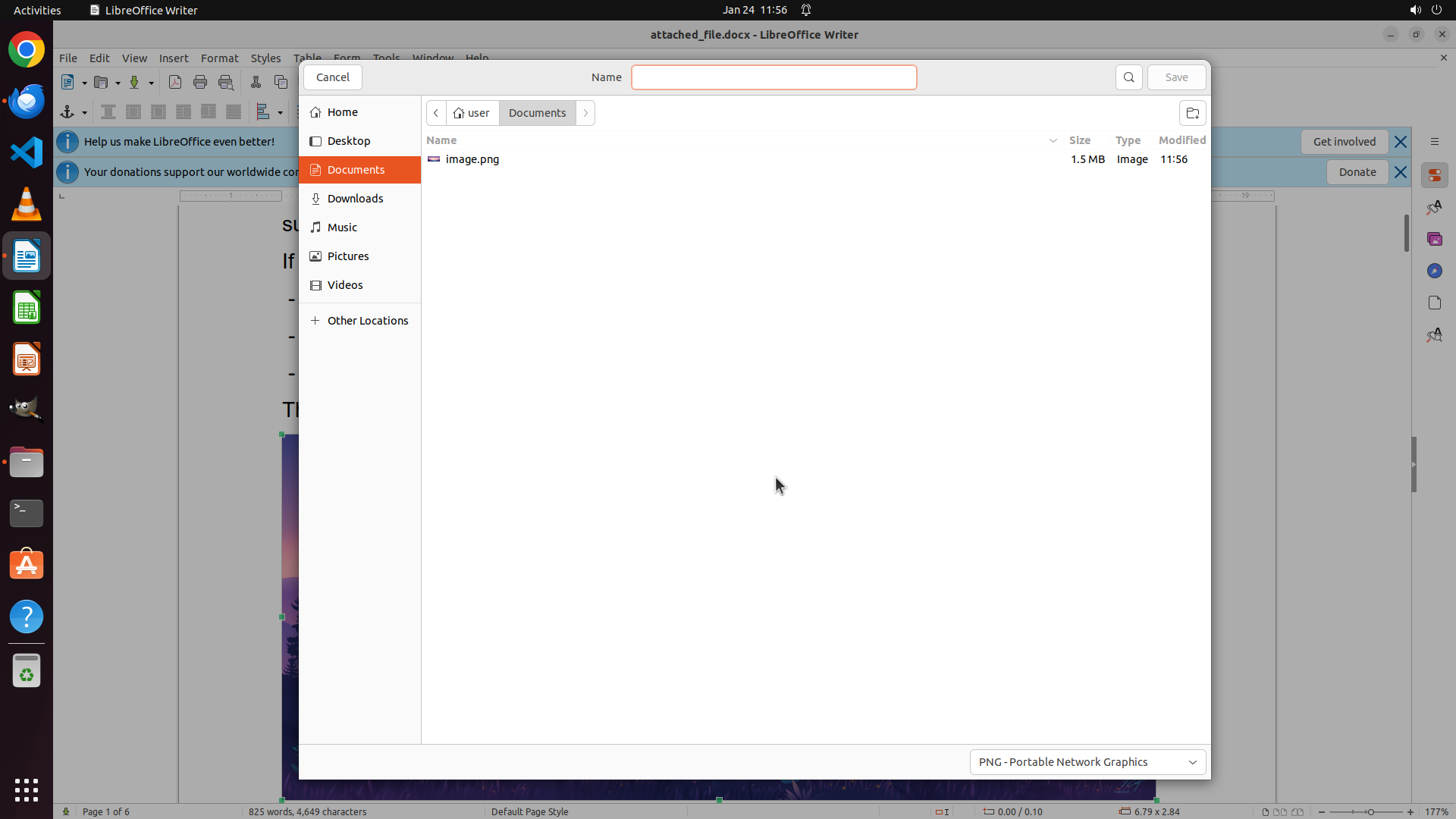}
    \\ \centering \texttt{click(1018, 624)}
  \end{subfigure}%
 \hfill   \begin{subfigure}[t]{0.19\linewidth}
    \includegraphics[width=\linewidth]{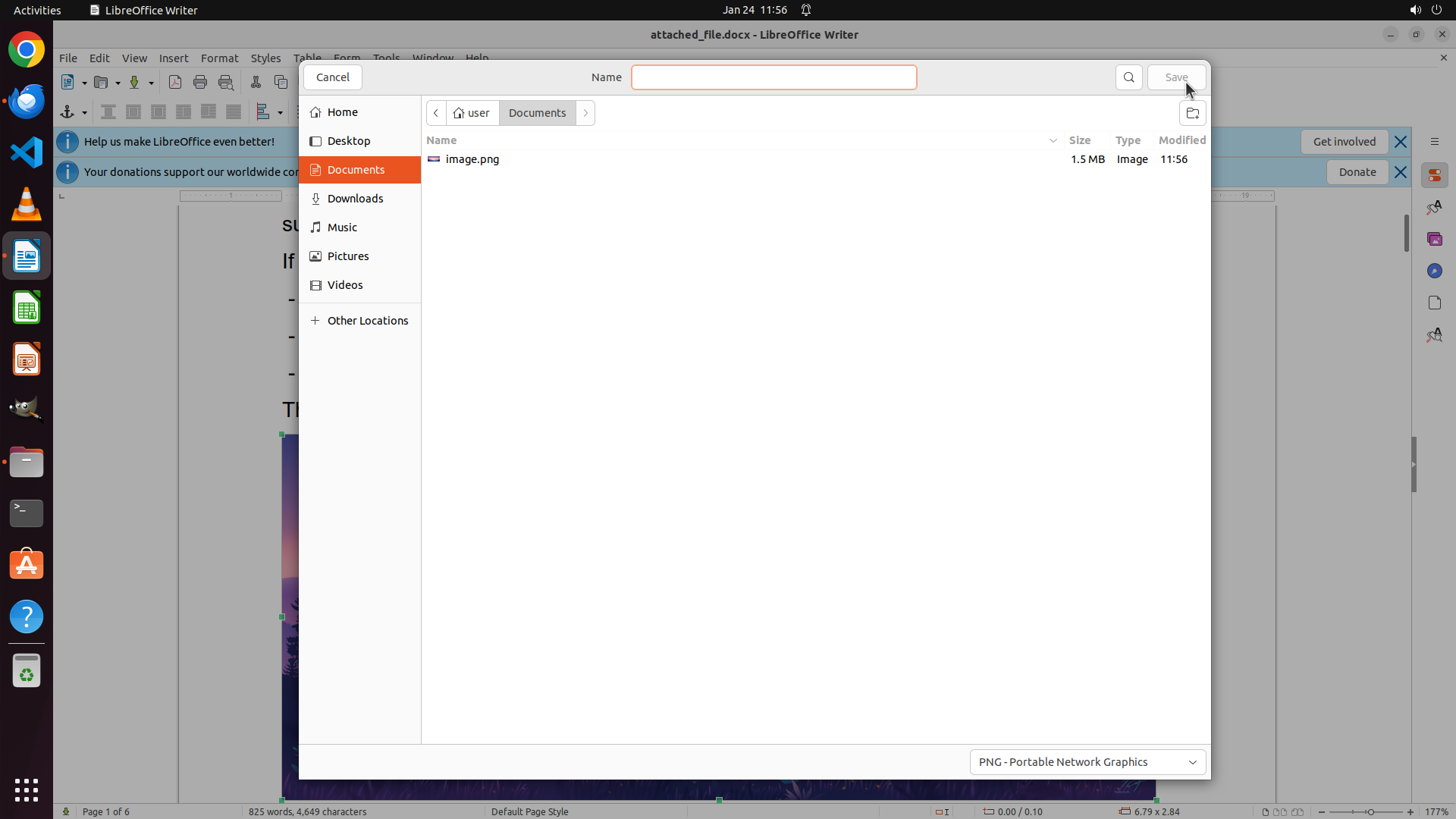}
    \\ \centering \texttt{click(1559, 103)}
  \end{subfigure}%
 \hfill   \begin{subfigure}[t]{0.19\linewidth}
    \includegraphics[width=\linewidth]{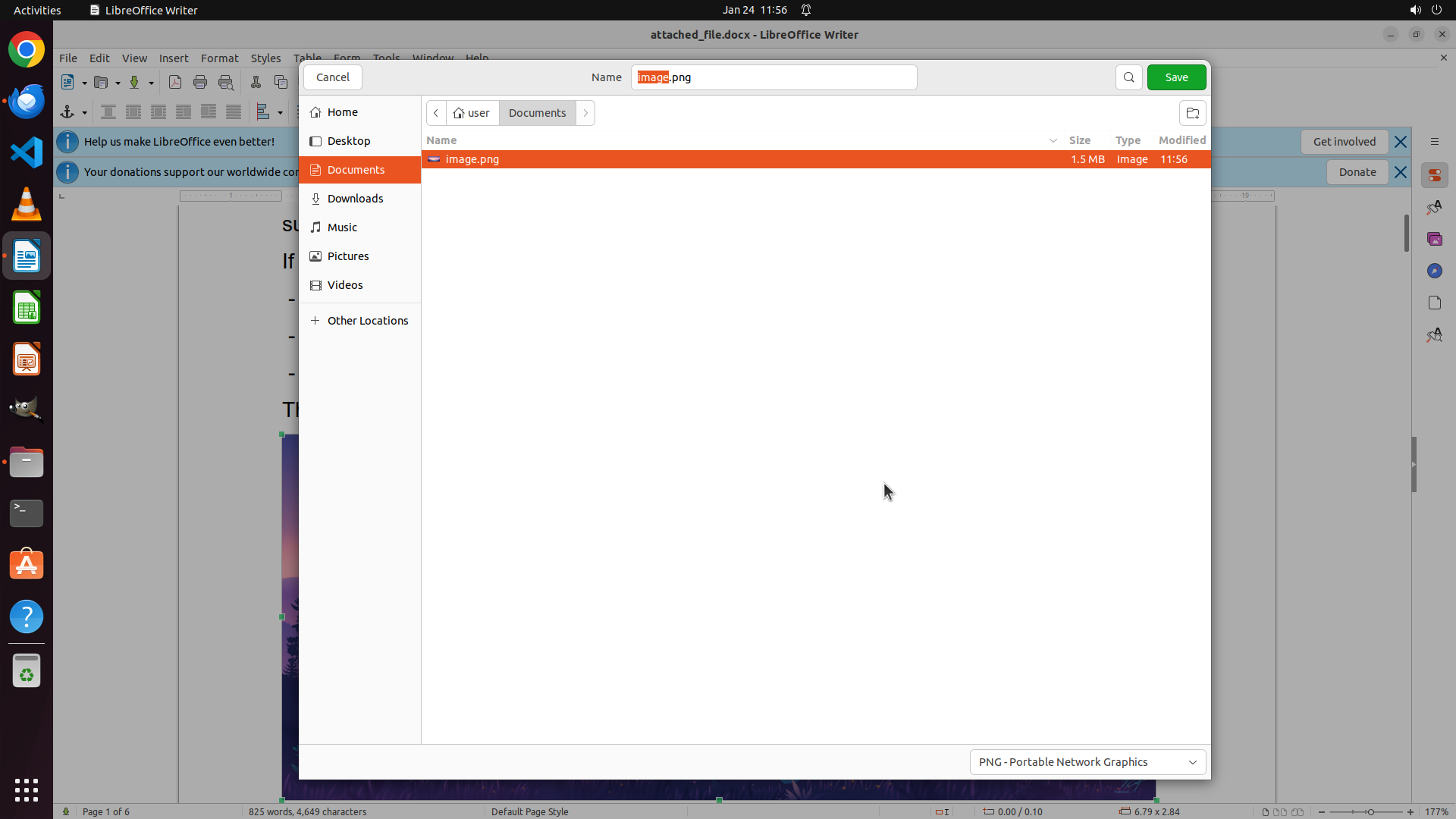}
    \\ \centering \texttt{click(1161, 632)}
  \end{subfigure}%
 \hfill   \begin{subfigure}[t]{0.19\linewidth}
    \includegraphics[width=\linewidth]{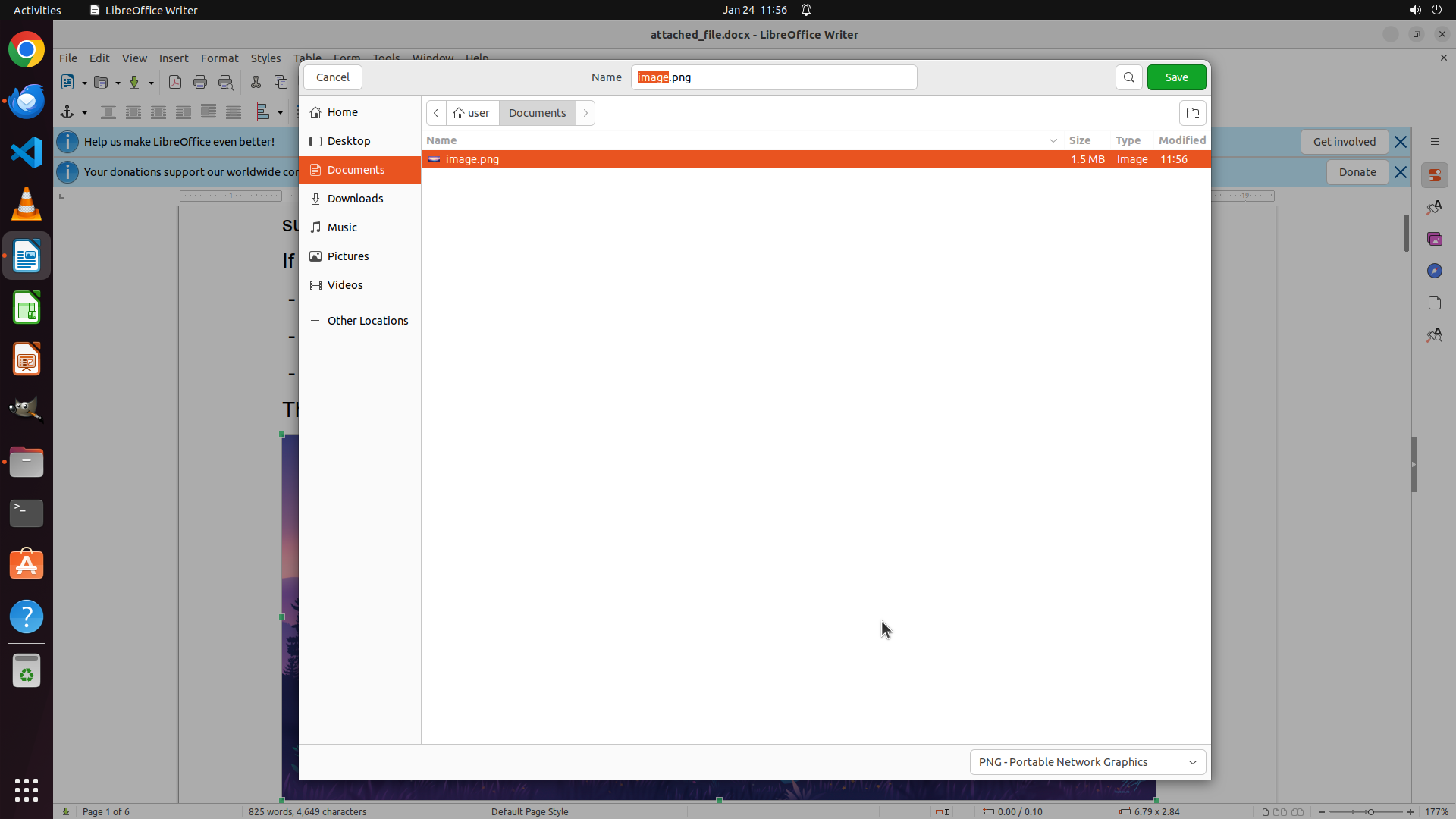}
    \\ \centering \texttt{click(1158, 814)}
  \end{subfigure}%

  \par\vspace{1em}
  \textbf{Node 1.1.4.5.3.1.3.2.1} \\
  \begin{subfigure}[t]{0.19\linewidth}
    \includegraphics[width=\linewidth]{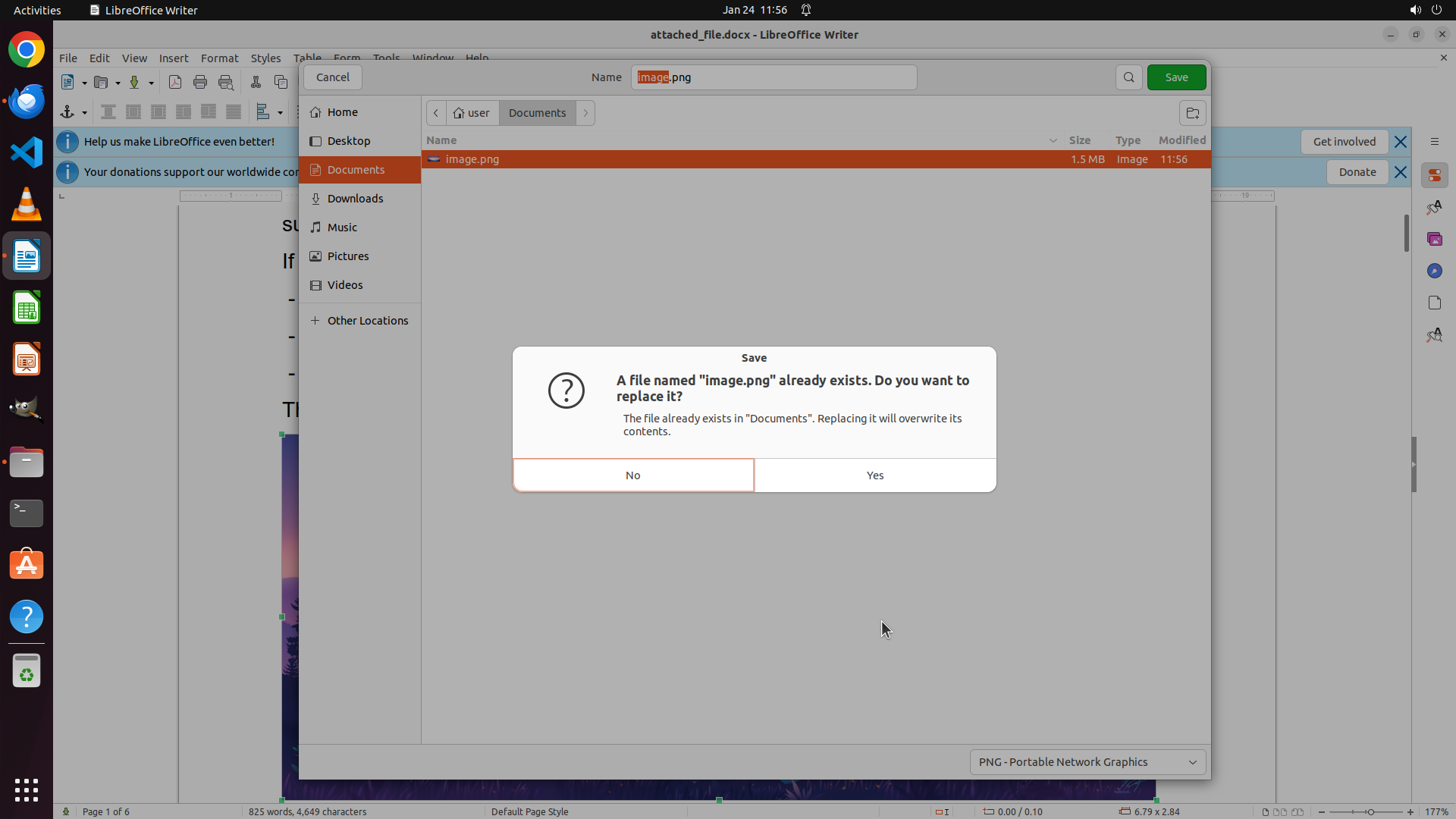}
    \\ \centering \texttt{hotkey(enter)}
  \end{subfigure}%
 \hfill   \begin{subfigure}[t]{0.19\linewidth}
    \includegraphics[width=\linewidth]{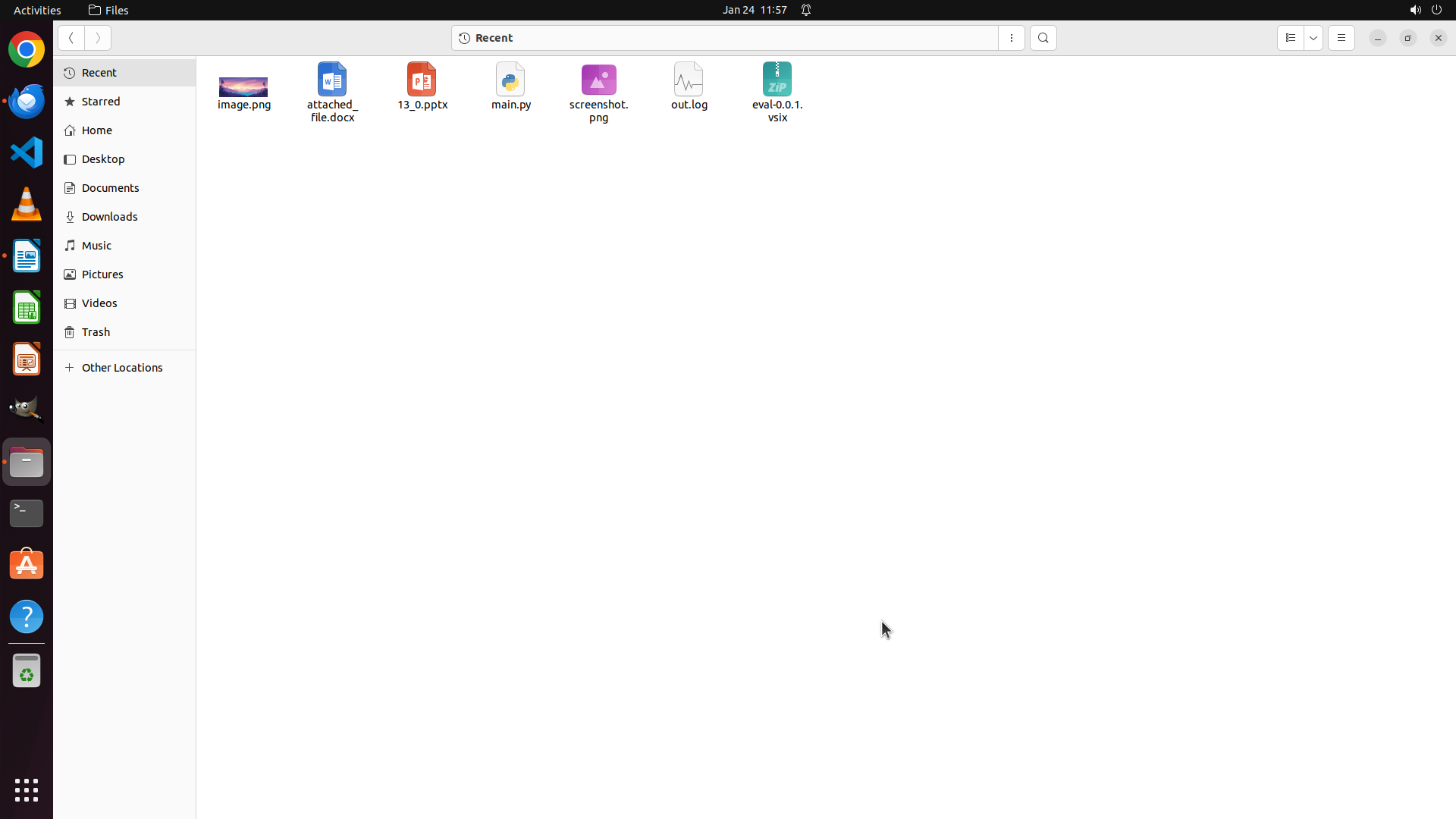}
    \\ \centering \texttt{hotkey(win)}
  \end{subfigure}%
 \hfill   \begin{subfigure}[t]{0.19\linewidth}
    \includegraphics[width=\linewidth]{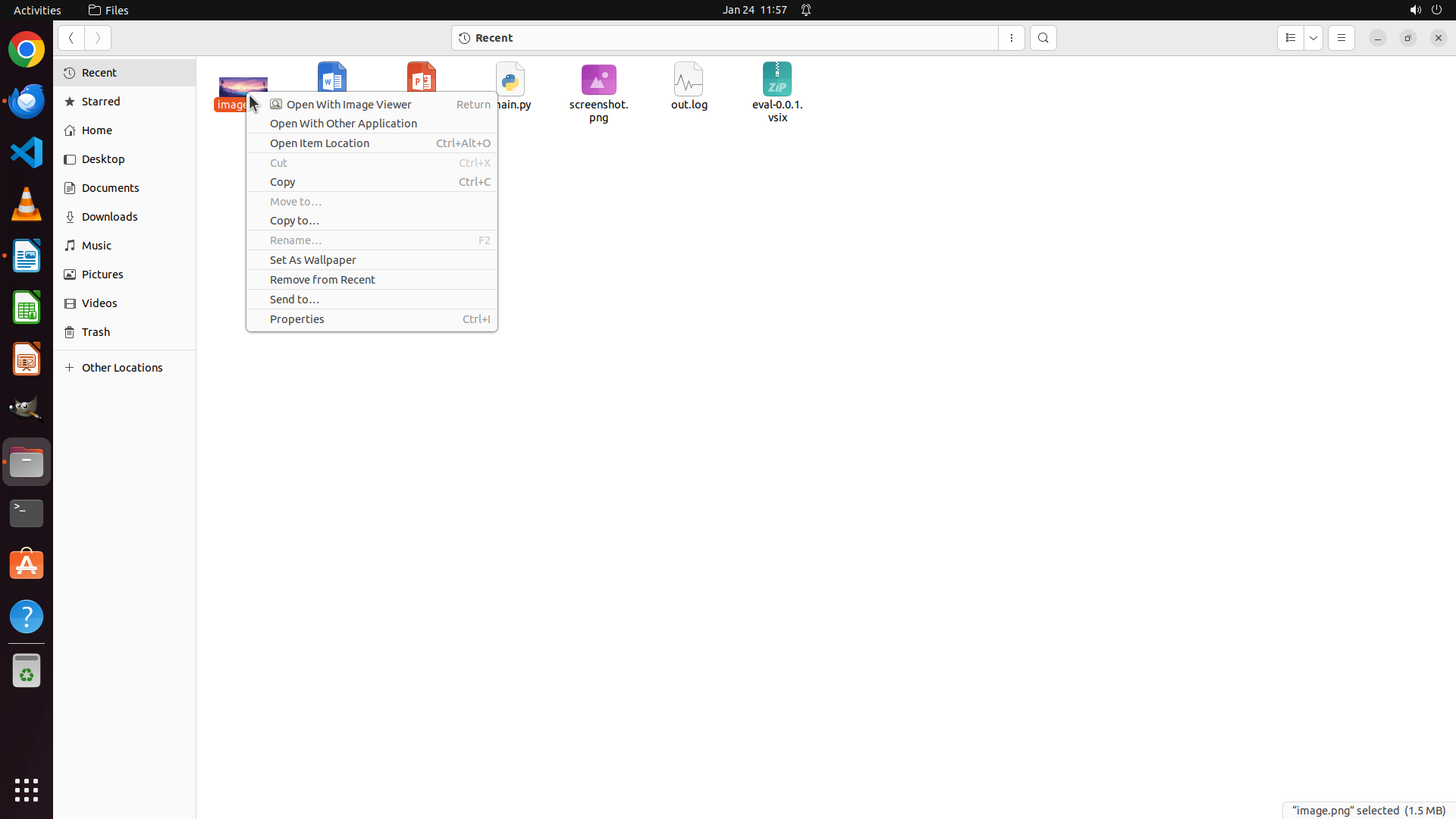}
    \\ \centering \texttt{click(324, 120)}
  \end{subfigure}%
 \hfill   \begin{subfigure}[t]{0.19\linewidth}
    \includegraphics[width=\linewidth]{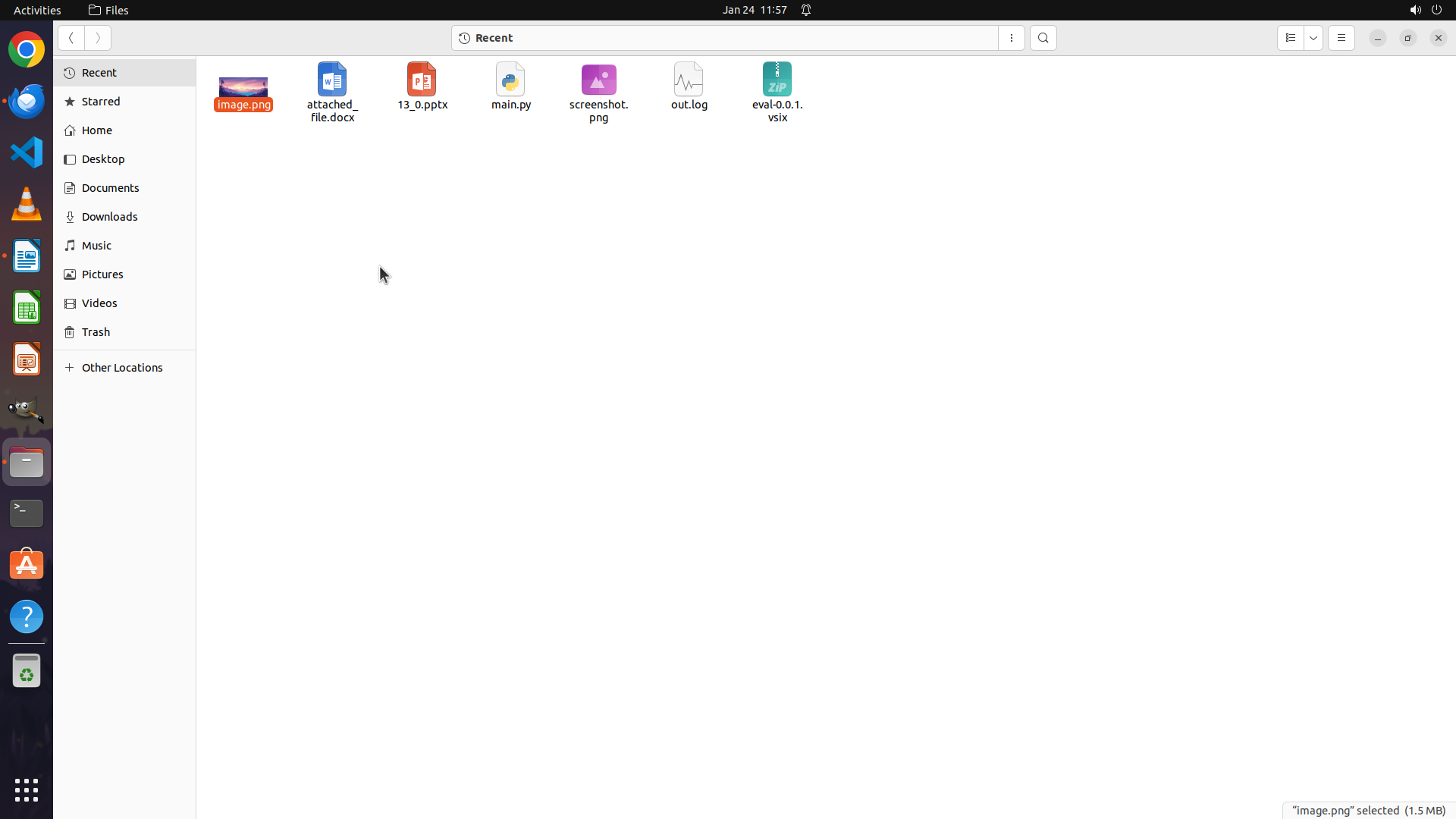}
    \\ \centering \texttt{click(496, 346)}
  \end{subfigure}%
 \hfill   \begin{subfigure}[t]{0.19\linewidth}
    \includegraphics[width=\linewidth]{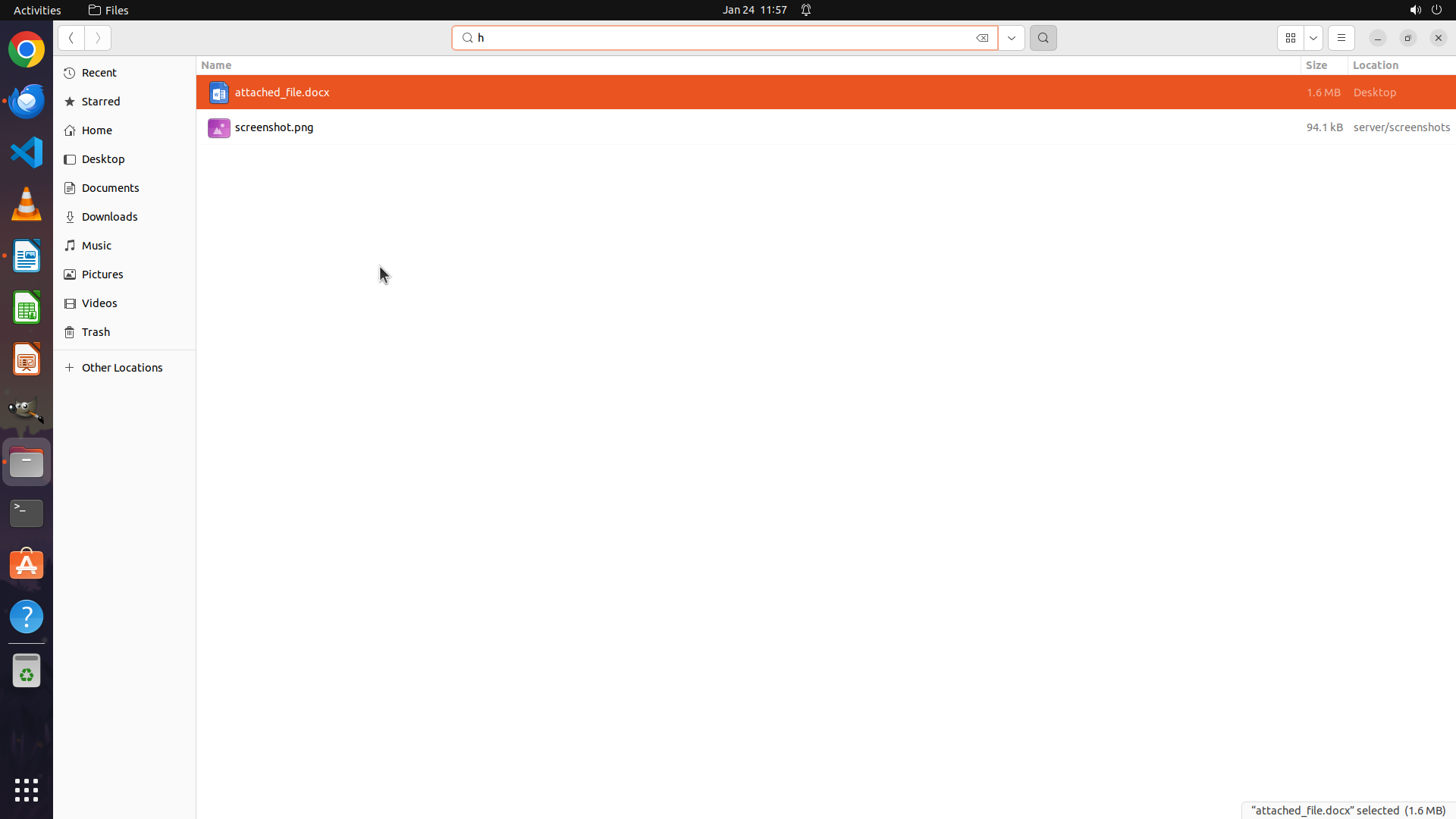}
    \\ \centering \texttt{hotkey(super,h)}
  \end{subfigure}%

  \caption{Trajectory Visualization of Task2: Part 2}
\end{figure}

\clearpage
\section{System Prompts}

In this section, we detail the comprehensive set of system prompts designed to govern the behavior and reasoning of our autonomous agents \cite{li2024crowdsensing, tang2025human, chang2024mixed, zhang2024modeling}. These prompts serve as the "cognitive framework" for the system~\cite{chen2025neurosymbolic,chen2025perception,li2026acdzero, tang2025malinzero, jiang2022intelligent}, encompassing various functional roles: from high-level task planning and dynamic reflection to stagnation detection and comparative evaluation. To ensure robust performance in complex GUI environments, we employ a modular prompt structure that includes specialized instructions for error recovery, tool-use selection (GUI vs. Code), and strict visual-based success criteria. The following figures illustrate the specific prompt templates used to guide the agents through trajectory analysis, multi-step planning, and final outcome verification.

\begin{figure}[h]
\centering
\small
\begin{tcolorbox}[colback=blue!5, colframe=blue!60!black, title=User Input, fonttitle=\bfseries]
\textbf{Instruction:} Can you help me change the default search engine from Chrome to Bing?
\end{tcolorbox}
\end{figure}

\begin{figure}[h]
\centering
\small
\begin{tcolorbox}[colback=orange!5, colframe=orange!90!black, title=Prompt for Reflection Generation, fonttitle=\bfseries]
\textbf{REFLECTION ON TRAJECTORY:} You are an expert computer use agent designed to reflect on the trajectory of a task and provide feedback on what has happened so far.

    You have access to the Task Description and the Current Trajectory of another computer agent. The Current Trajectory is a sequence of a desktop image, chain-of-thought reasoning, and a desktop action for each time step. The last image is the screen's display after the last action.
    
    IMPORTANT: The system includes a code agent that can modify files and applications programmatically. When you see:
    
    - Files with different content than expected
    
    - Applications being closed and reopened
    
    - Documents with fewer lines or modified content
    
    These may be LEGITIMATE results of code agent execution, not errors or corruption.
    
    Your task is to generate a reflection. Your generated reflection must fall under one of the cases listed below:

    Case 1. The trajectory is not going according to plan. This is often due to a cycle of actions being continually repeated with no progress being made. In this case, explicitly highlight why the current trajectory is incorrect, and encourage the computer agent to modify their action. However, DO NOT encourage a specific action in particular.
    
    Case 2. The trajectory is going according to plan. In this case, simply tell the agent to continue proceeding as planned. DO NOT encourage a specific action in particular.
    
    Case 3. You believe the current task has been completed. In this case, tell the agent that the task has been successfully completed.
    
    To be successful, you must follow the rules below:
    
    - **Your output MUST be based on one of the case options above**.
    
    - DO NOT suggest any specific future plans or actions. Your only goal is to provide a reflection, not an actual plan or action.
    
    - Any response that falls under Case 1 should explain why the trajectory is not going according to plan. You should especially lookout for cycles of actions that are continually repeated with no progress.
    
    - Any response that falls under Case 2 should be concise, since you just need to affirm the agent to continue with the current trajectory.
    
    - IMPORTANT: Do not assume file modifications or application restarts are errors - they may be legitimate code agent actions
    
    - Consider whether observed changes align with the task requirements before determining if the trajectory is off-track
\end{tcolorbox}
\end{figure}

\begin{figure}[h]
\centering
\small
\begin{tcolorbox}[colback=orange!5, colframe=green!60!black, title=Main prompt for Planning Generation, fonttitle=\bfseries]
 You are an expert in graphical user interfaces and Python code. You are responsible for executing the task: `TASK\_DESCRIPTION`.
You are working in CURRENT\_OS.

\# GUIDELINES

\#\# Agent Usage Guidelines

You have access to both GUI and code agents. Choose the appropriate agent based on the task requirements:

\#\#\# GUI Agent

- **Use for**: clicking, typing, navigation, file operations, tasks requiring specific application features...

\#\#\# Code Agent

{code\_agent\_section}  

\#\#\# Visual Confirmation for Evaluation

The evaluator judges success based on what is **VISIBLY** shown on the screen in the final screenshot.

- If you modified an image, you MUST open it in an image viewer to show the result.

- **Invisible changes count as failures.**

\#\#\# Environment-Instruction Alignment Analysis

Before planning your first action, analyze the relationship between the User Instruction and the Current Environment State.

1. **Initial State Check**: What application or website is currently open?

2. **Constraint Prediction**: Do NOT navigate away from the initial website unless explicitly asked.

3. **Action Alignment**: Use the search bar *of the current website*, not the browser's address bar.

\#\#\# Evaluation Awareness \& Success Criteria Analysis

The evaluator for this task is strict and programmatic. It often checks for specific URL parameters, file names, or HTML elements.

\#\#\# Pre-Completion Checklist

Before generating `agent.done()`, you MUST complete this checklist:

1. **Task Requirements Check**: Read the original task instruction word-for-word.

2. **Visual Proof Check**: Is the result clearly visible in the current screenshot?

3. **File Name Check**: If the task specified a filename, verify it EXACTLY matches.

4. **Core Task Verification**: Are you solving the RIGHT task?

\#\#\# State Stagnation \& Loop Detection (CRITICAL)

- **Compare Screenshots**: Look closely at the current screenshot versus the previous step's screenshot.

- **Identical State = Action Failed**: If the screen hasn't changed, your last action was **INEFFECTIVE**.

- **DO NOT REPEAT**: Never repeat an action that just failed.

\#\#\# END OF GUIDELINES

You are provided with:

1. A screenshot of the current time step.

2. The history of your previous interactions with the UI.

3. Access to the following class and methods to interact with the UI:

class Agent:
    {}

Your response should be formatted like this:

(Previous action verification)

Carefully analyze based on the screenshot if the previous action was successful.

(Screenshot Analysis)

Closely examine and describe the current state of the desktop.

(Next Action)

Based on the current screenshot and the history, decide on the next action in natural language.

(Grounded Action)

Translate the next action into code using the provided API methods:

```python

agent.click("The menu button at the top right of the window", 1, "left")
```
\end{tcolorbox}
\end{figure}

\begin{figure}[h]
\centering
\small
\begin{tcolorbox}[colback=orange!5, colframe=green!60!black, title=Dynamically added prompt for Planning Generation, fonttitle=\bfseries]

CRITICAL REFLECTION - YOU MUST READ AND ACT ON THIS 

\{reflection\}

MANDATORY INSTRUCTIONS BASED ON ABOVE REFLECTION:

1. If reflection says "NOT going according to plan" → You MUST try a DIFFERENT action

2. If reflection identifies a problem → You MUST address that specific problem

3. DO NOT repeat the same action if it has failed before

4. If clicking hasn't worked → Try scrolling, keyboard shortcuts, or different UI elements

5. Explore alternative paths: search boxes, filters, side panels, menus

6. Read the explanation(if available) of button or clickable element in the screenshot carefully.

7. Ignore options/settings that are NOT related to the instruction. Only focus on satisfying the specific conditions mentioned in the instruction.

8. If the task is about Chrome, DO NOT use the code agent. And do the task on initial webpage.

9. If you can't find the exact option/setting, try to find the similar or related option/setting that might reach the goal.

10. If the option list doesn't shows the target option in the first page, try to scroll the list to look for it.

\end{tcolorbox}
\end{figure}

\begin{figure}[h]
\centering
\small
\begin{tcolorbox}[colback=orange!5, colframe=green!60!black, title=Dynamically added prompt for Planning Generation(deep search alert), fonttitle=\bfseries]

DEEP SEARCH ALERT - CONSIDER ALTERNATIVE APPROACH

Current \{depth\_label\}: \{depth\_value\} (threshold exceeded: $>$ 8)
\{iter\_line\}

You have been exploring this path for a long time without success. This often indicates:

- The GUI approach is too complex or unreliable for this task

- The task might be better solved with code/command-line tools

- You are stuck in unproductive GUI navigation

MANDATORY ACTION REQUIRED:
If you have NOT already tried the code agent for this task, you MUST try it now:

1. **For application settings/config tasks**: Use agent.call\_code\_agent() to edit config files

2. **For file/data manipulation**: Use agent.call\_code\_agent() for text editing, file processing

3. **For extension installation**: Use code --install-extension instead of GUI navigation

4. **EXCEPTION FOR CHROME TASKS**: For Chrome-related tasks, DO NOT use the code agent.

DO NOT continue with GUI operations if coding agent could solve this more reliably.

\end{tcolorbox}
\end{figure}

\begin{figure}[h]
\centering
\small
\begin{tcolorbox}[colback=orange!5, colframe=green!60!black, title=Dynamically added prompt for Planning Generation(task infeasible alert), fonttitle=\bfseries]

CRITICAL: TASK MAY BE INFEASIBLE

Current \{depth\_label\}: \{depth\_value\} (CRITICAL THRESHOLD: 17)

 You have explored this task for an EXTREMELY long time without success.
This is a STRONG indicator that the task may be IMPOSSIBLE or INFEASIBLE.

CRITICAL ANALYSIS REQUIRED - Ask yourself:

1. **Application Capability Check**:

   - Does this application have the required feature?
   
     Examples: VS Code cannot open multiple workspaces in one window
              GIMP cannot edit/trim videos

2. **Fundamental Constraint Check**:

   - Is there a fundamental limitation preventing this task?
   
   - Have you tried multiple completely different approaches?

3. **Repeated Failure Pattern**:

   - Are you stuck in a cycle trying the same approaches?

IF YOU DETERMINE THE TASK MAYBE IMPOSSIBLE:
You MUST call agent.fail() immediately.

\end{tcolorbox}
\end{figure}

\begin{figure}[h]
\centering
\small
\begin{tcolorbox}[colback=orange!5, colframe=red!60!black, title=Prompt for Comparative Evaluation(fact generation), fonttitle=\bfseries]
\textbf{BEHAVIOR\_NARRATOR\_SYSTEM\_PROMPT:} You are an expert in computer usage responsible for analyzing what happened after a computer action is taken. 

**Reasoning Guidelines:**

You will analyze the before and after screenshots given an action and provide a clear summary of the changes observed. Some things to note:

- Pay attention to any circular visual markers that may suggest where clicks, mouse movements, or drags occurred.

  - Clicks will be marked with a red circle and labeled Click
  
  - Moving the mouse without clicking will be marked with a blue circle and labeled MoveTo
  
  - Drag and drops will have an initial blue circle labeled MoveTo, a green circle labeled DragTo, and a green line connecting the two circles.
  
- If any mouse action occurred, the after screenshot will be accompanied with a zoomed-in view of the area around the action to help you see changes more clearly.

- Focus on the changes that were induced by the action, rather than irrelevant details (e.g. the time change in the system clock).

- Note that even if the action is expected to cause a change, it may have not. Never assume that the action was successful without clear evidence in the screenshots.

- Do not rely on the coordinates of the action to determine what changed; always refer to the visual marker as the true location of the action.

- Your response will be used to caption the differences between before and after screenshots so they must be extremely precise.

Please format your response as follows below.

(thoughts)

[Your detailed reasoning about the before screenshot and any visual markers, the action being taken, and the changes in the after screenshot and zoomed-in view (if present).]

(thoughts)

(answer)

[An unordered list of the relevant changes induced by the action]

(answer)

\end{tcolorbox}
\end{figure}

\begin{figure}[h]
\centering
\small
\begin{tcolorbox}[colback=orange!5, colframe=red!60!black, title=Prompt for Comparative Evaluation(Comparative Scoring Prompt), fonttitle=\bfseries]
You are evaluating \{num\_trajectories\} different action sequences to determine which one makes the most progress toward completing this task:

**Task:** \{instruction\}

**CRITICAL: Show, Don't Tell**

You must base your judgment ONLY on the provided screenshots and explicit text output visible in them. Do NOT assume the task is done just because the agent says so (or because the facts claim so). If the agent ran a verification command (like ls or cat), you MUST see the output in the terminal screenshot confirming the result. If the proof is not visible, do not give a high score.

**Evaluation Criteria:**

1. Does the sequence move closer to completing the task?

2. Are the actions relevant and purposeful?

3. Does the final state show VISIBLE progress toward the goal?

**Differentiation Requirement:**

For nodes that appear very similar (in terms of action content, trajectory history, and produced results), please **carefully compare their subtle differences**. Look for efficiency, unnecessary steps, or minor visual improvements. **You must assign differentiated scores based on this careful analysis.** Avoid giving identical scores unless the trajectories are absolutely indistinguishable. Ideally, the scores should clearly reflect which trajectory is better, even if only slightly.

You will be provided:

- Initial screenshot (starting state)

- For each trajectory: fact captions describing what changed + final screenshot

**Output Format:**

Provide scores from -1.0 to 1.0 for each trajectory, where:

- 1.0 = Completes or nearly completes the task

- 0.5 to 0.9 = Makes significant progress

- 0 to 0.5 = Makes some progress

- -0.5 to 0 = Little or no progress / wrong direction

- -1.0 to -0.5 = Harmful or completely wrong

Return ONLY a JSON array of scores, e.g., [0.8, -0.5, -0.6]
\end{tcolorbox}
\end{figure}

\end{document}